\documentclass[12pt,twoside,a4paper]{article}

\newcommand{\IpiArchiv}[1]{#1}

\usepackage[OT4]{fontenc}
\usepackage[cp1250]{inputenc}
\usepackage[dvips]{color}
\usepackage{amsfonts}
\usepackage{amsmath}
\usepackage{amsthm}
\usepackage{graphicx}
\usepackage{hyperref}
\usepackage{float}
\usepackage{url}
\usepackage{array}
\usepackage{algorithm,algpseudocode}

\IpiArchiv{\usepackage{longtable}}
\newcommand{\nonIpiArchiv}[1]{#1}
\IpiArchiv{\renewcommand{\nonIpiArchiv}[1]{}}

\newcommand{\figaddr}[1]{#1}

\newcommand{\Bem}[1]{}
\newcommand{\Kuerze}[1]{}

\newtheorem{definition}{Definition}%[section]
 
\newtheorem{theorem}{Theorem}
\begin{document}
\newcommand{\MoovjTytulv}{
An Aposteriorical Clusterability Criterion for 
$k$-Means++ 
and Simplicity of Clustering \IpiArchiv{- Extended Version}}
\newcommand{\MaInstytucja}{\small Institute of Computer Science of the Polish Academy of Sciences\\\small  ul. Jana Kazimierza 5, 01-248 Warszawa 
Poland\\ \small \url{ klopotek@ipipan.waw.pl  }}
\title{
\MoovjTytulv  }
%\titlerunning{\MoovjTytulv}
\author{
Mieczys{\l}aw A. K{\l}opotek  
\\
\MaInstytucja
}
%\institute{\MaInstytucja}
\newcommand{\sto}{ d }
 
\maketitle

\begin{abstract}
We define the notion of a well-clusterable data set combining  the point  of view of the objective of $k$-means clustering algorithm (minimising the centric spread of data elements) and common sense (clusters shall be separated by gaps).
We identify conditions under which the optimum of $k$-means objective coincides with a clustering under which the data is separated by predefined gaps.

We investigate two cases: when the whole clusters are separated by some gap and when only the cores of the clusters meet some separation condition. 

We overcome a major obstacle in using clusterability criteria due to the fact that
known approaches to clusterability checking had the disadvantage that they are related to the optimal clustering which is NP hard to identify.

Compared to other approaches to clusterability, the novelty consists in the possibility  of an  a posteriori (after running $k$-means) check  if the data set is well-clusterable or not. 
As the $k$-means algorithm applied for this purpose has polynomial complexity so does therefore the appropriate check.
Additionally, if $k$-means++ fails to identify a clustering that meets clusterability criteria, with high probability the data is not well-clusterable.

\end{abstract}
% keywords:   k-means, clusterability, well-clusterable
\noindent

%-----------------------------------

\section{Introduction}
It is a commonly observed phenomenon that most practically used clustering algorithms (like $k$-means) have a high theoretical computational complexity (are NP-hard), but at the same time in many (though not all) practical applications they perform quite well (converge quickly enough) yielding more or less usable output. Apparently then, the data must have the property that some data sets are better clusterable than other. 

Though a number of attempts have been made to capture formally the intuition behind clusterability, none of these efforts seems to have been successful, as Ben-David exhibits in \cite{Ben-David:2015} in depth. He points at three important shortcomings of current state-of-the-art research results:  the clusterability cannot be checked prior to applying a potentially NP-hard clustering algorithm to the data,  known clusterability criteria impose strong (impractical) separation constraints and the research nearly does not address popular algorithms. A recent paper by Ackerman \cite{Ackerman:2016} partially eliminates some of these problems, but regrettably at the expense of introducing user-defined parameters that do not seem to be intuitive (in terms of one's imagination about what well-clusterable data are.).

Therefore in this paper we try a different approach to defining what well clusterable data are. 
As Ben-David mentioned,   the research in the area  does not address popular algorithms  except for $\epsilon$-Separatedness clusterability criterion related to $k$-means proposed  by Ostrovsky et al. \cite{Ostrovsky:2013}.
Herewith we want also to contribute to applicability of clusterability criteria in that, following Ostrovsky's example, we deal with $k$-means, and in particular with its special version called $k$-means++, as in fact Ostrovsky did. 
Furthermore, we share  Ben-David's concern that it is not a solution to the problem if we shift the NP hardness from the data clustering algorithm to the data clusterability checking algorithm, because the problem becomes even worse. 
Last not least we have to handle somehow the issue of the impractical gaps imposed by the clusterability criteria  in the literature. 
Ben-David argues in \cite{Ben-David:2015} (see his Section 5), that apparently the efforts in clusterability research on finding support for the hypothesis of "clustering is hard only if the clustering does not matter" have failed, mainly due to the fact that gaps between clusters that are required are too  large for practical applications, as the popular algorithms behave reasonably even with significantly smaller gaps. 
But a closer look at the $k$-means objective shows (see Section \ref{sec:gapinsufficiency}) that it does not make sense to build a clusterability criterion solely on the grounds of the gaps because $k$-means criterion does not rely solely on gaps between clusters, but also on their cardinalities as well as in fact on their internal spread, which cannot be known in advance. Under changed cardinalities $k$-means may prefer to split larger clusters and merge parts of them with smaller, though clearly separated clusters. If the $k$-means criterion shall coincide with separation of clusters, the gaps need to be large. 
Therefore in our research, instead of seeking smaller gaps, we rather concentrate on redefining the goal of clusterability research efforts.   
So it is proposed here to change the perspective. 
Instead of (or in addition to) seeking conditions for easiness of clustering of a given data set, 
let us look for definition of a data set (data set generator) for which we know optimal clustering in advance and  some algorithm returns the ground truth nearly for sure. 
This change of perspective will lead us to a practical application of clusterability concept consisting in testing the algorithm behaviour under varying degrees of violating the clusterability conditions. 

Having been freed from the need to seek the smallest possible gaps, we can also weaken the problems with NP hardness of stating the clusterability of the data set. 
In particular we do not require that we have to say beforehand (before clustering) whether or not the data is well-clusterable. Instead we require that one shall be able to state aposteriorically \emph{whether or not} the data is well-clusterable according to well-clusterability criteria that were assumed, in polynomial time. 
Note that this is a tremendous progress over clusterability criteria defined so far. 
None of the clusterability criteria discussed by 
Ben-David \cite{Ben-David:2015} fits this requirement and the criterion proposed in  \cite{Ackerman:2016} can be shown to be invalid for simple data sets (see Section \ref{sec:gapinsufficiency}, Figure \ref{fig:antiAckerman}). 
We believe to resolve in this way a serious bottleneck in the clusterability research. 
We do not cover here the issue of measuring the deviation from well-separatedness, but are convinced that by presenting a clusterability criterion that is verifiable ex post and where the data can be checked for clusterability by at least one popular algorithm we open a way to attack this issue also. 

In this study we will restrict ourselves to the $k$-means family of algorithms.
The restriction is in fact not too serious as the algorithms of this family are broadly used and in fact there exist quite large number of variants, starting with early work of Lloyd, Forgy,  MacQueen and    
Hartigan-Wong, to $k$-means++, spherical $k$-means, their fuzzified versions, and many other\footnote{See e.g. Chapter 3 \cite{Wierzchonv:2015} for a review of the $k$-means algorithm family} 

Within the $k$-means family we have to face the following challenges \cite{Ben-David:2015}:
\begin{itemize}
\item the clusterability criteria in the literature (e.g. \cite{Ostrovsky:2013}) refer to the optimal cost function value of $k$ means (see equation (\ref{eq:Q::kmeans})) - but the actual value of this optimal solution is not known
\item people are accustomed to associate well-clusterable data with ones of large gaps between clusters - but the optimal cost function of $k$-means is also influenced by cluster sizes, so that the gap sufficient for one set of clusters will prove insufficient for some other (see Section \ref{sec:gapinsufficiency})
\item the cost function of $k$-means usually has multiple local minima and the real world $k$-means algorithms usually tend to stick at some local minimum
(see e.g.  \cite[Chapter 3]{Wierzchonv:2015}). 
\end{itemize}

For these reasons, when comparing the results of various $k$-means brands on real data we have a hard time to distil the reason why their results differ: is it because the data are not clusterable, or that the cost function optimum does not agree with common sense split into well separated clusters or the algorithm is unable to discover the optimal clustering (systematically misses it). 

In order to enable making such distinctions, we decided to seek such a clusterability criterion that:
\begin{itemize}
\item the clusterability criterion is based on the gap size between clusters and other cluster characteristics, that can be computed by inspection of an obtained clustering (not referring to the optimal one),  
\item if the clustering obtained meets the clusterability criteria, then this is the real optimal clustering,
\item if a special algorithm (here we mean $k$-means++)  fails to find a clustering meeting our clusterability criteria, then with high probability the data is not well-clusterable at all by any algorithm, 
\item there exists the possibility to generate a data set matching the clusterability conditions for various constellations of the cluster sizes (cardinality, spread), dimensionalities, number of clusters etc.
\end{itemize}

Given such a tool at disposal we can investigate algorithm's capability to find the optimal clustering in the easy case, compare the algorithms in their performance in an easy case, and then compare their relative performance when the clusterability property degenerates, for example via decreasing the size of the gap between clusters. 

In this research we confine ourselves to providing the tool in terms of the new clusterability criterion, and make only a small demonstration, how the degenerative behaviour of algorithms may be studied.  

Our contribution encompasses:
\begin{itemize}
\item Two brands of well-clusterability criteria for data to be clustered via $k$-means algorithm, that can be verified ex-post (both positively and negatively) without great computational burden
(inequalities (\ref{eq:globalgcase2})  and (\ref{eq:globalgcase1}) in Section \ref{sec:wellclusteredsep},
and  inequalities (\ref{eq:globalgcase1core}) and  (\ref{eq:globalgcase2core}) in Section \ref{sec:core}). 
\item Demonstration, that the structure of well-clusterable data (according to these criteria)  is easy to recover
(see Theorems \ref{th:clusterabilitydecidable}(i) and \ref{th:coreclusterabilitydecidable}(i)).  
\item Demonstration that if well-clusterable data structure (in that sense) was not discovered by $k$-means++, then there is no such structure in the data (with high probability - see Theorems \ref{th:clusterabilitydecidable}(ii) and \ref{th:coreclusterabilitydecidable}(ii)).  
\item Demonstration that large gaps between data clusters are not sufficient to ensure well-clusterability by $k$-means
(see Section \ref{sec:gapinsufficiency}). 
\end{itemize}

The structure of this paper is as follows: In Section \ref{sec:previous} we recall the previous work on the topic of clusterability and give a brief introduction to the $k$-means algorithm and its special case $k$-means++. 
In Section \ref{sec:gapinsufficiency} we show that large gaps are not sufficient for well-clusterability. 
In Section \ref{sec:wellclusteredsep} we introduce the first version of well-clusterability concept and show that data well-clustered in this sense are easily learnable via $k$-means++.  This concept has the drawback that no data points (outliers) can lie in wide areas between the clusters. Therefore in Section \ref{sec: wellclusteredcore} we propose a core-based well-clusterability concept and show that data well-clustered in this sense are also easily learnable via $k$-means++. The concept of cluster core itself is introduced and  investigated in Section \ref{sec:core} and a method determining proper gap size under these new conditions is derived in Section \ref{sec:corebasedglobalkmeansminimum}. 
In Section \ref{sec:exper} some experimental results are reported concerning 
performance of various brands of $k$-means algorithms for data fulfilling the clusterability criteria proposed in this paper. 
  Section \ref{sec:discussion} contains a brief comparison of our clusterability criteria with those discussed by Ben-David \cite{Ben-David:2015}. 
In Section \ref{sec:conclusions} we draw some conclusions from this research.

\section{The  problem of clusterability in the previous work}\label{sec:previous}
Intuitively the clusterability shall be a function taking a set of points and returning a real value saying how "strong" or "conclusive" is the clustering structure of the data~\cite{Ackerman:2009}. This intuition, however, turns out not to be formalized in a uniform way so that quite a large number of formal definitions have been proposed. 
  Ackerman and Ben-David in \cite{Ackerman:2009} studied several of these notions.
They concluded that across the various formalizations, two phenomena co-occur: on the one hand well-clusterable data sets
(with high "clusterability" value) are computationally easy to cluster (in polynomial time), but on the other hand identification whether or not the data is well-clusterable is NP-hard. 

Ben-David 
\cite{Ben-David:2015} 
performed an interesting investigation of the concepts of clusterability from the point of view of the capability of "not too complex" algorithms to discover the cluster structure,  (negatively) verifying the working hypothesis that  “Clustering  is difficult  only
when it does not matter” (the $CDNM$ thesis). 

He considered   the following notions of clusterability, present in the literature:
\begin{itemize}
\item \emph{Perturbation Robustness} meaning that   small
perturbations of distances / positions in space of set elements  do not result in a change of the optimal clustering for that data 
set. Two brands may be distinguished: additive \cite{Ackerman:2009} and multiplicative ones \cite{Bilu:2012} (the limit of perturbation is upper-bounded either by an absolute value or by a coefficient).  
\item \emph{$\epsilon$-Separatedness} meaning that the cost of optimal  clustering into $k$ clusters is less than $\epsilon^2$ times the cost of optimal clustering into $k-1$ clusters \cite{Ostrovsky:2013} - here an explicit reference to the $k$-means objective is made.
\item \emph{
$(c, \epsilon)$-Approximation-
Stability} \cite{Balcan:2009} meaning that if the cost function values of two partitions differ by the factor $c$, then the distance (in some space) between the partitions is at most $\epsilon$.
As Ben-David recalls, this implies the uniqueness of optimal solution.
\item \emph{
$\alpha$-Centre Stability} \cite{Awasthi:2012} meaning, for any centric clustering, that the distance of an element to its cluster centre is $\alpha$ times smaller than the distance to any other cluster centre under optimal clustering. 
\item \emph{$(1+\alpha)$ Weak Deletion Stability} \cite{Awasthi:2010} meaning that given an optimal cost function value $OPT$ for $k$ centric clusters, then the cost function of a clustering obtained by deleting one of the cluster centres and assigning elements of that cluster to one of the remaining clusters should be bigger than  $(1+\alpha)\cdot OPT$.
\end{itemize}

 Under these notions of clusterability algorithms have been developed clustering the data nearly optimally in polynomial times, when some constraints are matched by the mentioned parameters. 
 
However, these conditions seem to be rather extreme.
For example, given the $(c, \epsilon)$-Approximation-
Stability~\cite{Balcan:2009}, 
polynomial time clustering requires that,  in the optimal
clustering (beside its uniqueness),     all  but an
$\epsilon$-fraction  of  the  elements,  
are 20 times closer to their own cluster centre than to every other cluster centre. 
  $\epsilon$-Separatedness requires that the distance to its own cluster centre must be at least 200 times closer than to every other cluster element \cite{Ostrovsky:2013}. And this is still insufficient if the clusters are not balanced. A ratio of $10^7$ is deemed by these authors as sufficient.  
($1+\alpha$) Weak Deletion Stability  \cite{Awasthi:2010} demands  distances to other clusters being 
$\log(k)$  
times the "average radius" of the own cluster.
The perturbational stability \cite{Ackerman:2009} induces exponential dependence on the sample size. 

Anyway,  we can  draw a certain important conclusion from these concepts of clusterability mentioned above:
People agree that  a data set is well clusterable if each cluster  is distant (widely separated) from the other clusters.

This idea occurs in many other clusterability concepts. 
Epter et al. \cite{Epter:1999} considers the data as clusterable  when the minimum between-cluster separation exceeds the maximum in-cluster distance (called elsewhere "perfect separation").%  
\footnote{It has been shown in the literature that under this notion of well-clusterability single link algorithm can detect clusters separated in such a way. It has also been shown that centre based algorithms like $k$-means may fail to detect such clusters, see e.g. \cite{Ackerman:2014nips}. 
}
Balcan et al. \cite{Balcan:2008} proposes to consider data as clusterable if   each element is closer to all elements in its cluster than to all other data (called also "nice separation").\footnote{It has been shown in the literature that this notion of well-clusterability is hard to decide in a data set, see e.g. \cite{Ackerman:2014nips}. } 
Interestingly, $k$-means reflects the Balcan concept "on average" that is each element average squared distance to elements of the same cluster is smaller than the minimum (over other clusters) averaged squared distance to elements of a different cluster. 
  Kumar and  Kannan \cite{Kumar:2010}, explicitly concentrating on $k$-means objective, define clusterability via a proximity condition stating that any point projected on a line connecting its own cluster centre and some other cluster centre should be closer to its own cluster centre by a "sufficiently large" gap depending on the number of clusters and inverted squared cluster cardinalities. 
 
Kushagra et al. \cite{Kushagra:2016} consider clusterability from the point of view of a structure in the data. They allow for noise in the data, but insist that the noise does not create structures by itself. They refrain from optimising a cost function. They show that without assumption of structure in the data or without assumption of structureless noise discovery of clusters is not possible.

 Ackerman  and  Dasgupta \cite{Ackerman:2014nips}
move the focus on clusterability from the clusterability as a property of the data alone to the pair of 
(data type, algorithm type). 
In that paper, they  are  interested in incremental algorithms only and show that an incremental version of $k$-means performs poorly under perfect and nice separation.  

In the same spirit  Ben-David and   Haghtala \cite{Ben-David:2014} investigated
clusterability by $k$-centroidal algorithms (a class of algorithms including $k$-means) via robustifying  an algorithm against noise in the data by either clustering the noise into separate clusters or cutting off too distant points.

Ackerman et al. \cite{Ackerman:2013} 
consider the clusterability from the perspective of distortion of clusters by malicious points. It turns out that from this perspective $k$-means performs better than various other algorithms. With respect to our research they also insist that the proportions between cluster sizes play a significant role ensuring proper clustering.

Cohen-Addad \cite{Cohen-Addad:2017} raises the claim that data are clusterable (in terms of various stability criteria) if the global clustering can be well approximated by local one. 
 Our work can be perceived in this spirit in that we try to achieve coincidence of clusters based on separability with global cost function minimum. 

Tang \cite{Tang:2016} investigates a clusterability criterion for his own version of  $k$-means, based on the requirement that the cluster centres are separated by some distance, which is dependent upon \emph{ground truth optimal clustering}.

Recently   Ackerman et al. \cite{Ackerman:2016} derived a method for testing clusterability of data based on the large gap assumption.  They investigate the histogram of (all) mutual dissimilarities between data points. If there is no data structure, the distribution should be unimodal. If there are distant clusters, then there will occur one mode for short distances (within clusters) and at least one for long distances (between clusters). 
Hence, to detect clusterability, they apply  tests of multimodality, namely the Dip \cite{Hartigan:1985} and Silverman \cite{Silverman:1981} tests.

But the criterion of a sufficiently large gap between clusters is not reflected in various clustering function objectives, like for example $k$-means which may reach an optimum with poorly separated clusters in spite of the fact that there exists an alternative partition of data with a clear separation between clusters in the data, as we will demonstrate in Section \ref{sec:gapinsufficiency}.  Also in Section \ref{sec:gapinsufficiency} we will demonstrate, that multimodal distributions can be detected by Ackerman's method even if there is no structure in the data. 
   
Ben-David 
\cite{Ben-David:2015} raises a further important point  
 that it is usually (in practically all above mentioned methods except \cite{Ackerman:2016}, which has a flaw by itself) impossible to verify apriori if the data fulfils the clusterability criterion because the conditions refer either to  all possible clusterings or to optimal clustering so that we do not have the possibility to verify whether or not
the data set is clusterable, before one starts clustering (but usually computing the optimum is NP-hard).  

In this paper, however, we would like to stress that the situation is even worse. Even at the termination of the clustering algorithm we are unable to say whether or not the clustered data set turned out to be well-clusterable. 
For example, the 
  $\epsilon$-Separatedness criterion requires that we know the   nearly optimal solution for clustering into $k$ and $k-1$ elements. 
While we can usually get the upper approximations for the cost functions in both cases, we need actually the lower approximation for $k-1$ in order to decide ex post if the data was well-clusterable, and hence whether or not we can say that we approximated the correct solution in some way. But we get it only for $k=2$, hence for higher $k$ the issue is not decidable. 
Tang's  \cite{Tang:2016} criterion is certainly better, though also based on solution to optimality criterion,  because we can sometimes decide ex-post that the clusterability criterion was fulfilled (the distance between clusters needs to be greater than a product of optimal clustering cost function and reversed squared roots of cluster cardinalities, which may be upper-bounded by the actual clustering cost function and the number 2). Still in this case upon finding the optimal clustering we will be still unsure that it is so even if the clusterability criterion is met. 

The issue of ex-post decision on clusterability seems nevertheless to be simpler to solve than the apriorical one, therefore we will attack it in this paper. We are unaware that such an issue was even raised in the past. 
Though the criteria of \cite{Epter:1999} and   \cite{Balcan:2008} can clearly be applied ex post  to see that in the resulting clustering the clusterability criteria hold, but these approaches lack the solving of the inverse issue: what if the clusterability criteria are not matched by the result clustering - is the data unclusterable? Could no other algorithm discover the clusterable structure?

One shall note at this point that the approach in \cite{Ackerman:2016} is different with this respect. 
Compared to methods requiring finding the optimum first, Ackerman's approach seems to fulfil Ben-David requirement, that we can see if there is clusterability in the data before starting the clustering process as the clusterability method  is  computationally optimal   because the computation of the histogram of dissimilarities is quadratic in sample size. 
But at an in-depth-investigation, the Ackerman's clusterability determination method misses one important point: it requires a user-defined parameter and the user may or may not make the right guess. Furthermore, even if clusterability is decided by Ackerman's tests, it is still uncertain if $k$-means algorithm will be willing to find such a clustering that fits Ackerman's clusterability criterion.  Beside this, as visible in Figure \ref{fig:antiAckerman}, one can easily find counterexamples to their concept of clusterability.
The left image shows that there is a single cluster there, but the histogram to the right has two modes, indicating a two-cluster structure. 

So in summary the issue of an aposteriorical determination if the data were clusterable, remains an open issue. 
 
Therefore it seems to be justified to restrict oneself to a problem as simple as possible in order to show that the issue is solvable at all. 
So in this paper we will limit ourselves to the issue of clusterability for the purposes of $k$-means algorithm.%
\footnote{ The $k$-means algorithm  seems to be quite popular in various variants both in traditional, kernel and spectral clustering. 
Hence the results may be still of sufficiently broad importance. } Furthermore we restrict ourselves to determine such cases when the clusterability is decidable "for sure".

The first problem to solve seems to be to get rid of the dependence on the undecidedness of  optimality of the obtained solution. 

But before proceeding let us recall the $k$-means cost function definition. 

 \begin{equation} \label{eq:Q::kmeans}
Q(\mathcal{C})=\sum_{i=1}^m\sum_{j=1}^k u_{ij}\|\textbf{x}_i - \boldsymbol{\mu}_j\|^2
=\sum_{j=1}^k \frac{1}{n_j} \sum_{\mathbf{x}_i, \mathbf{x}_l \in C_j} \|\mathbf{x}_i - \mathbf{x}_l\|^2 
\end{equation} 
for a dataset $\mathbf{X}$
under some partition $\mathcal{C}=\{C_1,\dots,C_k\}$ into the predefined number $k$ of clusters, $C_1\cup \dots\cup C_k=\mathbf{X}$, 
where  $u_{ij}$ is an indicator of the membership of data point $\textbf{x}_i$ in the cluster $C_j$ having the centre at $\boldsymbol{\mu}_j=\frac{1}{|C_j|}\sum_{\textbf{x}\in C_j}\textbf{x}$.

The \textbf{$k$-means algorithm} starts with some initial guess of the positions of  $\boldsymbol{\mu}_j$ for $j=1,\dots,k$ and then alternating two steps: cluster assignment and centre update till some convergence criterion is reached, e.g. no changes in cluster membership. The cluster assignment step updates $u_{ij}$ values so that each element $\textbf{x}_i$ is assigned to a cluster represented by the closest $\boldsymbol{\mu}_j$.
The centre update step uses the update formula $\boldsymbol{\mu}_j=\frac{1}{|C_j||}\sum_{\textbf{x}\in C_j}\textbf{x}$.

The \textbf{$k$-means++} algorithm is a special case of $k$-means where the initial guess of cluster centres proceeds as follows. $\boldsymbol{\mu}_1$ is set to be a data point uniformly sampled from $\mathbf{X}$. The subsequent cluster centres are data points picked from $\mathbf{X}$ with probability proportional to the squared distance to the closest cluster centre chosen so far. For details check \cite{CLU:AV07}. Note that the algorithm proposed by \cite{Ostrovsky:2013} differs from the $k$-means++ only by the non-uniform choice of the first cluster centre (the first pair of cluster centres should be distant, and the choice of this pair is proportional in probability to the squared distances between data elements).

\newcommand{\CMDFIGantiAckerman}{ 
\begin{figure}
\centering
 (a)\includegraphics[width=0.4\textwidth]{\figaddr{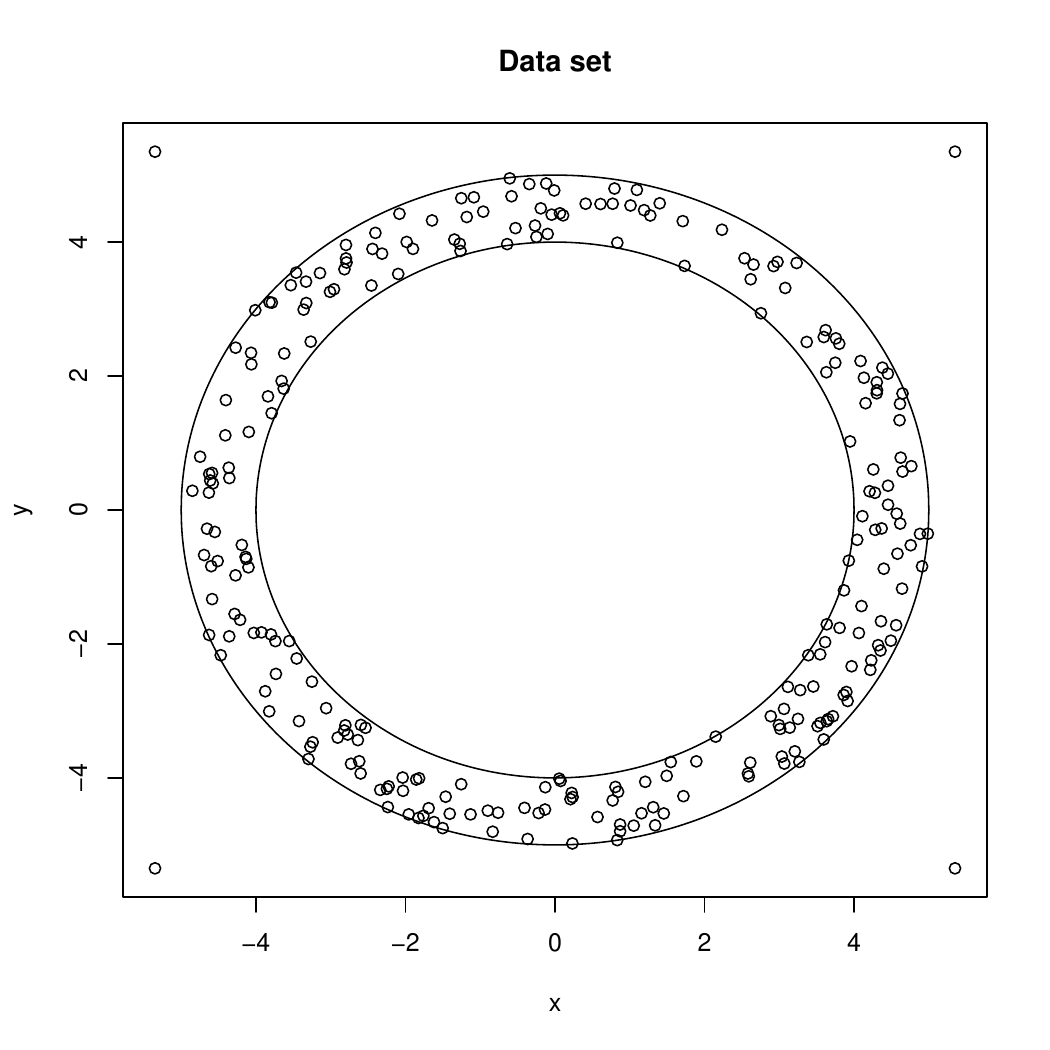}}  %
(b)\includegraphics[width=0.4\textwidth]{\figaddr{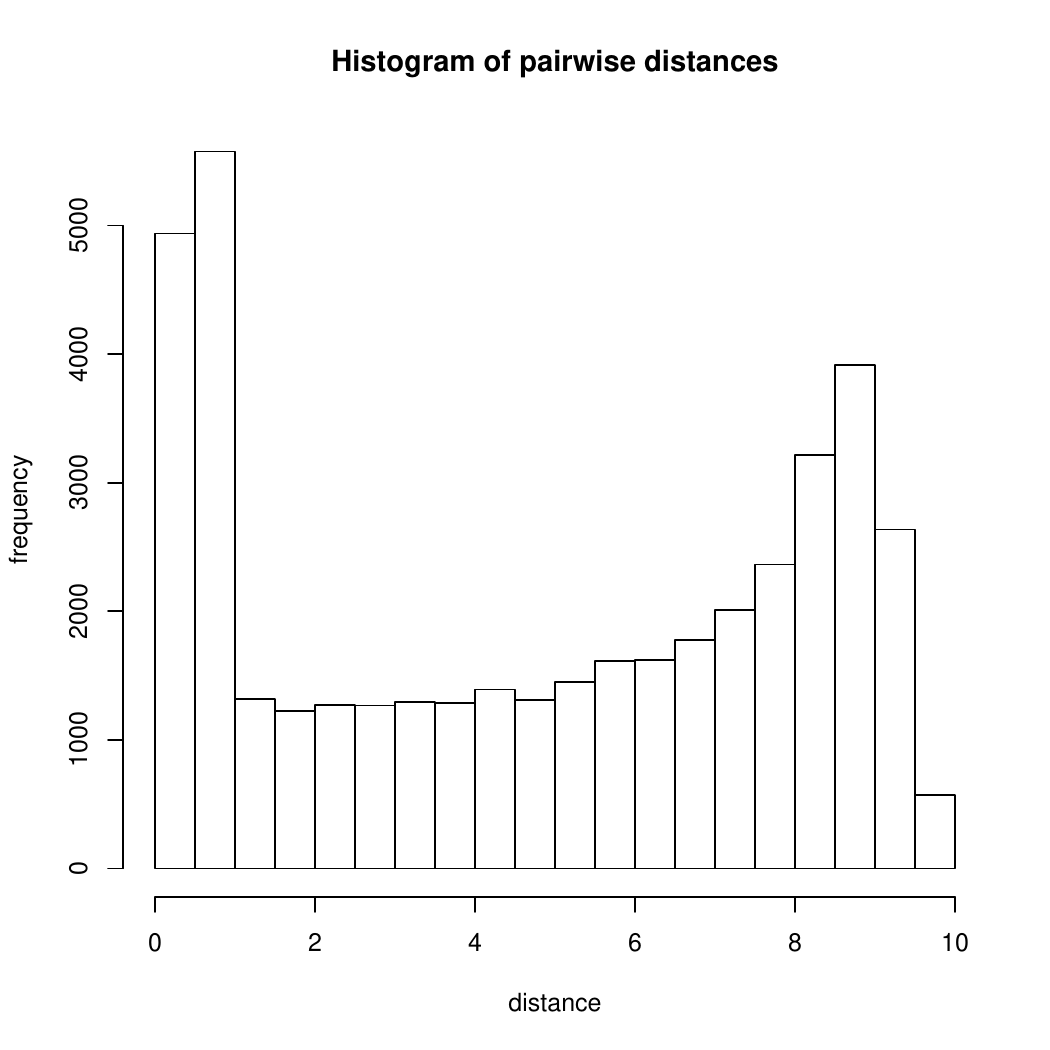}}  %
\caption{Illustration of a special case where Ackerman's method \cite{Ackerman:2016} would falsely recognize clustering structure in the data (a) the data (b) the histogram of pair-wise distances - two modes visible
}\label{fig:antiAckerman}
\end{figure}
}

\section{Non-suitability of gap-based  clusterability criteria for $k$-means}\label{sec:gapinsufficiency}
\CMDFIGantiAckerman
Let us discuss more closely the relationship between the gap-based well-clusterability concepts developed in the literature and the actual optimality criterion of $k$-means. 
Specifically let us consider  the   approaches to clusterability of \cite{Ackerman:2016}, 
 \cite{Epter:1999},  \cite{Awasthi:2012}   and  \cite{Balcan:2008}.

Human intuition will tell us that if the groups of  data points occur in the data and there are large spaces between these groups, then it should be these groups that will be chosen as the actual clustering. On the other hand if there are no gaps between the groups of data points, then one would expect that the data are not considered as well-clusterable.  
Furthermore, if the data is well-clusterable, one would expect a reasonable clustering algorithm to  discover easily such a well-clusterable data structure. 

However, these intuitions prove wrong in case of $k$-means. 

Let us first point to the fact that \cite{Ackerman:2016} may indicate a clear bimodal structure in the data where there are no gaps in the data. 
   We are unaware of anybody pointing at this weakness of well-clusterability in \cite{Ackerman:2016}:
Imagine a thin ring uniformly covered with data points (see Figure \ref{fig:antiAckerman}(a)). We would be reluctant to say that there is a clustering structure in such data. Nonetheless we will see two obvious modes in such data.   The thinner the ring (the closer to a circle), the more obvious the reason for the multimodality will be: we will get closer and closer to the following  function.
Consider the angle $\alpha$ centred at the centre of the circle ("thin ring"). 
As we are interested in calculating distances between points, we restrict ourselves to angles with measure $0^o\le \alpha\le 180^o$ (or $0\le \alpha\le \pi$). 
The number of elements within the angle would be approximately proportional to this angle. 
The distance between cutting points of this angle on the circle, given a   radius $r$ of the circle, will amount to 
$x=2r\sin \frac{\alpha}{2}$. 
Consequently $\alpha =2\arcsin \frac{x}{2r}$. 
To determine the density of distances we need to compute a   derivative
$\frac{d \alpha}{dx} =\frac{1}{dx/d\alpha} $
$= \frac{1}{d(2r\sin \frac{\alpha}{2}) /d\alpha}
$
$= \frac{1}{r\cos  \frac{\alpha}{2} }
$
$= \frac{1}{r\sqrt{1-\sin^2   \frac{\alpha}{2} }}
$
$= \frac{1}{r\sqrt{1- \frac{x^2}{4r^2}  }}
$.  
This function has a minimum at $x=r$ and grows towards both $x=0$ and $x=2r$.
If it is actually not a circle, but a ring, more distances close to zero occur, hence the shape of the histogram. In our case the radius was 5, so we need to multiply these numbers with 5 to get what is visible in the histogram in Figure \ref{fig:antiAckerman}(b).

On the other hand, even if there are gaps between groups of data, for example those required by   \cite{Epter:1999},  \cite{Awasthi:2012} or \cite{Balcan:2008}, $k$-means optimum may not lie in the partition exhibiting gap based well-clusterability property in spite of its existence, And not only for these gaps, but also for any arbitrary many times larger ones.   
As  \cite{Epter:1999} is concerned, it may be considered as a special case of   \cite{Balcan:2008}.  
 \cite{Awasthi:2012} may be viewed in turn as a strengthening of the concept of
\cite{Epter:1999}.  
So let us discuss a situation in which both perfect and nice separation  criteria are identical that is of two clusters. 
We will show that whatever $\alpha$ we assume in the $\alpha$-stability concept, $k$-means fails to be optimal under unequal class cardinalities. 
Let these clusters, $C_A,C_B$ be each enclosed in a ball of radius $r$ and the distance between ball centres should be at least $4r$. 
We have demonstrated in \cite{MAK:2017:kleinbergaxioms} that under these circumstances the clustering of data into $C_A,C_B$ reflects a local minimum of $k$-means cost function. 
But it is not the global minimum, as we will show subsequently. 
So at least for $k$-means the criteria of Epter and Balcan and Awasthi   cannot be viewed as realistic definitions of well-clusterability. 
Subsequently, whenever we say that a cluster is enclosed in a ball of radius $r$, we mean at the same time that the ball is centred at gravity centre of the cluster. 

For purposes of demonstration we assume that 
both clusters are of different cardinalities $n_A, n_B$
and let $n_A> n_B$. We show that whatever distance between both clusters, we can get such a proportion of $n_A/n_B$ that the clustering into $C_A,C_B$ is not optimal.

Let us consider a $d$-dimensional space.
Let us select the dimension that contributes most to the variance in cluster $C_A$. 
So the variance along this direction  amounts to at least the overall variance divided by  $d$.
Let us denote this variance component as $V_d$. 
Consider this coordinate axis responsible for $V_d$ to have the origin at the cluster centre of $C_A$. 
Project all the points of cluster $C_A$ on this axis. The variance of projected points will be just $V_d$. 
Split the projected data set into two parts $P_1,P_3$, one with coordinate above 0 and the rest. 
Let the centres of $P_1,P_3$ lie $x_1,x_3$ away from the cluster centre.
Let   $n_1$ data points of $P_1$ be at most $x_1$ distant from the origin, 
and $n_2$ more than $x_1$ from the origin. 
Let there be $n_3$ data points of $P_3$ be at most $x_3$ distant from the origin, 
and $n_4$ more than $x_3$ from the origin. 
Obviously, $n_1+n_2+n_3+n_4=n_A$,
  $|P_1|=n_1+n_2$, 
$|P_3|=n_3+n_4$.
As zero is assumed to be the $C_A$ cluster centre on this line, also  
$x_1 \cdot(n_1+n_2)=x_3 \cdot(n_3+n_4)$ holds. 
Furthermore, as the cluster is enclosed in a ball of radius $r$ centred at its gravity centre, 
both $x_1\le r$ and $x_3\le r$.  
Under these circumstances, let us ask the question whether for a 
$V_d$ some minimal values of $x_1,x_3$ are implied. 
Because if so, then by splitting the cluster $C_A$ into $P_1,P_3$ and by increasing the cardinality of $C_A$, the split of data into $P_1, P_2\cup  C_B$ will deliver a lower $Q$ value so that for sure the clustering into $C_A,C_B$ will not be optimal. 
 
Note that $V_d=(Var(P_1)+x_1^2)\cdot (n_1+n_2)+ (Var(P_3)+x_3^2)\cdot (n_3+n_4))/n_A$.
The $n_1$ points of $P_1$ closer to origin than $x_1$ are necessarily   
not more than $x_1$ distant from $P_1$ gravity centre.
Therefore, the remaining $n_2$ points cannot be more distant than $x_1\frac{n_1}{n_2}$.  
Hence $Var(P_1)\le x_1^2 n_1+\left(x_1\frac{n_1}{n_2}\right)^2n_2$. 
By analogy
  $Var(P_3)\le x_3^2 n_3+\left(x_3\frac{n_3}{n_4}\right)^2n_4$.

So we observe that 
$$V_d\le 
\frac{ \frac{x_1^2 (n_1 +n_1^2/n_2 +n_1+n_2)}{ n_1+n_2}\cdot (n_1+n_2)+ 
          \frac{x_3^2\cdot (n_3 +n_3^2/n_4 +n_3+n_4)}{(n_3+n_4)}  \cdot  (n_3+n_4) }
{n_A}
$$
that is 
$$V_d\le 
\frac{
(x_1^2\cdot (n_1 +n_1^2/n_2 +n_1+n_2) + 
          x_3^2\cdot (n_3 +n_3^2/n_4 +n_3+n_4) }
{n_A}
 $$

Note that we can delimit $n_2, n_4$ from below due to the relationship:
$  (r-x_1)\cdot n_2\ge n_1\cdot x_1$, 
$(r-x_3)\cdot n_4\ge n_3\cdot x_3$.

Therefore
$$V_d\le 
 (x_1^2\cdot (n_1 +n_1^2\cdot (r-x_1)/n_1/x_1 +n_1+n_2) + 
          x_3^2\cdot (n_3 +n_3^2\cdot (r-x_3)/n_3/x_3 +n_3+n_4) )/
n_A
$$

Hence

$$V_d\le 
(x_1^2\cdot (2\cdot n_1+n_2) +n_1^2\cdot (r-x_1)\cdot x_1/n_1   + 
          x_3^2\cdot (2\cdot n_3+n_4) +n_3^2\cdot (r-x_3)\cdot x_3/n_3  )/
n_A
$$
 
$$V_d\le 
(x_1^2\cdot (2\cdot n_1+n_2) +n_1\cdot (r-x_1)\cdot x_1  + 
          x_3^2\cdot (2\cdot n_3+n_4) +n_3\cdot (r-x_3)\cdot x_3  )/
n_A
$$

$$V_d\le (x_1^2\cdot (n_1+n_2) +n_1\cdot r\cdot x_1  + 
          x_3^2\cdot (n_3+n_4) +n_3\cdot r\cdot x_3  )/
n_A
$$
 
Recall that $x_1 \cdot(n_1+n_2)=x_3 \cdot(n_3+n_4)$. 
So we obtain equivalently 
%s
$$V_d\le (x_1^2\cdot (n_1+n_2) +n_1\cdot r\cdot x_1  + 
          (x_1\cdot (n_1+n_2)/(n_3+n_4)) ^2\cdot (n_3+n_4) +n_3\cdot r\cdot x_3  )/
n_A
$$ 
which is equivalent to
$$V_d\le (x_1^2\cdot (n_1+n_2) +n_1\cdot r\cdot x_1  + 
          x_1^2\cdot (n_1+n_2)^2/(n_3+n_4)  +n_3\cdot r\cdot x_3  )/
n_A
$$

By rearranging the terms we have:
$$V_d\le (x_1^2\cdot (n_1+n_2)\cdot n_A/(n_3+n_4) +n_1\cdot r\cdot x_1  + 
   +n_3\cdot r\cdot x_3  )/
n_A
$$

Let us increase the right hand side  by adding to the nominator 
$n_2 \cdot r\cdot x_1             +n_4 \cdot r\cdot x_3 $. 
This implies 
$$V_d\le (x_1^2                           \cdot (n_1+n_2)\cdot n_A/(n_3+n_4) 
          +(n_1+n_2)\cdot r\cdot x_1   
          +(n_3+n_4)\cdot r\cdot x_3  )/
n_A
$$ 
 
Let us substitute 
$x_1=\frac{x_3 \cdot(n_3+n_4)}{n_1+n_2}$. 

$$V_d\le \left(\left(x_3\cdot (n_3+n_4)/(n_1+n_2)\right)^2\cdot (n_1+n_2)\cdot n_A/(n_3+n_4)   
   +2\cdot (n_3+n_4)\cdot r\cdot x_3  \right)/
n_A
$$

Hence
$$V_d\le ( x_3^2  (n_3+n_4)^2/(n_1+n_2) \cdot n_A/(n_3+n_4)   
   +2\cdot (n_3+n_4)\cdot r\cdot x_3  )/
n_A
$$

We can delimit $n_1+n_2$ from below due to relationship 
$x_3 \cdot (n_3+n_4)=(n_1+n_2) \cdot x_1 \le (n_1+n_2)\cdot r$ because $x_1\le r$.  It implies that $ \frac{1}{n_1+n_2}\le \frac{  r}{x_3 \cdot (n_3+n_4)}$.  Therefore 
$$V_d\le (x_3^2\cdot (n_3+n_4)^2\cdot r/x_3/(n_3+n_4) \cdot n_A/(n_3+n_4)   
   +2\cdot (n_3+n_4)\cdot r\cdot x_3  )/
n_A
$$
which simplifies to 
$$V_d\le (x_3 \cdot r  \cdot n_A    
   +2\cdot (n_3+n_4)\cdot r\cdot x_3  )/
n_A
$$

$$V_d\le  x_3 \cdot r  \cdot (n_A    
   +2\cdot (n_3+n_4)   )/
n_A
$$
Clearly $n_3+n_4<n_A$, 
so we obtain 
$$V_d\le 
  3\cdot x_3 \cdot r  $$

This means that $$x_3\ge V_d/3/r$$ 
Now let us show that when scaling up $n_A$ it pays off to split the first cluster and to attach the contents of the second one to one of the parts of the first.
Let us increase the cardinality of $C_A$ $b$ times  simply by  replacing each data element by $b$ data elements collocated at the same place in space.
  In this way 
  we keep $V_d$ when increasing $|C_A|$.
So the sum of squared distances between  centre and elements
of the cluster $C_A$, $SSC(C_A)$
 will be kept 
below $V_d \cdot d \cdot n_A b$ ($SSC(C_A)\le V_d \cdot d \cdot n_A b$). 

Let $n_1+n_2$ be the minority among data points - then $x_1$ is larger  and $x_3$ is smaller of the two, because of $x_1 \cdot(n_1+n_2)=x_3 \cdot(n_3+n_4)$.
Let $P'_1,P'_3$ be the subsets of $C_A$ yielding upon the aforementioned projection the mentioned sets $P_1,P_3$. 
Then 
if we would split $C_A$ into $P'_1, P'_3$, the sum of squared distances 
to respective  cluster centres of $P'_1, P'_3$
would decrease by at least $ x_3^2 n_A b$, because 
 $SSC(P_1\cup P_3)-x_3^2 n_A b \ge SSC(P_1\cup P_3)-x_1^2(n_1+n_2)b-x_3^2(n_3+n_4) b \ge  SSC(P_1)+SSC(P_3)$,
and the distances  between elements of   $P'_1$ and $P'_3$ (and so respective gravity centres) are at least as big as between $P_1$ and $P_3$,
so that 
 $SSC(C_A)-x_3^2 n_A b = SSC(P'_1\cup P'_3)-x_3^2 n_A b  \ge  SSC(P'_1)+SSC(P'_3)$,

On the other hand combining $P'_1, P'_3$  with disjoint parts $P'_6,P'_7$ of   $C_B$
 will increase the sum of squared distances by at most $n_B x_5^2$, where $x_5$ is the distance between extreme elements of $C_A$ and $C_B$:
 $SSC(P'_1\cup P'_6)+SSC(P'_3\cup P'_7)\le SSC(P'_1)+| P_6|x_5^2+SSC(P'_3)+| P'_7|x_5^2
=SSC(P'_1)+ SSC(P'_3)+n_Bx_5^2$. 

Combining these two relations we get  
$$SSC(P'_1\cup P'_6)+SSC(P'_3\cup P'_7)\le 
SSC(C_A)-x_3^2 n_A b +n_Bx_5^2$$

Therefore, as soon as we set $ b \ge
  \frac{n_Bx_5^2}{(V_d/3/r)^2 n_A}
\ge 
  \frac{n_Bx_5^2}{x_3^2 n_A}$, 
we will obtain 
$$
Q(\{P'_1\cup P'_6, P'_3\cup P'_7\}) = 
SSC(P'_1\cup P'_6)+SSC(P'_3\cup P'_7)\le SSC(C_A)
$$ $$\le SSC(C_A)+SSC(C_B)=Q(\{C_A,C_B\})$$
that is that for suitably large $b$   it pays off to split $C_A$ and merge $C_B$ 
into parts of $C_A$, because the optimum lies at other partition than the one of well-separatedness in terms of big distance between centres of cluster enclosing balls. 
See also the discussion in Section  \ref{sec:exper}   
  on the table 
\ref{tab:fixdistM}.

\section{Our basic approach to  clusterability}\label{sec:wellclusteredsep}
Let us stress at this point that the issue of well-clusterability is not only a theoretical issue, but it is of practical interest too. 
For example when we intend to  create synthetic data sets for investigating suitability of various clustering algorithms.
But also after having performed the clustering process with whatever method we have, we need to  answer one important question: whether or not the obtained clustering meets the expectation of the analyst. 

These expectations may be divided into several categories:

\begin{itemize}
\item matching business goals,
\item matching underlying algorithm assumptions,
\item proximity to the optimal solutions. 
\end{itemize}
 
 Business goals of the clustering may be difficult to express in terms of data for an algorithm, or may not fit the algorithm domain or data may be too expensive to collect prior to performing an approximate clustering. 
 
 For example, when one seeks a clustering that would enable efficient collection of cars to be scrapped (disassembly  network), then one has to match multiple goals, like covering the whole country, maximum distance from client to  
the disassembly station, and of course the number of prospective clients, which is known with some degree of uncertainty. The distances to the clients are frequently not Euclidean in nature (due to geographical obstacles like rivers mountains etc.), while the preferred $k$-means  algorithm works best with geometrical distances, no upper distance can be imposed etc. Other algorithms may induce same or different problems. So a posteriori one has to check if the obtained solution meets all criteria, does not violate constraints and is stable under fluctuation of the actual set of clients. 

The other two problems are somehow related to one another.
For example, you may have clustered the data being a subsample of the proper data set and the question may be raised how close the sub-sample cluster centres are to the cluster centres of the proper data set. Known methods allow to estimate this discrepancy given that we know that the cluster sizes do not differ too much. So prior to evaluating the correctness of cluster centre estimation we have to check if cluster proportions are within a required range (or if sub-sample size is relevant for such a verification). 
As another example consider 
  methods of estimating closeness to optimal clustering solution under some general data distributions (like for the  $k$-means++\cite{CLU:AV07}), but the guarantees are quite loose. But at the same time the guarantees can be much tighter if the clusters are well-separated in some sense. So if we want to be sure with a reasonable probability that the obtained solution is sufficiently close to the optimum, we would need to check if the obtained clusters are well separated in the defined sense. 

With this in mind, as mentioned, a number of researchers developed the concept of data clusterability.  
The notion of clusterability should intuitively reflect the following idea: if it is easy to see that there are clear-cut clusters in the data, then one would say that the data set is clusterable. "Easy to see" may mean either a visual inspection or some algorithm that quickly identifies the clusters. 
The well-established notion of clusterability would improve our understanding of the concept of the cluster itself - a well-defined clustering would be a clustering of clusterable points. This also would be a foundation for objective evaluation of clustering algorithms.
The algorithm shall perform well for well-clusterable data and  when the clusterability condition would be violated to some degree, the performance of a clustering algorithm is allowed to  deteriorate also, but the algorithm quality would be measured on how the clusterability violation impacts the deterioration of algorithm performance. 

However, the issue turns out not to be that simple. 
As is well known, each algorithm seeking to discover a clustering may be betrayed somehow to fail to discover a clustering structure that is visible upon human inspection of data. 
So instead of trying to reflect human vision of clusterability of the data set independently of the algorithm, let us rather concentrate on finding a concept of clusterability that is both reflecting human perception and the minimum of cost function of a concrete algorithm, in our case $k$-means. We will particularly concentrate on its version called $k$-means++. 

So let us define:

\begin{definition}\label{def:WellClus}
A data set is \emph{well-clusterable} with respect to $k$-means if 
(a) the data points may be split into subsets that are clearly separated by an appropriately chosen gap such that (b) the global minimum of $k$-means cost function coincides with this split and (c) with high probability (over 0.95) the $k$-means++ algorithm discovers this split and (d) if the split was found, it may be verified that the data subsets are separated by the abovementioned gap and (e) if the $k$-means++ did not discover a split of the data fulfilling the requirement of the existence of the gap, then with high probability the split described by points (a) and (b) does not exist. 
\end{definition}

In the paper \cite{MAK:2017:kleinbergaxioms} we have investigated conditions under which one can ensure that the minimum of $k$-means cost function 
is related to a clustering with (wide) gaps between clusters. 

The conditions for clusterable data set therein are rather rigid, but serve the purpose of demonstration that it is possible to define properties of the data set that ensure this property of the minimum of $k$-means. Let us recall below the main result in this respect. 

So assume that the data set encompassing $n$ data points consists of $k$ subsets such that each subset $i=1,\dots,k$ can be enclosed  in a ball of radius $r_i$. Let the gap (distance between surfaces of enclosing balls) between each pair of subsets amount to at least $g$, that is described below.
 
\begin{equation}\label{eq:globalgcase2} 
g \ge r_{max}  \sqrt{k\frac{M+n}{   m} }
\end{equation}
and
\begin{equation}\label{eq:globalgcase1}
g\ge 
k r_{max}   \sqrt{n_p/2+n_q/2+n/2} \sqrt{
\frac{2 n}{    n_{p}n_{q} }
} 
\end{equation}
for any $p,q=1,\dots,k; p\ne q$,
when $n_i, i=1,\dots,k$ is the cardinality of the cluster $i$, 
$M=\max_i n_i$, $m=\min_i n_i$, 

Please note that the quotient of the cardinality of the largest to the smallest cluster increases the size of the required gap, as may be expected from Section \ref{sec:gapinsufficiency}.
From formula (\ref{eq:globalgcase2}) we see that both the relationship $M/m$
and $n/m$ matter. This formula gives the impression that this relationship may be like square root of the sum of the two. 
But note that $g$ is controlled also by formula (\ref{eq:globalgcase1}), where 
the dependence of $g$ on  $n/m$  may become close to linear, while that on $M/m$ will still be close to square root.  
As visible from Section \ref{sec:gapinsufficiency}, the sum of squared distances to cluster centre within the cluster and between clusters decides on the point when the shift in minimal costs occurs when the disproportion between cluster sizes grows. Hence $g$ needs to grow as square root with this disproportion $M/m$. 
The impact of $n/m$ shall be rather viewed in the context of the number of clusters $k$, as with fixed $m$ and growing $n$ $n/m$ may be deemed as a reflection of $k$. 
If one looks at formula (\ref{eq:globalgcase1eqsr}), one sees that $g$ depends approximately quadratically on $k$. 
This relates probably  to the fact  that the number of possible misassignments between clusters grows quadratically with $k$.

It is claimed in \cite{MAK:2017:kleinbergaxioms} that  the optimum of $k$-means objective is reached when splitting the data into the aforementioned subsets.

What are the implications?
The most fundamental one is that the problem is decidable. 

\begin{theorem}\label{th:clusterabilitydecidable}
(i) If the data set is well-clusterable with a gap defined by formulas (\ref{eq:globalgcase2}) and (\ref{eq:globalgcase1}), then with high probability $k$-means++ (after an appropriately chosen number of repetitions)  will discover the respective clustering. 
(ii) If  $k$-means++ (after an appropriately chosen number of repetitions) does not discover  a clustering matching formulas (\ref{eq:globalgcase2}) and (\ref{eq:globalgcase1}), then with high probability the data set is not well clusterable  with a gap defined by formulas (\ref{eq:globalgcase2}) and (\ref{eq:globalgcase1}). 
\end{theorem} 

The rest of the current section is devoted to the proof of the claims of this new theorem, proposed in the current paper. 

If we obtained the split, then for each cluster we are able to compute the cluster centre, the radius of the ball containing all the data points of the cluster, and finally we can check if the gaps between the clusters meet the requirement of formulas (\ref{eq:globalgcase2}) and (\ref{eq:globalgcase1}). So we are able to decide that we have found that the data set is well-clusterable. 

%%%%% FIG 

\newcommand{\CMDFIGgfun}{\begin{minipage}{.48\textwidth} 
%\begin{figure}
\centering
\includegraphics[width=0.8\textwidth]{\figaddr{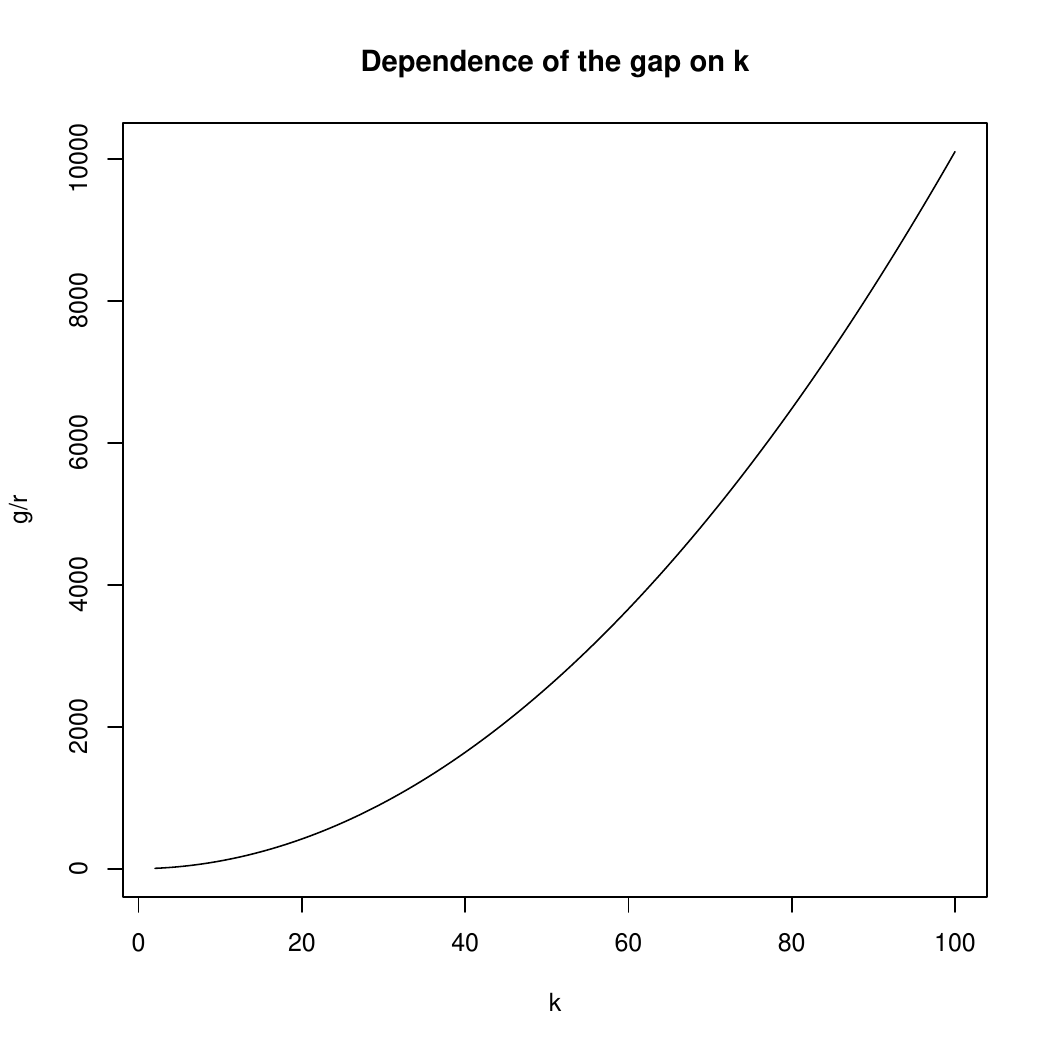}}  %
\caption{Dependence of the gap $g$ on $k$ 
for clusters of equal radius and equal cardinalities. 
}\label{fig:g_fun}
%\end{figure}
\end{minipage} 
}

\newcommand{\CMDFIGpfun}{\begin{minipage}{.48\textwidth} 
%\begin{figure}
\centering
\includegraphics[width=0.8\textwidth]{\figaddr{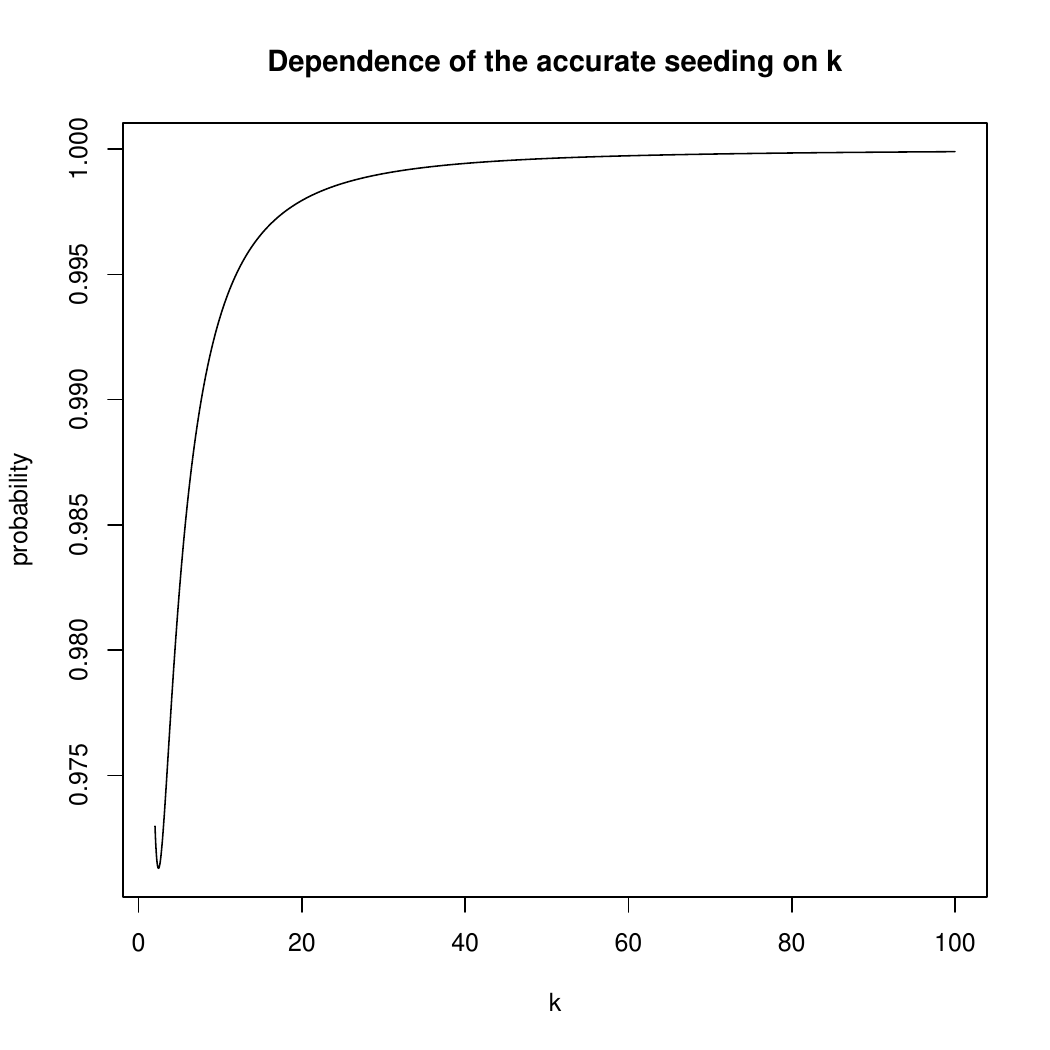}}  %
\caption{Probability of seeding each cluster    on $k$ 
for clusters of equal radius and equal cardinalities. 
}\label{fig:p_fun}
%\end{figure}
\end{minipage} 
}

\begin{figure}
\centering
\CMDFIGgfun
\quad
\CMDFIGpfun
\end{figure}

So let us look at the claim (i). As we already know, the global minimum of $k$-means coincides with the separation by abovementioned gaps. Hence if there exists a positive probability, that $k$-means++ discovers  the appropriate split, then by repeating independent runs of $k$-means++ and picking the split minimising $k$-means cost function we will increase the probability of finding the global minimum. We will show that we know the number of repetitions needed in advance, if we assume the maximum value of the quotient $M/m$.

First consider the easiest case of all clusters being of equal sizes ($M=m$). Then the above equations
(\ref{eq:globalgcase2}) and (\ref{eq:globalgcase1})
 can be reduced to ($r=r_{max}$) 
\begin{equation}\label{eq:globalgcase2eqsr} 
g \ge r \sqrt{k(k+1)  }
\end{equation}
\begin{equation}\label{eq:globalgcase1eqsr}
g\ge  
r k\sqrt{2k +k^2}    
 \end{equation}

%which is satisfied if $g\ge r k (k+1) $
A diagram of dependence of $g/r$ on $k$ is depicted in Figure \ref{fig:g_fun}

Now let us turn to $k$-means++ seeding.
If already $i$ distinct clusters were seeded, 
then the probability that a new cluster will be seeded (under our assumptions)
amounts to at least  
$$\frac{(k-i) g^2}{(k-i) g^2+i r^2}
\ge 
\frac{(k-i) r^2 k^2 (k+1)^2}{(k-i) r^2 k^2 (k+1)^2+i r^2}
$$ $$= 
\frac{(k-i) k^2 (k+1)^2}{(k-i)   k^2 (k+1)^2+i  }
\ge 
\frac{ k^2 (k+1)^2}{   k^2 (k+1)^2+(k-1)  }
$$

%%%% FIG 

Hence the probability of accurate  seeding ($PAS(k)$) amounts to 
$$PAS(k)\ge \left(\frac{ k^2 (k+1)^2}{   k^2 (k+1)^2+(k-1)  }\right)^{k-1}$$
The diagram of dependence of this expression on $k$ is depicted in Figure \ref{fig:p_fun}. 

Let us denote with $Pr_{succ}$ the required probability of success in finding the global minimum. 
To ensure that the seeding was successful in   $Pr_{succ}$  (e.g. 95\% ) of cases, we need to rerun $k$-means++ at least $R$ times, with $R$ given by  
$$
(1-PAS(k))^R\le 
\left(1-\left(\frac{ k^2 (k+1)^2}{   k^2 (k+1)^2+(k-1)  }\right)^{k-1}\right)^R <1-   Pr_{succ}$$

$$R\ge \frac{\log (1- Pr_{succ}  )}
{\log\left(
 1-\left(\frac{ k^2 (k+1)^2}{   k^2 (k+1)^2+(k-1)  }\right)^{k-1}\right) } $$

But look at the following relationship:
$$\left(\frac{ k^2 (k+1)^2}{   k^2 (k+1)^2+(k-1)  }\right)^{k-1}$$
$$=\left(1-\frac{  k-1 }{   k^2 (k+1)^2+(k-1)  }\right)^{k-1}$$
$$=\left(1-\frac{  (k-1)^2 }{   k^2 (k+1)^2+(k-1)  }\frac{1}{k-1}\right)^{k-1}$$
$$\approx e^{-\left( \frac{  (k-1)^2 }{   k^2 (k+1)^2+(k-1)  }\right) } $$

The exponent of the last expression approaches rapidly zero, 
so that with increasing $k$ within a single pass of $k$-means++ the optimum is reached.
In fact, already for k=2 we have an error of below 3\%, 
for k=8, below 1\%, for k=30 below 0.1\%. 
See the Figure \ref{fig:p_fun} for illustration.

\Bem{
k=5 
log(0.05)/log(1-(1-(k-1)/(k^2\cdot (k+1)^2+k-1))^(k-1))

}%Bem

%
%%% FIG

\newcommand{\CMDFIGgmMfun}{\begin{minipage}{.48\textwidth} 
%\begin{figure}
\centering
\includegraphics[width=0.8\textwidth]{\figaddr{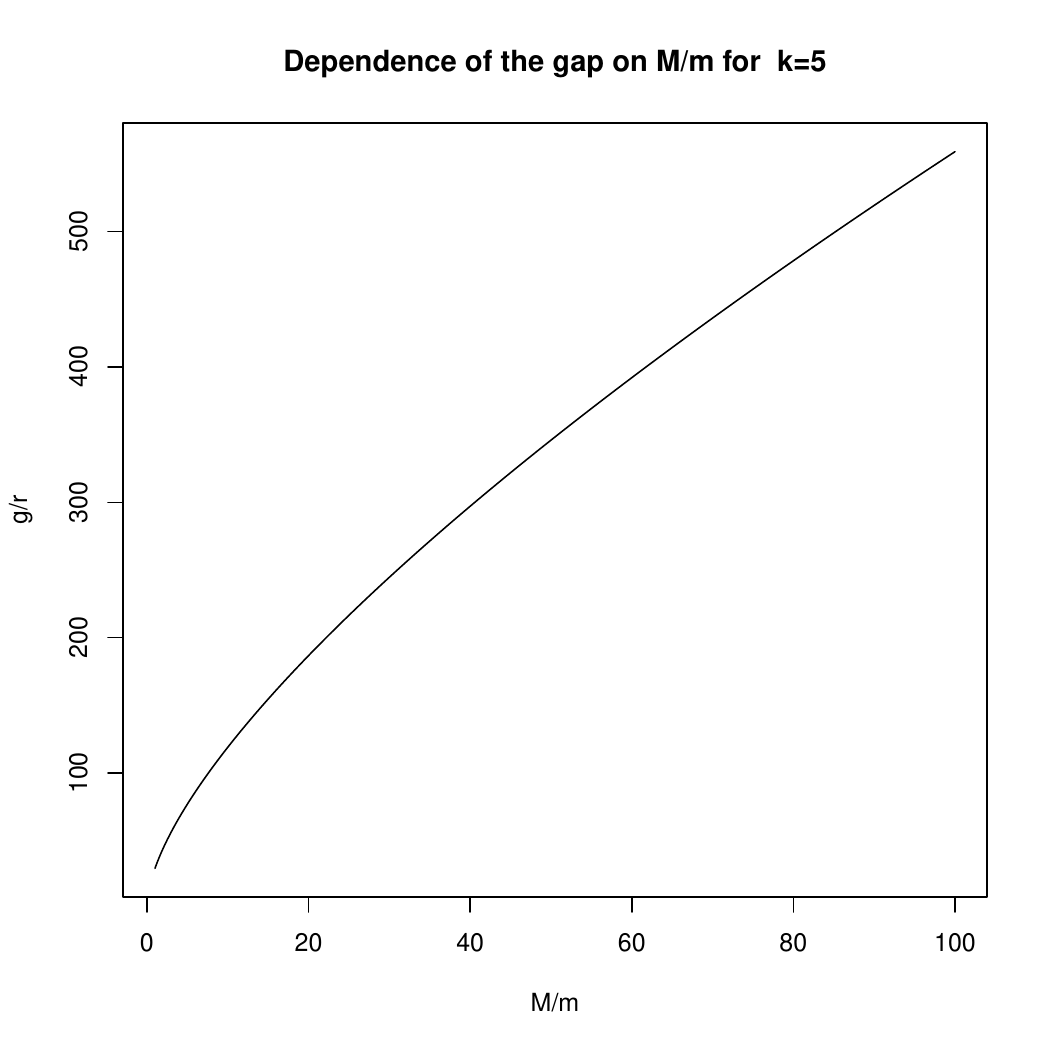}}  %
\caption{Dependence of the gap $g$ for $k=5$ 
for clusters of equal radius when varying cluster cardinalities. 
}\label{fig:gmM_fun}
%\end{figure}
\end{minipage} 
}

\newcommand{\CMDFIGpmMfun}{\begin{minipage}{.48\textwidth} 
%\begin{figure}
\centering
\includegraphics[width=0.8\textwidth]{\figaddr{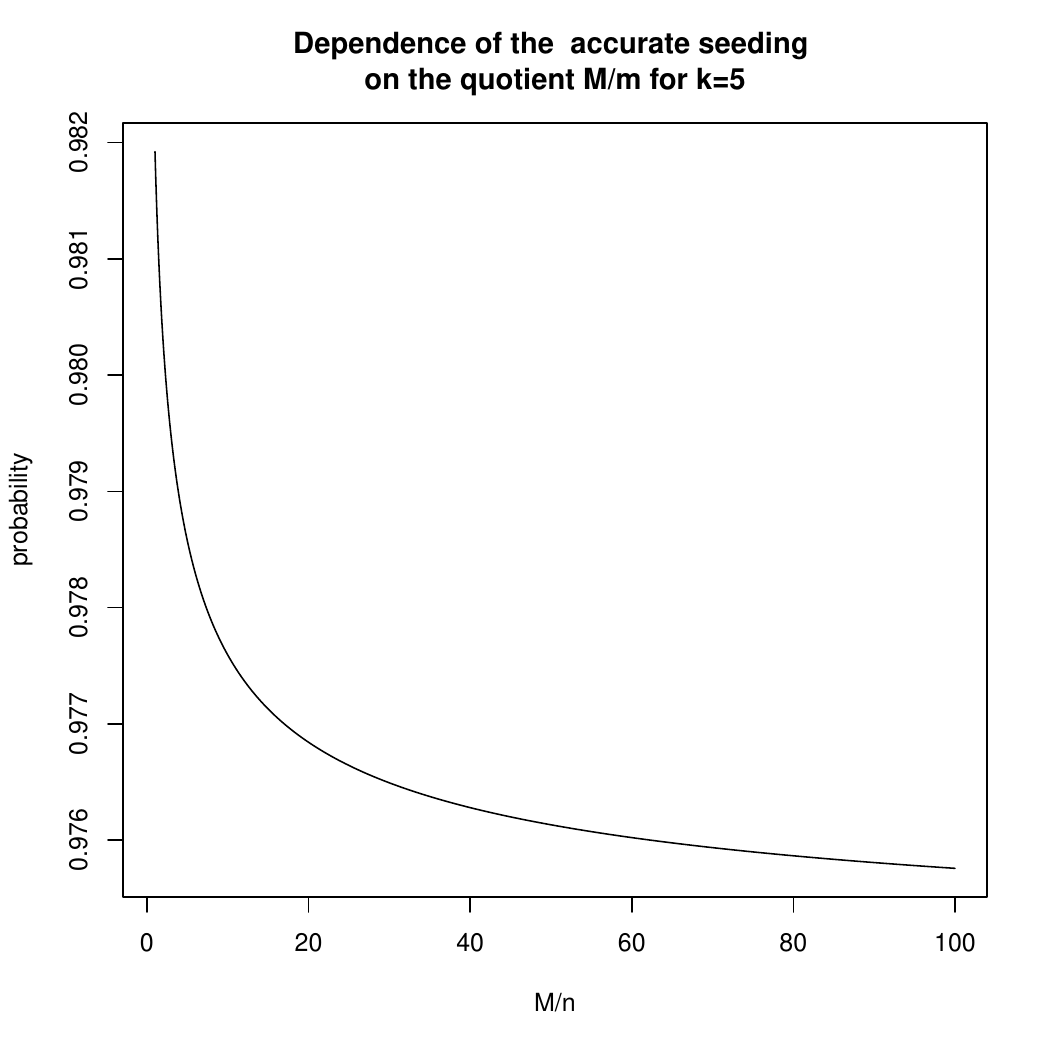}}  %
\caption{
Dependence of the accurate seeding
on the quotient $M/m$ for $k=5$
}\label{fig:pmM_fun}
%\end{figure}
\end{minipage} 
}

\begin{figure}
\CMDFIGgmMfun
\centering
\quad
\CMDFIGpmMfun
\end{figure}

Let us discuss clusters with same radius, but different cardinalities. 
Let $m$ be the cluster minimum cardinality, and $M$ respectively the maximum. 
 
\begin{equation}\label{eq:globalgcase2xx} 
g \ge r \sqrt{k\frac{M+n}{   m} }
\end{equation}
\begin{equation}\label{eq:globalgcase1xx}
g\ge 
k r \sqrt{
\frac{  n (n_p +n_q +n )}{    n_{p}n_{q} }
} 
\end{equation}
for any $p,q=1,\dots,k; p\ne q$,
when $n_i, i=1,\dots,k$ is the cardinality of the cluster $i$, 
$M=\max_i n_i$, $m=\min_i n_i$, 
Worst case $g/r$ values are illustrated in Figure \ref{fig:gmM_fun}.

Now let us turn to $k$-means++ seeding.
If already $i$ distinct clusters were seeded, 
then the probability that a new cluster will be seeded (under our assumptions)
amounts to at least  
$$\frac{(k-i) m g^2}{(k-i) m g^2+i M r^2}
\ge 
\frac{(k-i) m k^2   n (1/m  +1/m +n/m^2 ) }
{(k-i) m k^2   n (1/m  +1/m +n/m^2 ) +i M }
$$ $$= 
\frac{(k-i)   k^2   n (2 +n/m  ) }
{(k-i)   k^2   n (2 +n/m  ) +i M }
\ge
\frac{    k^2   n (2 +n/m  ) }
{    k^2   n (2 +n/m  ) +(k-1) M }
$$
So again the probability of successful seeding will amount to at least:
$$\left(
\frac{    k^2   n (2 +n/m  ) }
{    k^2   n (2 +n/m  ) +(k-1) M }
\right)^{k-1}
$$
$$=\left(
1-
\frac{   (k-1) M  }
{    k^2   n (2 +n/m  ) +(k-1) M }
\right)^{k-1}
$$
$$=\left(
1-
\frac{   (k-1)^2 M  }
{    k^2   n (2 +n/m  ) +(k-1) M }\frac{1}{k-1}
\right)^{k-1}
$$
$$\approx  
 exp \left(-
\frac{   (k-1)^2 M  }
{    k^2   n (2 +n/m  ) +(k-1) M }
\right)
$$
\Bem{
n=1000;
k=2;
Mdom=20;
a=n/k;
M=a*sqrt(Mdom);
m=a/sqrt(Mdom);

exp(-((k-1)^2 *M )/(    k^2 *  n *(2 +n/m  ) +(k-1)* M))

}%Bem

Even if $M$ is 20 times as big as $m$, still the convergence to 1 is so rapid that already for $k=2$ 
the clustering success is achieved with $95\%$ success probability in a single repetition. 
An illustration is visible in Figure \ref{fig:pmM_fun}

%%% FIG 

So far we have concentrated on showing that if the data is well-clusterable, then within practically a single clustering run the seeding will have the property that each cluster obtains a single seed. But what about the rest of the run of $k$-means? As in all these cases $g \ge 2r$, then, as shown in \cite{MAK:2017:kleinbergaxioms}, the cluster centres will never switch to balls encompassing other clusters, so that eventually the true cluster structure is detected and minimum of $Q$ is reached. 
This would complete the proof of claim (i).
The demonstration of claim (ii) is straight forward. 
Note that if a clustering discovered by $k$-means fulfils the conditions of well-clusterability, 
then the data set is clusterable for sure, by definition. 
If the data were not well-clusterable then $k$-means++ 
for sure not find a clustering with the property 
of being well-clusterable, because it does not exist. 
If the data were well-clusterable then $k$-means++ would have failed to identify it with probability of at most $1-Pr_{succ}$. 
 
So denote with $W$ the event that the data is well-clusterable. 
Further denote with $D$ the event that the $k$-means++ algorithm states that the data is well-clusterable. 
We are now interested in approximating $P(\lnot W | \lnot D)$,
or more precisely stating that this probability is high.  

$$Pr(\lnot W | \lnot D)
=
\frac{Pr(\lnot D | \lnot W) Pr( \lnot W) }
{Pr(\lnot D | \lnot W) Pr( \lnot W) + Pr(\lnot D |   W) Pr(  W)
}
$$ $$
=
\frac{  Pr( \lnot W) }
{  Pr( \lnot W) + Pr(\lnot D |   W) Pr(  W)
}
=
\frac{1  }
{  1 + Pr(\lnot D |   W) \frac{Pr(  W)}{Pr( \lnot W)}
}
$$ $$
\ge 
\frac{1  }
{  1 + (1-Pr_{succ}) \frac{Pr(  W)}{Pr( \lnot W)}
} 
%$$ $$
\ge 
   1 - (1-Pr_{succ}) \frac{Pr(  W)}{Pr( \lnot W)}
\ge Pr_{succ}
$$
The last inequality is true because   the well-clusterable data are in practice extremely rare,
and for sure less frequent than not well-clusterable ones.

Please note that what we have discussed here is a kind of worst case analysis. 
Already from this discussion it is obvious that the probability of seeding of $k$ distinct clusters depends on the characteristics of the data.
We refer always to the smallest gaps between clusters, but it may turn out that some clusters are stronger separated. This will automatically increase their probability of being hit so that the overall probability of hitting unhit   clusters will increase significantly.

\section{Smaller gaps between clusters}\label{sec:core}.
In the previous section we considered well-clusterability under the assumption of large areas between clusters where no data points of any cluster will occur. Subsequently we show that this assumption may be relaxed so that spurious points are allowed between the major concentrations of cluster points. 
But to ensure that the presence of such points will not lead the $k$-means procedure astray, we will distinguish core parts of the clusters and will ensure by the subsequent Theorem \ref{th:clusterpreservationbycorepreservation} that once a cluster core is hit by $k$-means initialisation procedure, the cluster is preserved over subsequent $k$-means iterations.

In \cite{MAK:2017:kleinbergaxioms} we have proven that 
\begin{theorem}{}\label{th:clusterpreservationbygap}
Let $A,B$ be cluster centres. 
Let $\rho_{AB}$ be the radius of a ball centred 
at $A$ and enclosing its cluster 
and it also is the radius of a ball centred 
at $B$ and enclosing its cluster.  
If the distance between   the cluster centres $A,B$
amounts to $2\rho_{AB}+g$, $g>0$ ($g$ being the "gap" between clusters),  
if we pick any two points, $X$ from the cluster of $A$ and $Y$ from the cluster of $B$, 
and recluster both clusters around $X$ and $Y$, 
then the new clusters will preserve the balls centred at $A$ and $B$ of radius $g/2$ (called subsequently "cores") each ($X$ the core of $A$, $Y$ the core of $B$).  
\end{theorem}  

Here we shall demonstrate the validity of a complementary theorem. 

\begin{theorem}{}\label{th:clusterpreservationbycorepreservation}
Let $A,B$ be cluster centres. 
Let $\rho_{AB}$ be the radius of a ball centred 
at $A$ and enclosing its cluster 
and it also is the radius of a ball centred 
at $B$ and enclosing its cluster.  
Let $\rho_{cAB}$ be the radius of a ball centred 
at $A$ and enclosing "vast majority" of its cluster 
and it also is the radius of a ball centred 
at $B$ and enclosing  "vast majority" of its cluster.  
If the distance between   the cluster centres $A,B$
amounts to $2\rho_{AB}+g$, $g>0$ ($g=2 r_{cAB}$ being the "gap" between clusters),  
if we pick any two points, $X$ from the ball $B(A,r_{cAB})$ and $Y$ from the ball $B(A,r_{cAB})$, 
and recluster both clusters around $X$ and $Y$, 
then the new clusters will 
be identical to the original clusters around $A$ and $B$.   
\end{theorem}

\begin{definition}
If the gap between each pair of clusters fulfils the condition of either of the above two theorems, then we say that we have core-clustering. 
\end{definition}

\begin{proof}
For the illustration of the proof see Figure \ref{fig:threecoreballs}.

\begin{figure}
\centering
\includegraphics[width=0.6\textwidth]{\figaddr{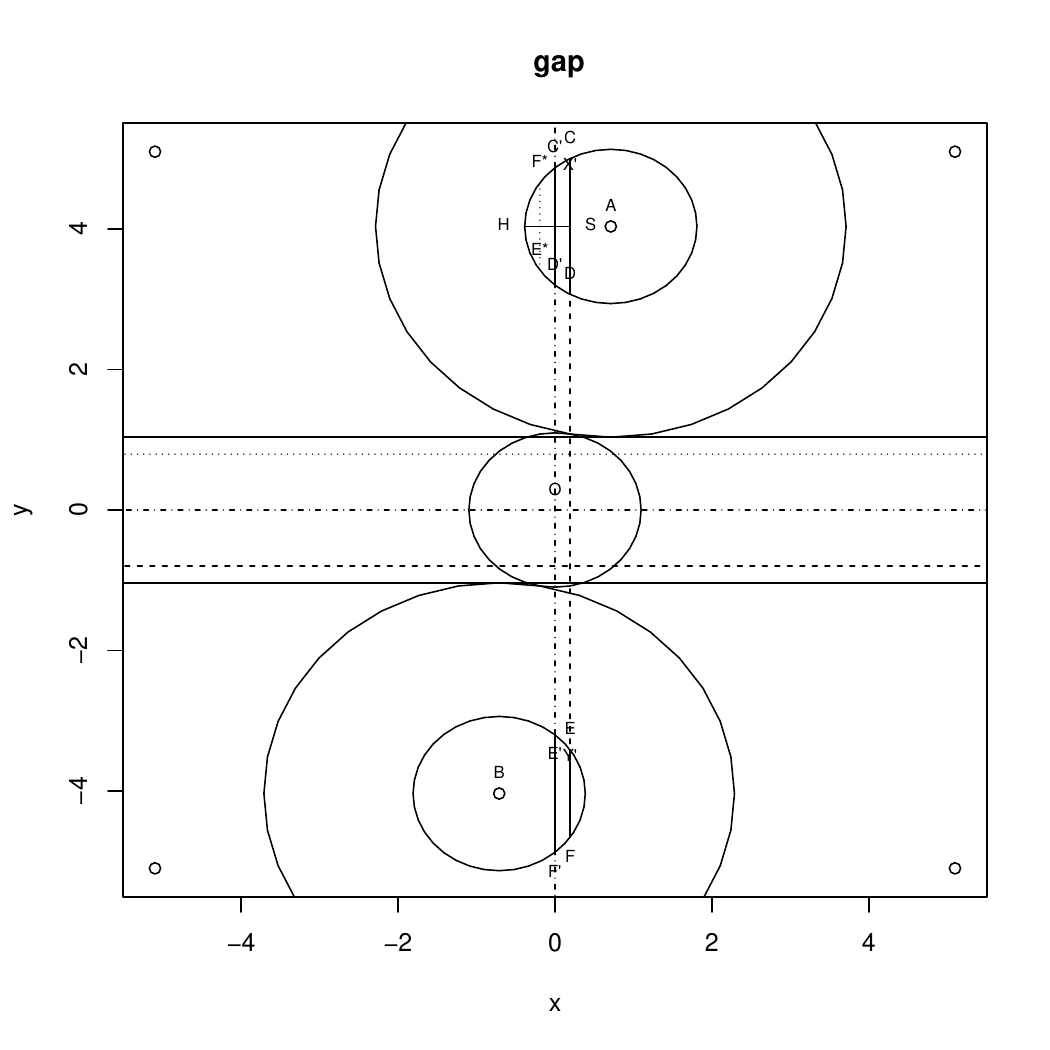}}  %
\caption{An illustrative figure for the proof of the cluster preservation under a gap between cluster enclosing balls. 
The figure represents a projection of two clusters centred at $A$ and $B$ onto a  plane containing the line $AB$. The inner circles centred at $A$ and $B$ contain the projections of cores of the respective clusters and outer circles contain the projections of whole clusters. The gap between clusters is represented by the circle with centre at $O$. It is proven that if the cluster centres change their positions  (via the initialisation procedure of $k$-means, for example) to any other place within the clusters $A$ and $B$, then the core of $A$ will remain in $A$, and the core of $B$ will remain in $B$, in spite of the fact that other elements of the clusters may change cluster membership.
Notation is explained in detail in the text. 
}\label{fig:threecoreballs}
\end{figure}

The proof does not differ too much from the previous one and in fact the previous Theorem \ref{th:clusterpreservationbygap} is  a special case of
Theorem \ref{th:clusterpreservationbycorepreservation}.

Consider the two points $A,B$ being the two centres of double balls.
The inner call represents the core of radius $r_{cAB}=g/2$, the outer ball of radius $\rho$ ($\rho=\rho_{AB}$), enclosing the whole cluster. 
Consider  two points, $X,Y$, one being in each core ball  (presumably the cluster centres at some stage of the $k$-means algorithm). To represent their distances faithfully, we need at most a 3D space. 

Let us consider the plane established by the line $AB$ and parallel to the line $XY$. 
Let $X'$ and $Y'$ be orthogonal projections of $X,Y$ onto this plane. 
Now let us establish that the hyperplane $\pi$ orthogonal to $XY$, and passing through the middle of the line segment $XY$, that is the hyperplane containing the boundary between clusters centred at $X$ and $Y$ does not cut any of the balls centred at $A$ and $B$. 
This hyperplane will be orthogonal to the plane of the Figure  \ref{fig:threecoreballs} and so it will manifest itself as an intersecting line $l$ that should not cross outer circles around $A$ and $B$, being projections of the respective balls.  Let us draw two solid lines $k,m$ between circles $O(A,\rho_{AB})$ and $O(B,\rho_{AB})$ tangential to each of them. Line $l$ should lie between these lines, in which case the cluster centre will not jump to the other ball. 

Let the line $X'Y'$ intersect with the circles $O(A,r_{cAB})$ and $O(B,r_{cAB})$ at points $C,D,E,F$ as in the figure. 

It is obvious that the line $l$ would get closer to circle $A$, if the points $X', Y'$ would lie closer to $C$ and $E$, or closer to circle $B$ if they would be closer to $D$ and $F$. 

Therefore, to show that it does not cut the circle $O(A,\rho)$ it is sufficient to consider $X'=C$ and $Y'=E$. (The case with ball $Ball(B,\rho)$ is symmetrical).

Let $O$ be the centre of the line segment $AB$. Let us draw through this point a line parallel to $CE$ that cuts the circles at points $C', D', E'$ and $F'$. 
Now notice that centric symmetry through point $O$ transforms the circles $O(A,r_{cAB})$,$O(B,r_{cAB})$ into one another, and point $C'$ in $F'$ and $D'$ in $E'$. Let $E^*$ and $F^*$ be images of points $E$ and $F$ under this symmetry. 

In order for the line $l$ to lie between $m$ and $k$, the middle point of the line segment $CE$ shall lie between these lines.
 
Let us introduce a planar coordinate system centred at $O$ with $\mathcal{X}$ axis parallel to lines $m,k$, such that $A$ has both coordinates non-negative, and $B$ non-positive. 
Let us denote with  $\alpha$ the angle between the lines $AB$ and $k$. 
As we assume that the distance between $A$ and $B$ equals $2\rho+2r_{cAB}$, then the distance between lines $k$ and $m$ amounts to $ 
2((\rho+r_{cAB})\sin(\alpha)-\rho)$. 
Hence the  $\mathcal{Y}$ coordinate of line $k$ equals $((\rho+r_{cAB})\sin(\alpha)-\rho)$.

So the  $\mathcal{Y}$ coordinate of the centre of line segment $CE$ shall be not higher than this. 
Let us express this in the coordinate system:
$$4(y_{OC}+y_{OE})/2 \le ((\rho+r_{cAB})\sin(\alpha)-\rho)$$
\noindent Where $y_{OC}$ is the $y$-coordinate of the vector $\overrightarrow{OC}$, etc..  

Note, however that 
$$y_{OC}+y_{OE}=y_{OA}+y_{AC}+y_{OB}+y_{BE}= y_{AC}+y_{BE} =  y_{AC}-y_{AE^*} 
=y_{AC}+y_{E^*A}$$ 

So let us examine the circle with centre at A. 
Note that the lines $CD$ and $E^*F^*$ are at the same distance from  the line $C'D'$. Note also that the absolute values  of direction coefficients of tangentials of circle $A$ at $C'$ and $D'$ are identical.
The more distant these lines are, as line $CD$ gets closer to $A$, the $y_{AC}$ gets bigger, and $y_{E^*A}$ becomes smaller. 
But from the properties of the circle we see that 
$y_{AC}$ increases at a decreasing rate, while  $y_{E^*A}$ decreases at an increasing rate. 
So the sum $y_{AC}+y_{E^*A}$  has the biggest value when $C$ is identical with $C'$ and we need hence to prove only that 
$$(y_{AC'}+y_{D'A} )/2=y_{AC'} \le  ((\rho+r_{cAB})\sin(\alpha)-\rho)$$

Let $M$ denote the middle point of the line segment $C'D'$. 
As point $A$ has the coordinates $((\rho+r_{cAB}) \cos(\alpha), (\rho+r_{cAB}) \sin(\alpha))$,
the point $M$ is at distance of $(\rho+r_{cAB}) \cos(\alpha)$ from $A$. 
But $C'M^2= r_{cAB}^2-((\rho+r_{cAB}) \cos(\alpha))^2 $.  

So we need to show that 
$$r_{cAB}^2-((\rho+r_{cAB}) \cos(\alpha))^2 \le ((\rho+r_{cAB})\sin(\alpha)-\rho)^2$$
In fact we get from the above 
$$r_{cAB}^2-((\rho+r_{cAB}) \cos(\alpha))^2 \le 
((\rho+r_{cAB})\sin(\alpha))^2
+\rho^2
-2(\rho+r_{cAB}) \rho \sin(\alpha)$$
$$r_{cAB}^2  \le  (\rho+r_{cAB})^2+\rho^2
-2(\rho+r_{cAB}) \rho \sin(\alpha)$$

$$r_{cAB}^2-\rho^2  \le  (\rho+r_{cAB})^2
-2(\rho+r_{cAB})\rho\sin(\alpha)$$
$$(r_{cAB} -\rho)(r_{cAB} +\rho)  \le  (\rho+r_{cAB})^2
-2(\rho+r_{cAB})(\rho)\sin(\alpha)$$
$$(r_{cAB} -\rho)  \le  (\rho+r_{cAB}) 
-2 \rho \sin(\alpha)$$
$$0   \le  2\rho  
-2 \rho \sin(\alpha)$$
$$0   \le  1-   \sin(\alpha)$$
which is obviously true, as $\sin$ never exceeds 1. 
%This ends the proof of the theorem.
\end{proof}

\section{Core based global $k$-means minimum }
\label{sec:corebasedglobalkmeansminimum}
In the paper \cite{MAK:2017:kleinbergaxioms} we have investigated conditions under which one can ensure that the minimum of $k$-means cost function 
is related to a clustering with (wide) gaps between clusters. 

Based on the result of the preceding Section \ref{sec:core}, we want to weaken these conditions requiring only that the big gaps exist between cluster cores and the clusters themselves are separated by much smaller gaps, equal to the size of the core. 

In particular, let us consider
the set of 
 $k$ clusters $\overline{\mathcal{C}}=\{\overline{C_1},\dots,\overline{C_k}\}$ of cardinalities $\overline{n_1},\dots,\overline{n_k}$ and with radii of balls enclosing the clusters (with centres located at cluster centres) $\overline{r_1},\dots, \overline{r_k}$.  
Let each of these clusters $\overline{C_i}$ have a core $C_i$
around the cluster  $\overline{C_i}$  centre 
 of 
radius $r_i$ 
and cardinality $n_i$ 
such that for $\mathfrak{p}\in [0,1)$ 
$$Q(\{C_i\})/Q(\{\overline{C_i}\})\ge 1-\mathfrak{p}$$

We are interested in a gap $g$ between cluster cores $C_1,\dots,C_k$ such that it does not make sense to split each cluster $\overline{C_i}$ into subclusters $\overline{C_{i1}},\dots, \overline{C_{ik}}$ and to combine them into
a set of 
 new clusters $\mathcal{S}=\{S_1,\dots,S_k\}$ such that 
$S_j=\cup_{i=1}^k \overline{C_{ij}}$. 

We seek a $g$ such that the highest possible central sum of squares combined over the clusters $\overline{C_i}$ would be lower than the lowest conceivable combined sums of squares around respective centres of clusters $S_j$. 
Let $Var(C)$ be the variance of the cluster $C$ (average squared distance to set $C$ gravity centre; with one exception, however:
if referring to the \emph{ core }  of any of the clusters 
$\overline{C_i}$, 
 we  compute against the cluster $\overline{C_i}$ gravity centre, not the core $C_i$ gravity centre, so also with the $Q$ function). 
Let $C_{ij}=\overline{C_{ij}}\cap C_i$ be the core part of the subcluster $\overline{C_{ij}}$. 
Let $r_{ij}$ be the distance of the centre of core subcluster $C_{ij}$ to the centre of cluster $\overline{C_i}$.
Let $v_{ilj}$ be the distance of the centre of core subcluster $C_{ij}$ to the centre of core subcluster $C_{lj}$.
So the total $k$-means function for the set of clusters $(C_1,\dots,C_k)$ will amount to:
\begin{equation} 
Q(\overline{\mathcal{C}})
=\frac{1}{1-\mathfrak{p}} Q( \mathcal{C} )
=\frac{1}{1-\mathfrak{p}} \sum_{i=1}^k \sum_{j=1}^k (n_{ij}Var(C_{ij})+n_{ij}r_{ij}^2)
\end{equation}
And the total $k$-means function for the set of clusters $(S_1,\dots,S_k)$ will amount to:
\begin{equation}
Q(\mathcal{S})\ge \sum_{j=1}^k \left(\left(\sum_{i=1}^k n_{ij}Var(C_{ij})\right)+
({\sum_{i=1}^k n_{ij}} )
\left(\sum_{i=1}^{k-1} \sum_{l=i+1}^k  \frac{n_{ij}}{\sum_{i=1}^k n_{ij}}\frac{n_{lj}}{\sum_{i=1}^k n_{ij}} v_{ilj}^2
\right)
\right)
\end{equation}

Should $(\overline{C_1},\dots,\overline{C_k})$ constitute the absolute minimum of the $k$-means target function, then $Q(\mathcal{S})\ge Q(\overline{\mathcal{C}})$  should hold, which is fulfilled if :
$$\sum_{j=1}^k \left(\left(\sum_{i=1}^k n_{ij}Var(C_{ij})\right)+
({\sum_{i=1}^k n_{ij}} )
\left(\sum_{i=1}^{k-1} \sum_{l=i+1}^k    \frac{n_{ij}}{\sum_{i=1}^k n_{ij}}\frac{n_{lj}}{\sum_{i=1}^k n_{ij}} v_{ilj}^2
\right)
\right)
%%%%%
$$ $$
\ge 
%%%%%
\frac{1}{1-\mathfrak{p}}\sum_{i=1}^k \sum_{j=1}^k (n_{ij}Var(C_{ij})+n_{ij}r_{ij}^2)
$$

Note that on the left hand-side of the inequality we ignored the portion of the data outside of the cores. this portion of the data would have made the left-hand-side even bigger. 

The above inequality is implied by: 

\begin{equation}\label{eq:e1}
\sum_{j=1}^k   
\left(\sum_{i=1}^{k-1} \sum_{l=i+1}^k   \frac{n_{ij}n_{lj}}{\sum_{i=1}^k n_{ij}}  v_{ilj}^2
\right)
%%%%%
\ge
%%%%%
\frac{1}{1-\mathfrak{p}}\sum_{i=1}^k \sum_{j=1}^k (\mathfrak{p}n_{ij}Var(C_{ij})+n_{ij}r_{ij}^2)
\end{equation}

Note that $Var(C_{ij})\le r_{ij}^2$, so 
\begin{align} 
\frac{1}{1-\mathfrak{p}}\sum_{i=1}^k \sum_{j=1}^k (\mathfrak{p}n_{ij}Var(C_{ij})+n_{ij}r_{ij}^2)
\le & 
\frac{1}{1-\mathfrak{p}}\sum_{i=1}^k \sum_{j=1}^k (1+\mathfrak{p})n_{ij} n_{ij}r_{ij}^2
\nonumber \\ = &
\frac{1+\mathfrak{p}}{1-\mathfrak{p}}\sum_{i=1}^k \sum_{j=1}^k n_{ij} n_{ij}r_{ij}^2
\end{align}

To maximise $\sum_{j=1}^k  n_{ij}r_{ij}^2$ for a single cluster $C_i$ of enclosing ball radius $r_i$, note that you should set $r_{ij}$ to $r_i$. Let $m_j=\arg \max_{j \in \{1,\dots,k\}} n_{ij}$. 
If we set $r_{ij}=r_i$ for all   $j$ except $m_j$, then the maximal $r_{i{m_j}}$ is delimited by the relation
$\sum_{j=1; j\ne m_j}^k  n_{ij}r_{ij}\ge n_{i{m_j}}r_{i{m_j}}$.
So 
\begin{align}\label{eq:e2}
\sum_{j=1}^k  n_{ij}r_{ij}^2 \le & 
 \left(\sum_{j=1; j\ne m_j}^k  n_{ij}\right) r_i^2\min\left(2,\left(1+\frac{\sum_{j=1; j\ne m_j}^k  n_{ij}}{n_{i{m_j}}} \right)\right)
\\ \le &
 2 \left(\sum_{j=1; j\ne m_j}^k  n_{ij}\right) r_i^2 
\nonumber 
\end{align}

So if we can guarantee that the gap between cluster balls (of clusters from $\mathcal{C}$) amounts to $g$  then surely 

\begin{equation}\label{eq:e3}
 \sum_{j=1}^k   
\left(\sum_{i=1}^{k-1} \sum_{l=i+1}^k   \frac{n_{ij}n_{lj}}{\sum_{i=1}^k n_{ij}}  v_{ilj}^2
\right)
\ge 
g^2
\sum_{j=1}^k   
\left(\sum_{i=1}^{k-1} \sum_{l=i+1}^k   \frac{n_{ij}n_{lj}}{\sum_{i=1}^k n_{ij}} 
\right)
\end{equation}
because in such case $g\le v_{ilj}$ for all $i,l,j$.

By combining inequalities (\ref{eq:e1}),  (\ref{eq:e2}) and (\ref{eq:e3}) we see 
that the global minimum is granted if the following holds:
\begin{equation}\label{eq:globalg}
g^2
\sum_{j=1}^k   
\left(\sum_{i=1}^{k-1} \sum_{l=i+1}^k   \frac{n_{ij}n_{lj}}{\sum_{i=1}^k n_{ij}}   
\right)
%%%%%
\ge
%%%%
 2 \frac{1+\mathfrak{p}}{1-\mathfrak{p}}\sum_{i=1}^k \left(\sum_{j=1; j\ne m_j}^k  n_{ij}\right) r_i^2 
\end{equation}

One can distinguish two cases: either 
(1) there exists a cluster $S_t$ containing two subclusters $C_{pt}$, $C_{qt}$ 
such that $t=\arg \max_j |C_{pj}|$ 
and $t=\arg \max_j |C_{qj}|$ 
(maximum cardinality   subclasses of their respective original clusters $C_p, C_q$
 or (2) not. 

Consider the first case. Let $C_p,C_q$ be the two clusters where $C_{pt}$ and $C_{qt}$ be two subclusters of highest cardinality within $C_p,C_q$ resp. 
This implies that $n_{pt}\ge \frac 1k n_p, n_{qt}\ge \frac 1k n_q$. 
Also this implies that for $i\ne p, i\ne q$  $n_{it}\le n_i/2$.

$$
\sum_{j=1}^k   
 \sum_{i=1}^{k-1} \sum_{l=i+1}^k   \frac{n_{ij}n_{lj}}{\sum_{i=1}^k n_{ij}}  
%$$ $$
\ge   
 \sum_{i=1}^{k-1} \sum_{l=i+1}^k   \frac{n_{it}n_{lt}}{\sum_{i=1}^k n_{it}}  
%$$ $$
\ge 
    \frac{n_{pt}n_{qt}}{\sum_{i=1}^k n_{it}}  
$$
$$
\ge 
  \frac{n_{pt}n_{qt}} {n_p/2+n_q/2+\sum_{i=1}^k n_{i}/2 }  
=   \frac{n_{pt}n_{qt}}{n_p/2+n_q/2+n/2}   
$$

$$
\ge \frac1{k^2}  \frac{n_{p}n_{q}}{n_p/2+n_q/2+n/2}   
$$

Note that 
$$
2 \sum_{i=1}^k \left(\sum_{j=1; j\ne m_j}^k  n_{ij}\right) r_i^2 
\le 2 \sum_{i=1}^k  n_{i} r_i^2 
$$
So, in order to fulfil inequality (\ref{eq:globalg}), it is sufficient to require that  
\begin{align}
g\ge & 
\sqrt{
\frac{2\frac{1+\mathfrak{p}}{1-\mathfrak{p}} \sum_{i=1}^k  n_{i} r_i^2 }
{ \frac1{k^2}  \frac{n_{p}n_{q}}{n_p/2+n_q/2+n/2} } 
}
\nonumber \\ = & 
k\sqrt{n_p/2+n_q/2+n/2} \sqrt{
\frac{2 \frac{1+\mathfrak{p}}{1-\mathfrak{p}}\sum_{i=1}^k  n_{i} r_i^2 }{    n_{p}n_{q} }
}
\nonumber \\ =  & \label{eq:globalgcase1core}
k\sqrt{n_p +n_q +n } \sqrt{
\frac{ \frac{1+\mathfrak{p}}{1-\mathfrak{p}} \sum_{i=1}^k  n_{i} r_i^2 }{    n_{p}n_{q} }
}
\end{align}
This of course maximized over all combinations of $p,q$. 

Let us proceed to the second case.
Here each cluster $S_j$ contains a subcluster of maximum cardinality of a different cluster $C_i$. 
As the relation between $S_j$ and $C_i$ is unique, we can reindex $S_j$ in such a way that actually $C_j$ contains its maximum cardinality subcluster $C_{jj}$. 
Let us rewrite the  inequality (\ref{eq:globalg}). 

$$
g^2
\sum_{j=1}^k   
\left(\sum_{i=1}^{k-1} \sum_{l=i+1}^k   \frac{n_{ij}n_{lj}}{\sum_{i=1}^k n_{ij}}   
\right)
-
 2 \frac{1+\mathfrak{p}}{1-\mathfrak{p}}\sum_{i=1}^k \left(\sum_{j=1; j\ne m_j}^k  n_{ij}\right) r_i^2 
%%%%%
\ge 0
%%%%
$$

This is met if 

$$
g^2
\sum_{j=1}^k   
\left(\sum_{i=1}^{j-1}  \frac{n_{ij}n_{jj}}{\sum_{i=1}^k n_{ij}} 
+
\sum_{l=j+1}^k   \frac{n_{jj}n_{lj}}{\sum_{i=1}^k n_{ij}}   
\right)
-
 2 \frac{1+\mathfrak{p}}{1-\mathfrak{p}}\sum_{i=1}^k (n_i-  n_{ii}) r_i^2 
%%%%%
\ge 0
%%%%
$$
This is the same as:

$$
g^2
\sum_{j=1}^k   
\left(\sum_{i=1,\dots, {j-1},{j+1},\dots,k}   \frac{n_{ij}n_{jj}}{\sum_{i=1}^k n_{ij}}    
\right)
-
 2 \frac{1+\mathfrak{p}}{1-\mathfrak{p}}\sum_{i=1}^k (n_i-  n_{ii}) r_i^2 
%%%%%
\ge 0
%%%%
$$

This is fulfilled if:

$$
g^2
\sum_{j=1}^k   
\left(\sum_{i=1,\dots, {j-1},{j+1},\dots,k}   \frac{n_{ij}n_{j}/k}
{n_j/2+\sum_{i=1}^k n_{i}/2}    
\right)
-
 2 \frac{1+\mathfrak{p}}{1-\mathfrak{p}}\sum_{i=1}^k (n_i-  n_{ii}) r_i^2 
%%%%%
\ge 0
%%%%
$$

Let $M$ be the maximum over $n_1,\dots,n_k$. The above holds if 

$$
g^2
\sum_{j=1}^k   
\left(\sum_{i=1,\dots, {j-1},{j+1},\dots,k}   \frac{n_{ij}n_{j}/k}
{M/2+n/2}    
\right)
-
 2 \frac{1+\mathfrak{p}}{1-\mathfrak{p}}\sum_{i=1}^k (n_i-  n_{ii}) r_i^2 
%%%%%
\ge 0
%%%%
$$
Let $m$ be the minimum over $n_1,\dots,n_k$. The above holds if 
$$
g^2
\sum_{j=1}^k   
\left(\sum_{i=1,\dots, {j-1},{j+1},\dots,k}   \frac{n_{ij}m/k}
{M/2+n/2}    
\right)
-
 2 \frac{1+\mathfrak{p}}{1-\mathfrak{p}}\sum_{i=1}^k (n_i-  n_{ii}) r_i^2 
%%%%%
\ge 0
%%%%
$$
This is the same as 
$$
g^2   \frac{m/k}
{M/2+n/2}    
\left(
\sum_{j=1}^k   
\sum_{i=1,\dots, {j-1},{j+1},\dots,k}   {n_{ij} }   
\right)
-
 2 \frac{1+\mathfrak{p}}{1-\mathfrak{p}}\sum_{i=1}^k (n_i-  n_{ii}) r_i^2 
%%%%%
\ge 0
%%%%
$$

$$
g^2   \frac{m/k}
{M/2+n/2}    
\left(
\sum_{j=1}^k
\left(   
\left(   
\sum_{i=1}^k   {n_{ij} } \right) - n_{jj}  
\right)
-
 2 \frac{1+\mathfrak{p}}{1-\mathfrak{p}}\sum_{i=1}^k (n_i-  n_{ii}) r_i^2 \right)
%%%%%
\ge 0
%%%%
$$

$$
g^2   \frac{m/k}
{M/2+n/2}    
\left(   
\left(   
\sum_{j=1}^k\sum_{i=1}^k   {n_{ij} } \right) - (\sum_{j=1}^kn_{jj} ) 
\right)
-
 2 \frac{1+\mathfrak{p}}{1-\mathfrak{p}}\left( \sum_{i=1}^k (n_i-  n_{ii}) r_i^2 \right)
%%%%%
\ge 0
%%%%
$$

$$
g^2   \frac{m/k}{M/2+n/2}    
\left(   
\left(   
\sum_{i=1}^k   {n_{i} } \right) - \left(\sum_{j=1}^kn_{jj} \right) 
\right)
-
 2 \frac{1+\mathfrak{p}}{1-\mathfrak{p}}\sum_{i=1}^k (n_i-  n_{ii}) r_i^2 
%%%%%
\ge 0
%%%%
$$

$$
g^2   \frac{m/k}
{M/2+n/2}    
\left(      
\sum_{i=1}^k  \left( {n_{i} -n_{ii}} \right)  
\right)
-
 2 \frac{1+\mathfrak{p}}{1-\mathfrak{p}}\sum_{i=1}^k (n_i-  n_{ii}) r_i^2 
%%%%%
\ge 0
%%%%
$$

$$
\sum_{i=1}^k  \left( {n_{i} -n_{ii}} \right) \left(
g^2   \frac{m/k}
{M/2+n/2}    
-
 2  \frac{1+\mathfrak{p}}{1-\mathfrak{p}} r_i^2 
\right)
%%%%%
\ge 0
%%%%
$$
The above will hold, if for every $i=1,\dots,k$
$$g \ge r_i \sqrt{\frac{1+\mathfrak{p}}{1-\mathfrak{p}}\frac{2}{  \frac{m/k}{M/2+n/2} }} $$
\begin{equation}\label{eq:globalgcase2core} 
g \ge r_i \sqrt{k\frac{1+\mathfrak{p}}{1-\mathfrak{p}}\frac{M+n}{   m} }
\end{equation}

So the  inequality (\ref{eq:globalg}) is fulfilled, if both 
 inequality (\ref{eq:globalgcase1core}) and  inequality (\ref{eq:globalgcase2core}) are held by an appropriately chosen $g$. 

In summary we have shown that 
\begin{theorem}
\label{th:coreclusterglobaloptimum}
Let $\overline{\mathcal{C}}=\{\overline{C_1},\dots,\overline{C_k}\}$
be a partition of a data set into  
 $k$ clusters of cardinalities $\overline{n_1},\dots,\overline{n_k}$ and with radii of balls enclosing the clusters (with centres located at cluster centres) $\overline{r_1},\dots, \overline{r_k}$.  
Let each of these clusters $\overline{C_i}$ have a core $C_i$ of radius $r_i$ 
and cardinality $n_i$ around the cluster centre 
such that for $p\in [0,1)$ 
$$Q(\{C_i\})/Q(\{\overline{C_i}\})\ge 1-\mathfrak{p}$$
Then if the   gap $g$ between cluster cores $C_1,\dots,C_k$ 
fulfils conditions 
expressed in formulas 
(\ref{eq:globalgcase1core}) and   (\ref{eq:globalgcase2core})
then the partition 
$\overline{\mathcal{C}}$ coincides with the global minimum 
of the $k$-means cost function for the data set. 
\end{theorem}

%-----------------------------------------------

\newcommand{\CMDFIGgmMpfun}{\begin{minipage}{.60\textwidth} 
%\begin{figure}
\centering
\includegraphics[width=0.8\textwidth]{\figaddr{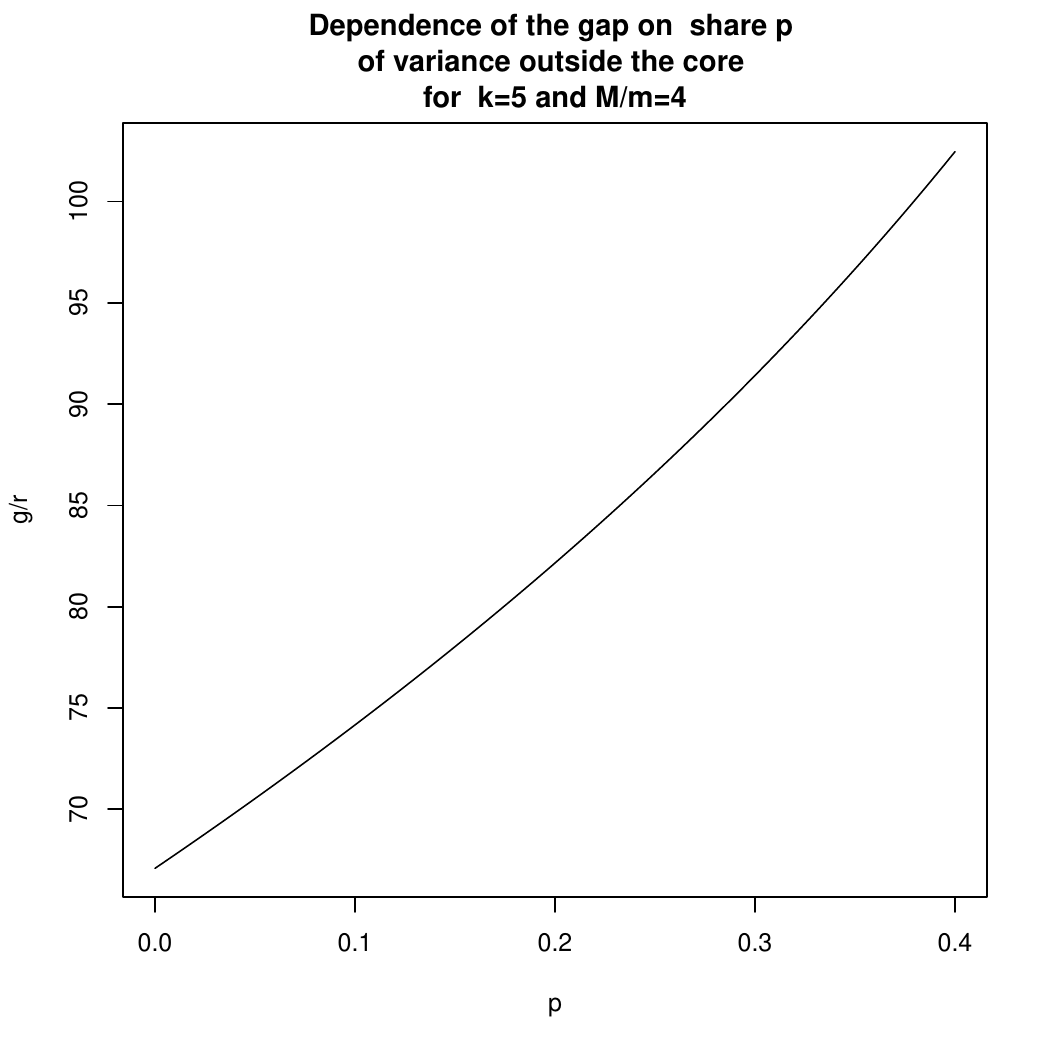}}  %
\caption{Dependence of $g/r$       for $k=5$  on the value of $\mathfrak{p}$ 
}\label{fig:gmMp_fun}
%\end{figure}
\end{minipage} 
}

\section{Core based  approach to  clusterability }
\label{sec: wellclusteredcore}
\begin{figure}
\centering
\CMDFIGgmMpfun
\end{figure}
After the preceding preparatory work, we want to prove a theorem analogous to  Theorem \ref{th:clusterabilitydecidable}, but now allowing for smaller gaps between clusters.  

\begin{theorem}\label{th:coreclusterabilitydecidable}
(i) If the data set is well-clusterable with a gap defined by formulas (\ref{eq:globalgcase2core}) and (\ref{eq:globalgcase1core}), with $r_i$ replaced by their maxima, then with high probability $k$-means++ (after an appropriately chosen number of repetitions)  will discover the respective clustering. 
(ii) If  $k$-means++ (after an appropriately chosen number of repetitions) does not discover  a clustering matching formulas (\ref{eq:globalgcase2core}) and (\ref{eq:globalgcase1core}) (with $r_i$ replaced by their maxima), then with high probability the data set is not well clusterable  with a gap defined by formulas (\ref{eq:globalgcase2core}) and (\ref{eq:globalgcase1core}. 
\end{theorem} 

The rest of the current section is devoted to the proof of the claims of this theorem. 

If we obtained the split, then for each cluster we are able to compute the cluster centre, the radius of the ball containing all the data points of the cluster but the most distant ones, constituting at most $\mathfrak{p}$ of the quality function for the cluster,
 and finally we can check if the gaps between the cluster cores meet the requirement of formulas (\ref{eq:globalgcase2core}) and (\ref{eq:globalgcase1core}). So we are able to decide that we have found that the data set is well-clusterable. 

So let us look at the claim (i). As we already know from preceding Section \ref{sec:corebasedglobalkmeansminimum}, the global minimum of $k$-means coincides with the separation by abovementioned gaps. Hence if there exists a positive probability, that $k$-means++ discovers  the appropriate split, then by repeating independent runs of $k$-means++ and picking the split minimising $k$-means cost function we will increase the probability of finding the global minimum. We will show that we know the number of repetitions needed in advance, if we assume the maximum value of the quotient $M/m$. 

\newcommand{\CMDFIGpmMpfun}{\begin{minipage}{.48\textwidth} 
%\begin{figure}
\centering
\includegraphics[width=0.8\textwidth]{\figaddr{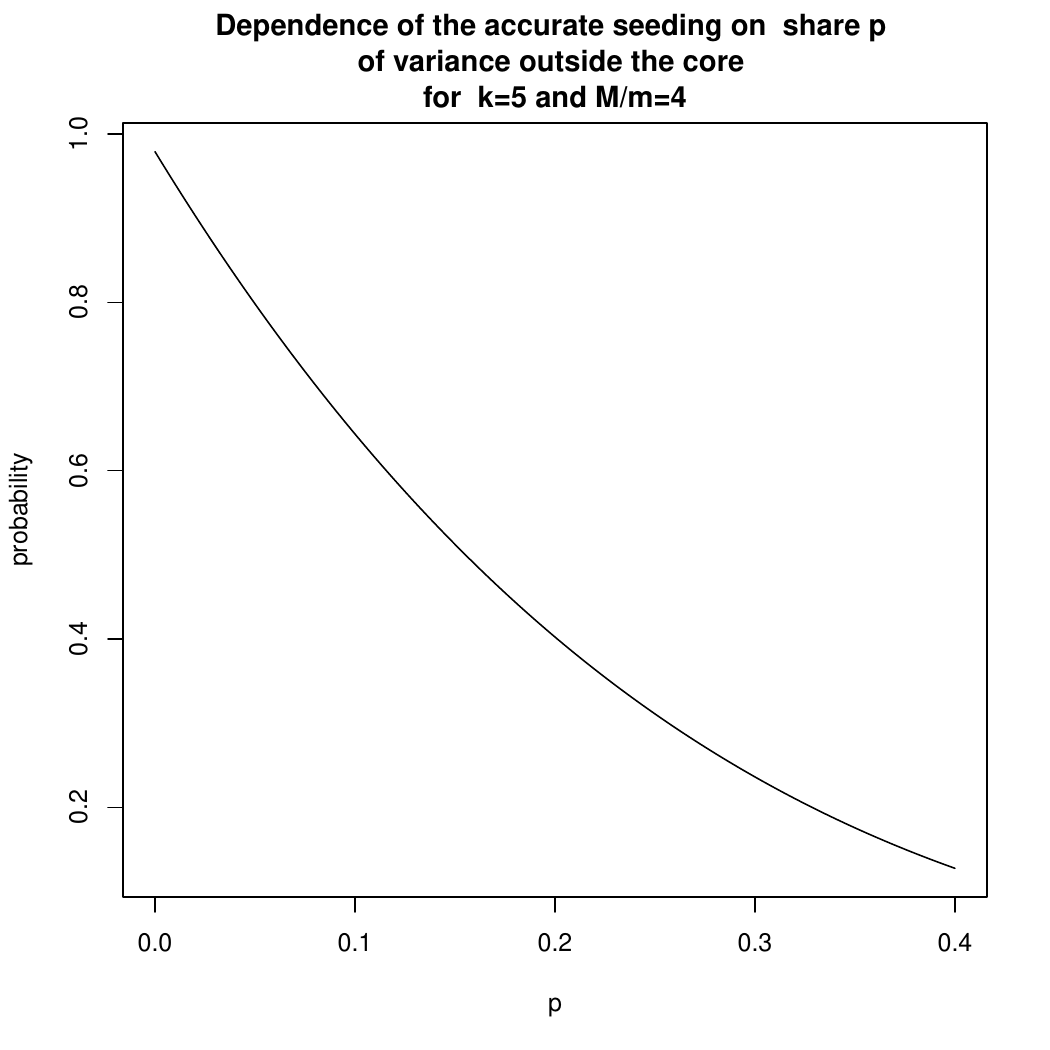}}  %
\caption{
Dependence of the accurate seeding on share $\mathfrak{p}$
of variance outside the core
for $k$=5 and $M/m$=4. 
%
%Probability of seeding each cluster      for $k=5$ 
%for clusters of equal radius when varying cluster cardinalities. 
}\label{fig:pmMp_fun}
%\end{figure}
\end{minipage} 
}

\newcommand{\CMDFIGRmMpfun}{\begin{minipage}{.48\textwidth} 
%\begin{figure}
\centering
\includegraphics[width=0.8\textwidth]{\figaddr{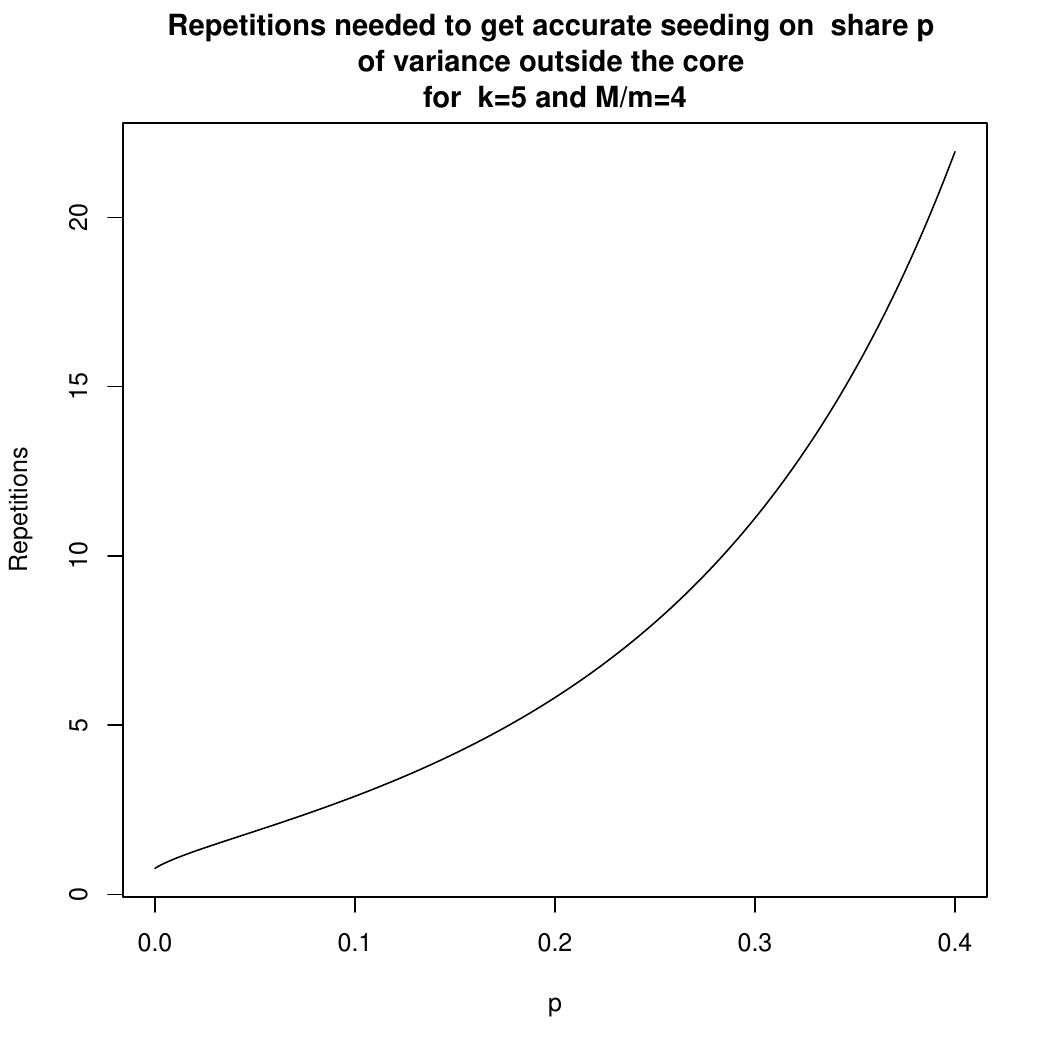}}  %
\caption{Repetitions needed to get accurate seeding on share $\mathfrak{p}$
of variance outside the core
for $k=5$ and $M/m=4$
}\label{fig:RmMp_fun}
%\end{figure}
\end{minipage} 
}

We assume it is granted that  
\begin{equation}\label{eq:globalgcase2p} 
g \ge r \sqrt{k\frac{1+\mathfrak{p}}{1-\mathfrak{p}}\frac{M+n}{   m} }
\end{equation}
for any $i=1,\dots,k$
\begin{equation}\label{eq:globalgcase1p}
g\ge 
k r \sqrt{\frac{1+\mathfrak{p}}{1-\mathfrak{p}}
\frac{  n (n_p +n_q +n )}{    n_{p}n_{q} }
} 
\end{equation}
for any $p,q=1,\dots,k; p\ne q$,
when $n_i, i=1,\dots,k$ is the cardinality of the cluster $i$, 
$M=\max_i n_i$, $m=\min_i n_i$, 
For an illustration of this dependence see  Figure \ref{fig:gmMp_fun}.

%%% FIG

\begin{figure}
\centering
\CMDFIGpmMpfun
\quad
\CMDFIGRmMpfun
\end{figure}

So  let us turn to $k$-means++ seeding.
If already $i$ distinct cluster cores were seeded, 
then the probability that a new cluster core will be seeded (under our assumptions)
amounts to at least  
$$\frac{(k-i) m  (1-\mathfrak{p}) g^2}{(k-i) m g^2+i M \frac{1}{1-\mathfrak{p}}r^2}
\ge 
\frac{(k-i) m k^2 (1-\mathfrak{p}) \frac{1+\mathfrak{p}}{1-\mathfrak{p}}  n (1/m  +1/m +n/m^2 ) }
{(k-i) m k^2 \frac{1+\mathfrak{p}}{1-\mathfrak{p}}  n (1/m  +1/m +n/m^2 ) +i M \frac{1}{1-\mathfrak{p}} }
$$ $$= 
\frac{(k-i)   k^2 (1-\mathfrak{p}) (1+\mathfrak{p})  n (2 +n/m  ) }
{(k-i)   k^2 (1+\mathfrak{p})  n (2 +n/m  ) +i M }
$$ $$\ge
\frac{    k^2  (1-\mathfrak{p}) (1+\mathfrak{p}) n (2 +n/m  ) }
{    k^2 (1+\mathfrak{p})  n (2 +n/m  ) +(k-1) M }
$$
So again the probability of successful seeding will amount to at least:
$$\left(
\frac{    k^2 (1-\mathfrak{p}) (1+\mathfrak{p})  n (2 +n/m  ) }
{    k^2  (1+\mathfrak{p}) n (2 +n/m  ) +(k-1) M }
\right)^{k-1}
$$
$$=(1-\mathfrak{p})^{k-1}\left(1- 
\frac{    (k-1) M }
{    k^2  (1+\mathfrak{p}) n (2 +n/m  ) +(k-1) M }
\right)^{k-1}
$$
$$=(1-\mathfrak{p})^{k-1}\left(1- 
\frac{    (k-1)^2 M }
{    k^2  (1+\mathfrak{p}) n (2 +n/m  ) +(k-1) M }
\frac{1}{k-1}
\right)^{k-1}
$$
 $$\approx  
(1-\mathfrak{p})^{k-1} exp \left(-
\frac{    (k-1)^2 M }
{    k^2  (1+\mathfrak{p}) n (2 +n/m  ) +(k-1) M }
\right)
$$
\Bem{
n=1000;
k=2;
Mdom=20;
a=n/k;
M=a*sqrt(Mdom);
m=a/sqrt(Mdom);

exp(-((k-1)^2 *M )/(    k^2 *  n *(2 +n/m  ) +(k-1)* M))

}%Bem
For an illustration of this dependence see Figure \ref{fig:pmMp_fun}

Apparently in the limit 
the above expression 
lies at about 
$(1-\mathfrak{p})^{k-1}$.

So to achieve the identification of the clustering with probability of at least 
$Pr_{succ}$ (e.g. 
$95\%$), we will need $R$ runs of $k$-means++ where 
$$R=\frac{\log (1-Pr_{succ})}{\log(1-(1-\mathfrak{p})^{k-1})}$$

Note that 
$$  1-(1-\mathfrak{p})^{k-1}\approx 1-e^{-\mathfrak{p}(k-1)}\approx 1- e^{-\mathfrak{p}k}   $$

The effect of doubling $k$ is
$$\frac{1- e^{-\mathfrak{p}2k}}{1- e^{-\mathfrak{p}k}}
= \frac{(1- e^{-\mathfrak{p} k})(1+ e^{-\mathfrak{p}2k})}{1- e^{-\mathfrak{p}k}}
=1+ e^{-\mathfrak{p}2k}$$
that is it is sublinear in the expression $  1-(1-\mathfrak{p})^{k-1}$,
hence $R$ grows slower than reciprocally logarithmically in $k$ and $p$. 
For an illustration of this relation see Figure \ref{fig:RmMp_fun}

%%%% FIG

So far we have concentrated on showing that if the data is well-clusterable, then within practically reasonable number of  $k$-means++ runs the seeding will have the property that each cluster obtains a single seed. But what about the rest of the run of $k$-means? As 
shown in Section \ref{sec:core}, the cluster centres will never switch to balls encompassing other clusters, so that eventually the true cluster structure is detected and minimum of $Q$ is reached. 
This would complete the proof of claim (i).
The demonstration of claim (ii) is straight forward. 
If the data were well-clusterable then $k$-means++ would have failed to identify it with probability of at most $1-Pr_{succ}$. As the well-clusterable data are in practice extremely rare,  the failure of the algorithm to identify a well-clusterable structure induces with probability of at least $Pr_{succ}$ that no such structure exists in the data. 
A detailed proof follows the reasoning of the last part of the proof of Theorem 
\ref{th:clusterabilitydecidable}

\section{Experimental results}\label{sec:exper} 
In order to illustrate the issues raised in this paper, three types of experiments were performed. 
The first experiment, performed on synthetic data, is devoted to the mismatch of gaps between subsets of data 
and the clusterings obtained by common clustering algorithms. The results are shown in Table \ref{tab:fixdistM}.  
The second, performed on synthetic data (Table \ref{tab:synth}), and third, performed on real data (Table \ref{tab:real})
are devoted to demonstration that $k$-means++ is able to discover well-clusterable data.  In particular it is shown that: 
1) If a dataset is well-clusterable as defined 
in Theorem \ref{th:clusterabilitydecidable} 
or Theorem \ref{th:coreclusterabilitydecidable}
(based on Definition \ref{def:WellClus})
then $k$-means++ is able to identify the best clustering (both for real world datasets and synthetic ones)  2) If $k$-means++ cannot find a clustering satisfying well-clusterability, there is no good clustering structure, fitting those definitions, hidden in data (with high probability) for all $k$-means style algorithms.

%-------------------TAB------------------- 
\begin{table} 
\caption{Dependence of the number of errors on   $M$ to $m$ proportion under fixed gap $g/r_{max}$= 2. Other parameters fixed at  $k$=2 $d$=2 $breakgap$=1.
 .}
\label{tab:fixdistM}
\begin{center}  
\begin{tabular}{|l|r|r|r|r|} 
\hline  
$M,m=$& 15 , 9& 480 , 9& 720 , 9& 960 , 9    \\ 
\hline  
Hartigan-Wong 1x& 0 $\pm$ 0& 961.3 $\pm$ 8.8& 974.3 $\pm$ 3.2& 980.9 $\pm$ 4.9 \\
1000 execs in s / relQ & 0.32 / 1& 0.57 / 1.07& 0.67 / 1& 0.79 / 1 \\
\hline  
Lloyd 1x& 1.7 $\pm$ 3.6& 962.5 $\pm$ 5.6& 974.1 $\pm$ 4.9& 980.4 $\pm$ 4.5 \\
1000 execs in s / relQ & 0.31 / 1.01& 0.63 / 1.08& 0.71 / 1.01& 0.79 / 1.01 \\
\hline  
Forgy 1x& 0.8 $\pm$ 1.8& 963.3 $\pm$ 6.5& 975.3 $\pm$ 3.7& 980.9 $\pm$ 4.6 \\
1000 execs in s / relQ & 0.3 / 1.01& 0.63 / 1.08& 0.71 / 1.01& 0.79 / 1.01 \\
\hline  
MacQueen 1x& 1.2 $\pm$ 3.2& 961.8 $\pm$ 4.6& 973.1 $\pm$ 5.2& 982.9 $\pm$ 5.9 \\
1000 execs in s / relQ & 0.31 / 1.01& 0.54 / 1.07& 0.64 / 1.01& 0.72 / 1 \\
\hline  
Hartigan-Wong 10x& 0 $\pm$ 0& 680.9 $\pm$ 25.2& 1000 $\pm$ 0& 1000 $\pm$ 0 \\
1000 execs in s / relQ & 0.81 / 1& 7.23 / 1.05& 10.24 / 1& 13.42 / 1 \\
\hline  
Lloyd 10x& 0 $\pm$ 0& 678.6 $\pm$ 10.1& 1000 $\pm$ 0& 1000 $\pm$ 0 \\
1000 execs in s / relQ & 0.72 / 1& 7.92 / 1.05& 10.73 / 1& 13.6 / 1 \\
\hline  
Forgy 10x& 0 $\pm$ 0& 666.7 $\pm$ 14.2& 1000 $\pm$ 0& 1000 $\pm$ 0 \\
1000 execs in s / relQ & 0.72 / 1& 7.94 / 1.05& 10.73 / 1& 13.52 / 1 \\
\hline  
MacQueen 10x& 0 $\pm$ 0& 677.2 $\pm$ 12& 1000 $\pm$ 0& 1000 $\pm$ 0 \\
1000 execs in s / relQ & 0.73 / 1& 7.15 / 1.05& 9.99 / 1& 12.92 / 1 \\
\hline  
Hartigan-Wong 20x& 0 $\pm$ 0& 457.5 $\pm$ 14& 1000 $\pm$ 0& 1000 $\pm$ 0 \\
1000 execs in s / relQ & 1.08 / 1& 8.75 / 1.03& 12.63 / 1& 16.59 / 1 \\
\hline  
Lloyd 20x& 0 $\pm$ 0& 451.6 $\pm$ 9.5& 1000 $\pm$ 0& 1000 $\pm$ 0 \\
1000 execs in s / relQ & 0.9 / 1& 9.96 / 1.03& 13.58 / 1& 16.79 / 1 \\
\hline  
Forgy 20x& 0 $\pm$ 0& 457.8 $\pm$ 12.4& 1000 $\pm$ 0& 1000 $\pm$ 0 \\
1000 execs in s / relQ & 0.9 / 1& 9.95 / 1.03& 13.44 / 1& 16.78 / 1 \\
\hline  
MacQueen 20x& 0 $\pm$ 0& 452.6 $\pm$ 14.7& 1000 $\pm$ 0& 1000 $\pm$ 0 \\
1000 execs in s / relQ & 0.89 / 1& 8.54 / 1.03& 12.11 / 1& 15.69 / 1 \\
\hline  
cmeans Fuzzy & 0 $\pm$ 0& 973.7 $\pm$ 7.9& 995.4 $\pm$ 6.6& 1000 $\pm$ 0 \\
1000 execs in s / relQ & 0.67 / 1& 5.37 / 1.09& 7.37 / 1.02& 9.43 / 1.02 \\
\hline  
ufcl Fuzzy  & 235.5 $\pm$ 59.6& 964.4 $\pm$ 8.9& 976.8 $\pm$ 6.5& 981.3 $\pm$ 5.4 \\
1000 execs in s / relQ & 0.8 / 1.52& 6.3 / 1.14& 8.79 / 1.06& 11.22 / 1.05 \\
\hline  
single link   & 0 $\pm$ 0& 0 $\pm$ 0& 0 $\pm$ 0& 0 $\pm$ 0 \\
1000 execs in s / relQ & 0.2 / 1& 13.76 / 1& 34.39 / 1.06& 49.01 / 1.15 \\
\hline  
kmeans++  & 0 $\pm$ 0& 734.7 $\pm$ 17& 811 $\pm$ 11.5& 844 $\pm$ 12.5 \\
1000 execs in s / relQ & 1.14 / 1& 16.01 / 1.05& 22.53 / 1.01& 28.77 / 1.02 \\
\hline  
kmeans++ 2x& 0 $\pm$ 0& 544.7 $\pm$ 28.2& 959.3 $\pm$ 6.1& 976.9 $\pm$ 4.6 \\
1000 execs in s / relQ & 2.26 / 1& 31.96 / 1.04& 44.38 / 1& 57.49 / 1 \\
\hline  
\end{tabular}  
\end{center}  
\end{table}

%-------------------TAB------------------- 
\begin{table} 
\caption{Reconstruction of the clusters in synthetic data. Notation in text.  }
\label{tab:synth}
\begin{center}  
\begin{tabular}{|p{5cm}|r|r|r|r|r|r|} 
\hline
Dataset   &
\multicolumn{2}{|c|}{Errors}& 
\multicolumn{2}{|c|}{WC disc.}& 
\multicolumn{2}{|c|}{WC not disc. }
\\
\cline{2-7}
& $k$-means   & $k$-means++ 
& cc 
& wc
& cc  
& wc 
\\
 \hline 
%\Bem{%global0
	$k= 2 , 	\mathfrak{p}= 0 , 
	\Bem{M= 75 ,	m= 45 ,
	d= 10 ,}	gp= 0.5 $
&  0  &  0 & 0 & 0 & 100 & 0 \\ 
\Bem{
 ==================================================
Paramweter setting:
	k= 2 
	pfrak= 0.1 
	M= 75 
	m= 45 
	d= 10 
	r_max= 5 
	breakgap= 2

*****test run*********
  10         20         30         40         50         60         70         80         90         100       
Errors in recognizing the intrinsic clustering 
kmeans++ 0 kmeans 0 

Decisions if the data set is clusterable or not - kmeans++  
clu if correct  0  clu if not correct  0  not clu if   correct  100  not clu if not correct  0 }%\Bem
\hline 
	$k= 2 , 	\mathfrak{p}= 0.1 , 
	\Bem{M= 75 ,	m= 45 ,
	d= 10 ,}	gp= 0.5 $
&  0  &  0 & 0 & 0 & 100 & 0 \\ 
\Bem{
 ==================================================
Paramweter setting:
	k= 3 
	pfrak= 0 
	M= 75 
	m= 45 
	d= 10 
	r_max= 5 
	breakgap= 2

*****test run*********
  10         20         30         40         50         60         70         80         90         100       
Errors in recognizing the intrinsic clustering 
kmeans++ 0 kmeans 38 

Decisions if the data set is clusterable or not - kmeans++  
clu if correct  0  clu if not correct  0  not clu if   correct  100  not clu if not correct  0 }%\Bem
\hline 
	$k= 3 , 	\mathfrak{p}= 0 , 
	\Bem{M= 75 ,	m= 45 ,
	d= 10 ,}	gp= 0.5 $
&  38  &  0 & 0 & 0 & 100 & 0 \\ 
\Bem{
 ==================================================
Paramweter setting:
	k= 3 
	pfrak= 0.1 
	M= 75 
	m= 45 
	d= 10 
	r_max= 5 
	breakgap= 2

*****test run*********
  10         20         30         40         50         60         70         80         90         100       
Errors in recognizing the intrinsic clustering 
kmeans++ 0 kmeans 26 

Decisions if the data set is clusterable or not - kmeans++  
clu if correct  0  clu if not correct  0  not clu if   correct  100  not clu if not correct  0 }%\Bem
\hline 
	$k= 3 , 	\mathfrak{p}= 0.1 , 
	\Bem{M= 75 ,	m= 45 ,
	d= 10 ,}	gp= 0.5 $
&  26  &  0 & 0 & 0 & 100 & 0 \\ 
\Bem{
 ==================================================
Paramweter setting:
	k= 5 
	pfrak= 0 
	M= 75 
	m= 45 
	d= 10 
	r_max= 5 
	breakgap= 2

*****test run*********
  10         20         30         40         50         60         70         80         90         100       
Errors in recognizing the intrinsic clustering 
kmeans++ 0 kmeans 74 

Decisions if the data set is clusterable or not - kmeans++  
clu if correct  0  clu if not correct  0  not clu if   correct  100  not clu if not correct  0 }%\Bem
\hline 
	$k= 5 , 	\mathfrak{p}= 0 , 
	\Bem{M= 75 ,	m= 45 ,
	d= 10 ,}	gp= 0.5 $
&  74  &  0 & 0 & 0 & 100 & 0 \\ 
\Bem{
 ==================================================
Paramweter setting:
	k= 5 
	pfrak= 0.1 
	M= 75 
	m= 45 
	d= 10 
	r_max= 5 
	breakgap= 2

*****test run*********
  10         20         30         40         50         60         70         80         90         100       
Errors in recognizing the intrinsic clustering 
kmeans++ 0 kmeans 66 

Decisions if the data set is clusterable or not - kmeans++  
clu if correct  0  clu if not correct  0  not clu if   correct  100  not clu if not correct  0 }%\Bem
\hline 
	$k= 5 , 	\mathfrak{p}= 0.1 , 
	\Bem{M= 75 ,	m= 45 ,
	d= 10 ,}	gp= 0.5 $
&  66  &  0 & 0 & 0 & 100 & 0 \\ 
\Bem{
 ==================================================
Paramweter setting:
	k= 2 
	pfrak= 0 
	M= 75 
	m= 45 
	d= 10 
	r_max= 5 
	breakgap= 1.42857142857143

*****test run*********
  10         20         30         40         50         60         70         80         90         100       
Errors in recognizing the intrinsic clustering 
kmeans++ 0 kmeans 0 

Decisions if the data set is clusterable or not - kmeans++  
clu if correct  0  clu if not correct  0  not clu if   correct  100  not clu if not correct  0 }%\Bem
\hline 
	$k= 2 , 	\mathfrak{p}= 0 , 
	\Bem{M= 75 ,	m= 45 ,
	d= 10 ,}	gp= 0.7 $
&  0  &  0 & 0 & 0 & 100 & 0 \\ 
\Bem{
 ==================================================
Paramweter setting:
	k= 2 
	pfrak= 0.1 
	M= 75 
	m= 45 
	d= 10 
	r_max= 5 
	breakgap= 1.42857142857143

*****test run*********
  10         20         30         40         50         60         70         80         90         100       
Errors in recognizing the intrinsic clustering 
kmeans++ 0 kmeans 0 

Decisions if the data set is clusterable or not - kmeans++  
clu if correct  0  clu if not correct  0  not clu if   correct  100  not clu if not correct  0 }%\Bem
\hline 
	$k= 2 , 	\mathfrak{p}= 0.1 , 
	\Bem{M= 75 ,	m= 45 ,
	d= 10 ,}	gp= 0.7 $
&  0  &  0 & 0 & 0 & 100 & 0 \\ 
\Bem{
 ==================================================
Paramweter setting:
	k= 3 
	pfrak= 0 
	M= 75 
	m= 45 
	d= 10 
	r_max= 5 
	breakgap= 1.42857142857143

*****test run*********
  10         20         30         40         50         60         70         80         90         100       
Errors in recognizing the intrinsic clustering 
kmeans++ 2 kmeans 50 

Decisions if the data set is clusterable or not - kmeans++  
clu if correct  0  clu if not correct  0  not clu if   correct  98  not clu if not correct  2 }%\Bem
\hline 
	$k= 3 , 	\mathfrak{p}= 0 , 
	\Bem{M= 75 ,	m= 45 ,
	d= 10 ,}	gp= 0.7 $
&  50  &  2 & 0 & 0 & 98 & 2 \\ 
\Bem{
 ==================================================
Paramweter setting:
	k= 3 
	pfrak= 0.1 
	M= 75 
	m= 45 
	d= 10 
	r_max= 5 
	breakgap= 1.42857142857143

*****test run*********
  10         20         30         40         50         60         70         80         90         100       
Errors in recognizing the intrinsic clustering 
kmeans++ 0 kmeans 37 

Decisions if the data set is clusterable or not - kmeans++  
clu if correct  0  clu if not correct  0  not clu if   correct  100  not clu if not correct  0 }%\Bem
\hline 
	$k= 3 , 	\mathfrak{p}= 0.1 , 
	\Bem{M= 75 ,	m= 45 ,
	d= 10 ,}	gp= 0.7 $
&  37  &  0 & 0 & 0 & 100 & 0 \\ 
\Bem{
 ==================================================
Paramweter setting:
	k= 5 
	pfrak= 0 
	M= 75 
	m= 45 
	d= 10 
	r_max= 5 
	breakgap= 1.42857142857143

*****test run*********
  10         20         30         40         50         60         70         80         90         100       
Errors in recognizing the intrinsic clustering 
kmeans++ 0 kmeans 78 

Decisions if the data set is clusterable or not - kmeans++  
clu if correct  0  clu if not correct  0  not clu if   correct  100  not clu if not correct  0 }%\Bem
\hline 
	$k= 5 , 	\mathfrak{p}= 0 , 
	\Bem{M= 75 ,	m= 45 ,
	d= 10 ,}	gp= 0.7 $
&  78  &  0 & 0 & 0 & 100 & 0 \\ 
\Bem{
 ==================================================
Paramweter setting:
	k= 5 
	pfrak= 0.1 
	M= 75 
	m= 45 
	d= 10 
	r_max= 5 
	breakgap= 1.42857142857143

*****test run*********
  10         20         30         40         50         60         70         80         90         100       
Errors in recognizing the intrinsic clustering 
kmeans++ 0 kmeans 73 

Decisions if the data set is clusterable or not - kmeans++  
clu if correct  0  clu if not correct  0  not clu if   correct  100  not clu if not correct  0 }%\Bem
\hline 
	$k= 5 , 	\mathfrak{p}= 0.1 , 
	\Bem{M= 75 ,	m= 45 ,
	d= 10 ,}	gp= 0.7 $
&  73  &  0 & 0 & 0 & 100 & 0 \\ 
\Bem{
 ==================================================
Paramweter setting:
	k= 2 
	pfrak= 0 
	M= 75 
	m= 45 
	d= 10 
	r_max= 5 
	breakgap= 1.11111111111111

*****test run*********
  10         20         30         40         50         60         70         80         90         100       
Errors in recognizing the intrinsic clustering 
kmeans++ 0 kmeans 0 

Decisions if the data set is clusterable or not - kmeans++  
clu if correct  0  clu if not correct  0  not clu if   correct  100  not clu if not correct  0 }%\Bem
\hline 
	$k= 2 , 	\mathfrak{p}= 0 , 
	\Bem{M= 75 ,	m= 45 ,
	d= 10 ,}	gp= 0.9 $
&  0  &  0 & 0 & 0 & 100 & 0 \\ 
\Bem{
 ==================================================
Paramweter setting:
	k= 2 
	pfrak= 0.1 
	M= 75 
	m= 45 
	d= 10 
	r_max= 5 
	breakgap= 1.11111111111111

*****test run*********
  10         20         30         40         50         60         70         80         90         100       
Errors in recognizing the intrinsic clustering 
kmeans++ 0 kmeans 0 

Decisions if the data set is clusterable or not - kmeans++  
clu if correct  0  clu if not correct  0  not clu if   correct  100  not clu if not correct  0 }%\Bem
\hline 
	$k= 2 , 	\mathfrak{p}= 0.1 , 
	\Bem{M= 75 ,	m= 45 ,
	d= 10 ,}	gp= 0.9 $
&  0  &  0 & 0 & 0 & 100 & 0 \\ 
\Bem{
 ==================================================
Paramweter setting:
	k= 3 
	pfrak= 0 
	M= 75 
	m= 45 
	d= 10 
	r_max= 5 
	breakgap= 1.11111111111111

*****test run*********
  10         20         30         40         50         60         70         80         90         100       
Errors in recognizing the intrinsic clustering 
kmeans++ 1 kmeans 41 

Decisions if the data set is clusterable or not - kmeans++  
clu if correct  0  clu if not correct  0  not clu if   correct  99  not clu if not correct  1 }%\Bem
\hline 
	$k= 3 , 	\mathfrak{p}= 0 , 
	\Bem{M= 75 ,	m= 45 ,
	d= 10 ,}	gp= 0.9 $
&  41  &  1 & 0 & 0 & 99 & 1 \\ 
\Bem{
 ==================================================
Paramweter setting:
	k= 3 
	pfrak= 0.1 
	M= 75 
	m= 45 
	d= 10 
	r_max= 5 
	breakgap= 1.11111111111111

*****test run*********
  10         20         30         40         50         60         70         80         90         100       
Errors in recognizing the intrinsic clustering 
kmeans++ 1 kmeans 29 

Decisions if the data set is clusterable or not - kmeans++  
clu if correct  0  clu if not correct  0  not clu if   correct  99  not clu if not correct  1 }%\Bem
\hline 
	$k= 3 , 	\mathfrak{p}= 0.1 , 
	\Bem{M= 75 ,	m= 45 ,
	d= 10 ,}	gp= 0.9 $
&  29  &  1 & 0 & 0 & 99 & 1 \\ 
\Bem{
 ==================================================
Paramweter setting:
	k= 5 
	pfrak= 0 
	M= 75 
	m= 45 
	d= 10 
	r_max= 5 
	breakgap= 1.11111111111111

*****test run*********
  10         20         30         40         50         60         70         80         90         100       
Errors in recognizing the intrinsic clustering 
kmeans++ 0 kmeans 74 

Decisions if the data set is clusterable or not - kmeans++  
clu if correct  0  clu if not correct  0  not clu if   correct  100  not clu if not correct  0 }%\Bem
\hline 
	$k= 5 , 	\mathfrak{p}= 0 , 
	\Bem{M= 75 ,	m= 45 ,
	d= 10 ,}	gp= 0.9 $
&  74  &  0 & 0 & 0 & 100 & 0 \\ 
\Bem{
 ==================================================
Paramweter setting:
	k= 5 
	pfrak= 0.1 
	M= 75 
	m= 45 
	d= 10 
	r_max= 5 
	breakgap= 1.11111111111111

*****test run*********
  10         20         30         40         50         60         70         80         90         100       
Errors in recognizing the intrinsic clustering 
kmeans++ 0 kmeans 78 

Decisions if the data set is clusterable or not - kmeans++  
clu if correct  0  clu if not correct  0  not clu if   correct  100  not clu if not correct  0 }%\Bem
\hline 
	$k= 5 , 	\mathfrak{p}= 0.1 , 
	\Bem{M= 75 ,	m= 45 ,
	d= 10 ,}	gp= 0.9 $
&  78  &  0 & 0 & 0 & 100 & 0 \\ 
\Bem{
 ==================================================
Paramweter setting:
	k= 2 
	pfrak= 0 
	M= 75 
	m= 45 
	d= 10 
	r_max= 5 
	breakgap= 0.909090909090909

*****test run*********
  10         20         30         40         50         60         70         80         90         100       
Errors in recognizing the intrinsic clustering 
kmeans++ 0 kmeans 0 

Decisions if the data set is clusterable or not - kmeans++  
clu if correct  100  clu if not correct  0  not clu if   correct  0  not clu if not correct  0 }%\Bem
\hline 
	$k= 2 , 	\mathfrak{p}= 0 , 
	\Bem{M= 75 ,	m= 45 ,
	d= 10 ,}	gp= 1.1 $
&  0  &  0 & 100 & 0 & 0 & 0 \\ 
\Bem{
 ==================================================
Paramweter setting:
	k= 2 
	pfrak= 0.1 
	M= 75 
	m= 45 
	d= 10 
	r_max= 5 
	breakgap= 0.909090909090909

*****test run*********
  10         20         30         40         50         60         70         80         90         100       
Errors in recognizing the intrinsic clustering 
kmeans++ 0 kmeans 0 

Decisions if the data set is clusterable or not - kmeans++  
clu if correct  100  clu if not correct  0  not clu if   correct  0  not clu if not correct  0 }%\Bem
\hline 
	$k= 2 , 	\mathfrak{p}= 0.1 , 
	\Bem{M= 75 ,	m= 45 ,
	d= 10 ,}	gp= 1.1 $
&  0  &  0 & 100 & 0 & 0 & 0 \\ 
\Bem{
 ==================================================
Paramweter setting:
	k= 3 
	pfrak= 0 
	M= 75 
	m= 45 
	d= 10 
	r_max= 5 
	breakgap= 0.909090909090909

*****test run*********
  10         20         30         40         50         60         70         80         90         100       
Errors in recognizing the intrinsic clustering 
kmeans++ 0 kmeans 41 

Decisions if the data set is clusterable or not - kmeans++  
clu if correct  100  clu if not correct  0  not clu if   correct  0  not clu if not correct  0 }%\Bem
\hline 
	$k= 3 , 	\mathfrak{p}= 0 , 
	\Bem{M= 75 ,	m= 45 ,
	d= 10 ,}	gp= 1.1 $
&  41  &  0 & 100 & 0 & 0 & 0 \\ 
\Bem{
 ==================================================
Paramweter setting:
	k= 3 
	pfrak= 0.1 
	M= 75 
	m= 45 
	d= 10 
	r_max= 5 
	breakgap= 0.909090909090909

*****test run*********
  10         20         30         40         50         60         70         80         90         100       
Errors in recognizing the intrinsic clustering 
kmeans++ 0 kmeans 39 

Decisions if the data set is clusterable or not - kmeans++  
clu if correct  100  clu if not correct  0  not clu if   correct  0  not clu if not correct  0 }%\Bem
\hline 
	$k= 3 , 	\mathfrak{p}= 0.1 , 
	\Bem{M= 75 ,	m= 45 ,
	d= 10 ,}	gp= 1.1 $
&  39  &  0 & 100 & 0 & 0 & 0 \\ 
\Bem{
 ==================================================
Paramweter setting:
	k= 5 
	pfrak= 0 
	M= 75 
	m= 45 
	d= 10 
	r_max= 5 
	breakgap= 0.909090909090909

*****test run*********
  10         20         30         40         50         60         70         80         90         100       
Errors in recognizing the intrinsic clustering 
kmeans++ 0 kmeans 80 

Decisions if the data set is clusterable or not - kmeans++  
clu if correct  100  clu if not correct  0  not clu if   correct  0  not clu if not correct  0 }%\Bem
\hline 
	$k= 5 , 	\mathfrak{p}= 0 , 
	\Bem{M= 75 ,	m= 45 ,
	d= 10 ,}	gp= 1.1 $
&  80  &  0 & 100 & 0 & 0 & 0 \\ 
\Bem{
 ==================================================
Paramweter setting:
	k= 5 
	pfrak= 0.1 
	M= 75 
	m= 45 
	d= 10 
	r_max= 5 
	breakgap= 0.909090909090909

*****test run*********
  10         20         30         40         50         60         70         80         90         100       
Errors in recognizing the intrinsic clustering 
kmeans++ 0 kmeans 80 

Decisions if the data set is clusterable or not - kmeans++  
clu if correct  100  clu if not correct  0  not clu if   correct  0  not clu if not correct  0 }%\Bem
\hline 
	$k= 5 , 	\mathfrak{p}= 0.1 , 
	\Bem{M= 75 ,	m= 45 ,
	d= 10 ,}	gp= 1.1 $
&  80  &  0 & 100 & 0 & 0 & 0 \\ 
\Bem{
 ==================================================
Paramweter setting:
	k= 2 
	pfrak= 0 
	M= 75 
	m= 45 
	d= 10 
	r_max= 5 
	breakgap= 0.769230769230769

*****test run*********
  10         20         30         40         50         60         70         80         90         100       
Errors in recognizing the intrinsic clustering 
kmeans++ 0 kmeans 0 

Decisions if the data set is clusterable or not - kmeans++  
clu if correct  100  clu if not correct  0  not clu if   correct  0  not clu if not correct  0 }%\Bem
\hline 
	$k= 2 , 	\mathfrak{p}= 0 , 
	\Bem{M= 75 ,	m= 45 ,
	d= 10 ,}	gp= 1.3 $
&  0  &  0 & 100 & 0 & 0 & 0 \\ 
\Bem{
 ==================================================
Paramweter setting:
	k= 2 
	pfrak= 0.1 
	M= 75 
	m= 45 
	d= 10 
	r_max= 5 
	breakgap= 0.769230769230769

*****test run*********
  10         20         30         40         50         60         70         80         90         100       
Errors in recognizing the intrinsic clustering 
kmeans++ 0 kmeans 0 

Decisions if the data set is clusterable or not - kmeans++  
clu if correct  100  clu if not correct  0  not clu if   correct  0  not clu if not correct  0 }%\Bem
\hline 
	$k= 2 , 	\mathfrak{p}= 0.1 , 
	\Bem{M= 75 ,	m= 45 ,
	d= 10 ,}	gp= 1.3 $
&  0  &  0 & 100 & 0 & 0 & 0 \\ 
\Bem{
 ==================================================
Paramweter setting:
	k= 3 
	pfrak= 0 
	M= 75 
	m= 45 
	d= 10 
	r_max= 5 
	breakgap= 0.769230769230769

*****test run*********
  10         20         30         40         50         60         70         80         90         100       
Errors in recognizing the intrinsic clustering 
kmeans++ 0 kmeans 43 

Decisions if the data set is clusterable or not - kmeans++  
clu if correct  100  clu if not correct  0  not clu if   correct  0  not clu if not correct  0 }%\Bem
\hline 
	$k= 3 , 	\mathfrak{p}= 0 , 
	\Bem{M= 75 ,	m= 45 ,
	d= 10 ,}	gp= 1.3 $
&  43  &  0 & 100 & 0 & 0 & 0 \\ 
\Bem{
 ==================================================
Paramweter setting:
	k= 3 
	pfrak= 0.1 
	M= 75 
	m= 45 
	d= 10 
	r_max= 5 
	breakgap= 0.769230769230769

*****test run*********
  10         20         30         40         50         60         70         80         90         100       
Errors in recognizing the intrinsic clustering 
kmeans++ 0 kmeans 38 

Decisions if the data set is clusterable or not - kmeans++  
clu if correct  100  clu if not correct  0  not clu if   correct  0  not clu if not correct  0 }%\Bem
\hline 
	$k= 3 , 	\mathfrak{p}= 0.1 , 
	\Bem{M= 75 ,	m= 45 ,
	d= 10 ,}	gp= 1.3 $
&  38  &  0 & 100 & 0 & 0 & 0 \\ 
\Bem{
 ==================================================
Paramweter setting:
	k= 5 
	pfrak= 0 
	M= 75 
	m= 45 
	d= 10 
	r_max= 5 
	breakgap= 0.769230769230769

*****test run*********
  10         20         30         40         50         60         70         80         90         100       
Errors in recognizing the intrinsic clustering 
kmeans++ 0 kmeans 80 

Decisions if the data set is clusterable or not - kmeans++  
clu if correct  100  clu if not correct  0  not clu if   correct  0  not clu if not correct  0 }%\Bem
\hline 
	$k= 5 , 	\mathfrak{p}= 0 , 
	\Bem{M= 75 ,	m= 45 ,
	d= 10 ,}	gp= 1.3 $
&  80  &  0 & 100 & 0 & 0 & 0 \\ 
\Bem{
 ==================================================
Paramweter setting:
	k= 5 
	pfrak= 0.1 
	M= 75 
	m= 45 
	d= 10 
	r_max= 5 
	breakgap= 0.769230769230769

*****test run*********
  10         20         30         40         50         60         70         80         90         100       
Errors in recognizing the intrinsic clustering 
kmeans++ 0 kmeans 75 

Decisions if the data set is clusterable or not - kmeans++  
clu if correct  100  clu if not correct  0  not clu if   correct  0  not clu if not correct  0 }%\Bem
\hline 
	$k= 5 , 	\mathfrak{p}= 0.1 , 
	\Bem{M= 75 ,	m= 45 ,
	d= 10 ,}	gp= 1.3 $
&  75  &  0 & 100 & 0 & 0 & 0 \\ 
\Bem{
 ==================================================
Paramweter setting:
	k= 2 
	pfrak= 0 
	M= 75 
	m= 45 
	d= 10 
	r_max= 5 
	breakgap= 0.666666666666667

*****test run*********
  10         20         30         40         50         60         70         80         90         100       
Errors in recognizing the intrinsic clustering 
kmeans++ 0 kmeans 0 

Decisions if the data set is clusterable or not - kmeans++  
clu if correct  100  clu if not correct  0  not clu if   correct  0  not clu if not correct  0 }%\Bem
\hline 
	$k= 2 , 	\mathfrak{p}= 0 , 
	\Bem{M= 75 ,	m= 45 ,
	d= 10 ,}	gp= 1.5 $
&  0  &  0 & 100 & 0 & 0 & 0 \\ 
\Bem{
 ==================================================
Paramweter setting:
	k= 2 
	pfrak= 0.1 
	M= 75 
	m= 45 
	d= 10 
	r_max= 5 
	breakgap= 0.666666666666667

*****test run*********
  10         20         30         40         50         60         70         80         90         100       
Errors in recognizing the intrinsic clustering 
kmeans++ 0 kmeans 0 

Decisions if the data set is clusterable or not - kmeans++  
clu if correct  100  clu if not correct  0  not clu if   correct  0  not clu if not correct  0 }%\Bem
\hline 
	$k= 2 , 	\mathfrak{p}= 0.1 , 
	\Bem{M= 75 ,	m= 45 ,
	d= 10 ,}	gp= 1.5 $
&  0  &  0 & 100 & 0 & 0 & 0 \\ 
\Bem{
 ==================================================
Paramweter setting:
	k= 3 
	pfrak= 0 
	M= 75 
	m= 45 
	d= 10 
	r_max= 5 
	breakgap= 0.666666666666667

*****test run*********
  10         20         30         40         50         60         70         80         90         100       
Errors in recognizing the intrinsic clustering 
kmeans++ 1 kmeans 33 

Decisions if the data set is clusterable or not - kmeans++  
clu if correct  99  clu if not correct  0  not clu if   correct  0  not clu if not correct  1 }%\Bem
\hline 
	$k= 3 , 	\mathfrak{p}= 0 , 
	\Bem{M= 75 ,	m= 45 ,
	d= 10 ,}	gp= 1.5 $
&  33  &  1 & 99 & 0 & 0 & 1 \\ 
\Bem{
 ==================================================
Paramweter setting:
	k= 3 
	pfrak= 0.1 
	M= 75 
	m= 45 
	d= 10 
	r_max= 5 
	breakgap= 0.666666666666667

*****test run*********
  10         20         30         40         50         60         70         80         90         100       
Errors in recognizing the intrinsic clustering 
kmeans++ 0 kmeans 33 

Decisions if the data set is clusterable or not - kmeans++  
clu if correct  100  clu if not correct  0  not clu if   correct  0  not clu if not correct  0 }%\Bem
\hline 
	$k= 3 , 	\mathfrak{p}= 0.1 , 
	\Bem{M= 75 ,	m= 45 ,
	d= 10 ,}	gp= 1.5 $
&  33  &  0 & 100 & 0 & 0 & 0 \\ 
\Bem{
 ==================================================
Paramweter setting:
	k= 5 
	pfrak= 0 
	M= 75 
	m= 45 
	d= 10 
	r_max= 5 
	breakgap= 0.666666666666667

*****test run*********
  10         20         30         40         50         60         70         80         90         100       
Errors in recognizing the intrinsic clustering 
kmeans++ 0 kmeans 75 

Decisions if the data set is clusterable or not - kmeans++  
clu if correct  100  clu if not correct  0  not clu if   correct  0  not clu if not correct  0 }%\Bem
\hline 
	$k= 5 , 	\mathfrak{p}= 0 , 
	\Bem{M= 75 ,	m= 45 ,
	d= 10 ,}	gp= 1.5 $
&  75  &  0 & 100 & 0 & 0 & 0 \\ 
\Bem{
 ==================================================
Paramweter setting:
	k= 5 
	pfrak= 0.1 
	M= 75 
	m= 45 
	d= 10 
	r_max= 5 
	breakgap= 0.666666666666667

*****test run*********
  10         20         30         40         50         60         70         80         90         100       
Errors in recognizing the intrinsic clustering 
kmeans++ 0 kmeans 79 

Decisions if the data set is clusterable or not - kmeans++  
clu if correct  100  clu if not correct  0  not clu if   correct  0  not clu if not correct  0 }%\Bem
\hline 
	$k= 5 , 	\mathfrak{p}= 0.1 , 
	\Bem{M= 75 ,	m= 45 ,
	d= 10 ,}	gp= 1.5 $
&  79  &  0 & 100 & 0 & 0 & 0 \\ 
\Bem{
 ==================================================
Paramweter setting:
	k= 2 
	pfrak= 0 
	M= 75 
	m= 45 
	d= 10 
	r_max= 5 
	breakgap= 0.588235294117647

*****test run*********
  10         20         30         40         50         60         70         80         90         100       
Errors in recognizing the intrinsic clustering 
kmeans++ 0 kmeans 0 

Decisions if the data set is clusterable or not - kmeans++  
clu if correct  100  clu if not correct  0  not clu if   correct  0  not clu if not correct  0 }%\Bem
\hline 
%\Bem{%global0

  \hline  
\end{tabular}  
\end{center}  
\end{table}

\nonIpiArchiv{
%-------------------TAB------------------- 
\begin{table} 
\caption{Reconstruction of the clusters in real data from $R$ library $datasets$. Notation in text.  }
\label{tab:real}
\begin{center}  
\begin{tabular}{|l|r|r|r|r|r|r|r|r|r|} 
\hline
Dataset & gap type& clusters &$\mathfrak{p}$ &
\multicolumn{2}{|c|}{Errors}& 
\multicolumn{2}{|c|}{WC disc.}& 
\multicolumn{2}{|c|}{WC not disc. }
\\
\cline{5-10}
&&&& $k$-means   & $k$-means++ 
& cc 
& wc
& cc  
& wc 
\\
 \hline  
\Bem{ 

------------------------------------------
The set to analyse DNase
 The numerical columns [1] "conc"    "density"

 The number of rows   176
 The sizes of ground truth clusters  [1]  44 132
\n }%\Bem{
\Bem{

 phase  orig  pfrak 0 ---- sufficient gap quot  0.0472230819828586 

 sufficient gap quot  0.0472230819828586 
  10         20         30         40         50         60         70         80         90         100       
Errors in recognizing the intrinsic clustering 
kmeans++ 39 kmeans 69 

Decisions if the data set is clusterable or not - kmeans++  
clu if correct  0  clu if not correct  0  not clu if   correct  61  not clu if not correct  39 
\hline
 }%\Bem{ 
    DNase & orig & 2 & 0 &  69  &  39 & 0 & 0 & 61 & 39 \\ 
\Bem{

 phase  orig  pfrak 0.1 ---- sufficient gap quot  0.058973060967884 

 sufficient gap quot  0.058973060967884 
  10         20         30         40         50         60         70         80         90         100       
Errors in recognizing the intrinsic clustering 
kmeans++ 38 kmeans 59 

Decisions if the data set is clusterable or not - kmeans++  
clu if correct  0  clu if not correct  0  not clu if   correct  62  not clu if not correct  38 

 }%\Bem{ 
 \hline   DNase & orig & 2 & 0.1 &  59  &  38 & 0 & 0 & 62 & 38 \\ 
\Bem{

 phase  g/2  pfrak 0 ---- sufficient gap quot  0.0472230819828586 

 sufficient gap quot  0.402152772402553 
  10         20         30         40         50         60         70         80         90         100       
Errors in recognizing the intrinsic clustering 
kmeans++ 0 kmeans 0 

Decisions if the data set is clusterable or not - kmeans++  
clu if correct  0  clu if not correct  0  not clu if   correct  100  not clu if not correct  0 

 }%\Bem{ 
 \hline   DNase & g/2 & 2 & 0 &  0  &  0 & 0 & 0 & 100 & 0 \\ 
\Bem{

 phase  g/2  pfrak 0.1 ---- sufficient gap quot  0.058973060967884 

 sufficient gap quot  0.415327826479856 
  10         20         30         40         50         60         70         80         90         100       
Errors in recognizing the intrinsic clustering 
kmeans++ 0 kmeans 0 

Decisions if the data set is clusterable or not - kmeans++  
clu if correct  0  clu if not correct  0  not clu if   correct  100  not clu if not correct  0 

 }%\Bem{ 
 \hline   DNase & g/2 & 2 & 0.1 &  0  &  0 & 0 & 0 & 100 & 0 \\ 
\Bem{

 phase  g  pfrak 0 ---- sufficient gap quot  0.0472230819828586 

 sufficient gap quot  1 
  10         20         30         40         50         60         70         80         90         100       
Errors in recognizing the intrinsic clustering 
kmeans++ 0 kmeans 0 

Decisions if the data set is clusterable or not - kmeans++  
clu if correct  100  clu if not correct  0  not clu if   correct  0  not clu if not correct  0 

 }%\Bem{ 
 \hline   DNase & g & 2 & 0 &  0  &  0 & 100 & 0 & 0 & 0 \\ 
\Bem{

 phase  g  pfrak 0.1 ---- sufficient gap quot  0.058973060967884 

 sufficient gap quot  1 
  10         20         30         40         50         60         70         80         90         100       
Errors in recognizing the intrinsic clustering 
kmeans++ 0 kmeans 0 

Decisions if the data set is clusterable or not - kmeans++  
clu if correct  100  clu if not correct  0  not clu if   correct  0  not clu if not correct  0 

 }%\Bem{ 
 \hline   DNase & g & 2 & 0.1 &  0  &  0 & 100 & 0 & 0 & 0 \\ 
\Bem{

 phase  2g  pfrak 0 ---- sufficient gap quot  0.0472230819828586 

 sufficient gap quot  2.1956944551949 
  10         20         30         40         50         60         70         80         90         100       
Errors in recognizing the intrinsic clustering 
kmeans++ 0 kmeans 0 

Decisions if the data set is clusterable or not - kmeans++  
clu if correct  100  clu if not correct  0  not clu if   correct  0  not clu if not correct  0 

 }%\Bem{ 
 \hline   DNase & 2g & 2 & 0 &  0  &  0 & 100 & 0 & 0 & 0 \\ 
\Bem{

 phase  2g  pfrak 0.1 ---- sufficient gap quot  0.058973060967884 

 sufficient gap quot  2.16934434704029 
  10         20         30         40         50         60         70         80         90         100       
Errors in recognizing the intrinsic clustering 
kmeans++ 0 kmeans 0 

Decisions if the data set is clusterable or not - kmeans++  
clu if correct  100  clu if not correct  0  not clu if   correct  0  not clu if not correct  0 

 }%\Bem{ 
 \hline   DNase & 2g & 2 & 0.1 &  0  &  0 & 100 & 0 & 0 & 0 \\ 
\Bem{ 

------------------------------------------
The set to analyse DNase
 The numerical columns [1] "conc"    "density"

 The number of rows   176
 The sizes of ground truth clusters  [1] 110  22  44
\n }%\Bem{
\Bem{

 phase  orig  pfrak 0 ---- sufficient gap quot  0.0141676868696718 

 sufficient gap quot  0.0112778684315601 
  10         20         30         40         50         60         70         80         90         100       
Errors in recognizing the intrinsic clustering 
kmeans++ 25 kmeans 81 

Decisions if the data set is clusterable or not - kmeans++  
clu if correct  0  clu if not correct  0  not clu if   correct  75  not clu if not correct  25 

 }%\Bem{ 
 \hline   DNase & orig & 3 & 0 &  81  &  25 & 0 & 0 & 75 & 25 \\ 
\Bem{

 phase  orig  pfrak 0.1 ---- sufficient gap quot  0.0195523414637491 

 sufficient gap quot  0.0132190591635998 
  10         20         30         40         50         60         70         80         90         100       
Errors in recognizing the intrinsic clustering 
kmeans++ 28 kmeans 80 

Decisions if the data set is clusterable or not - kmeans++  
clu if correct  0  clu if not correct  0  not clu if   correct  72  not clu if not correct  28 

 }%\Bem{ 
 \hline   DNase & orig & 3 & 0.1 &  80  &  28 & 0 & 0 & 72 & 28 \\ 
\Bem{

 phase  g/2  pfrak 0 ---- sufficient gap quot  0.0141676868696718 

 sufficient gap quot  0.471986857820362 
  10         20         30         40         50         60         70         80         90         100       
Errors in recognizing the intrinsic clustering 
kmeans++ 1 kmeans 63 

Decisions if the data set is clusterable or not - kmeans++  
clu if correct  0  clu if not correct  0  not clu if   correct  99  not clu if not correct  1 

 }%\Bem{ 
 \hline   DNase & g/2 & 3 & 0 &  63  &  1 & 0 & 0 & 99 & 1 \\ 
\Bem{

 phase  g/2  pfrak 0.1 ---- sufficient gap quot  0.0195523414637491 

 sufficient gap quot  0.475174296821998 
  10         20         30         40         50         60         70         80         90         100       
Errors in recognizing the intrinsic clustering 
kmeans++ 1 kmeans 59 

Decisions if the data set is clusterable or not - kmeans++  
clu if correct  0  clu if not correct  0  not clu if   correct  99  not clu if not correct  1 

 }%\Bem{ 
 \hline   DNase & g/2 & 3 & 0.1 &  59  &  1 & 0 & 0 & 99 & 1 \\ 
\Bem{

 phase  g  pfrak 0 ---- sufficient gap quot  0.0141676868696718 

 sufficient gap quot  1.00000000000001 
  10         20         30         40         50         60         70         80         90         100       
Errors in recognizing the intrinsic clustering 
kmeans++ 0 kmeans 58 

Decisions if the data set is clusterable or not - kmeans++  
clu if correct  100  clu if not correct  0  not clu if   correct  0  not clu if not correct  0 

 }%\Bem{ 
 \hline   DNase & g & 3 & 0 &  58  &  0 & 100 & 0 & 0 & 0 \\ 
\Bem{

 phase  g  pfrak 0.1 ---- sufficient gap quot  0.0195523414637491 

 sufficient gap quot  1 
  10         20         30         40         50         60         70         80         90         100       
Errors in recognizing the intrinsic clustering 
kmeans++ 0 kmeans 51 

Decisions if the data set is clusterable or not - kmeans++  
clu if correct  100  clu if not correct  0  not clu if   correct  0  not clu if not correct  0 

 }%\Bem{ 
 \hline   DNase & g & 3 & 0.1 &  51  &  0 & 100 & 0 & 0 & 0 \\ 
\Bem{

 phase  2g  pfrak 0 ---- sufficient gap quot  0.0141676868696718 

 sufficient gap quot  2.05602628435926 
  10         20         30         40         50         60         70         80         90         100       
Errors in recognizing the intrinsic clustering 
kmeans++ 0 kmeans 50 

Decisions if the data set is clusterable or not - kmeans++  
clu if correct  100  clu if not correct  0  not clu if   correct  0  not clu if not correct  0 

 }%\Bem{ 
 \hline   DNase & 2g & 3 & 0 &  50  &  0 & 100 & 0 & 0 & 0 \\ 
\Bem{

 phase  2g  pfrak 0.1 ---- sufficient gap quot  0.0195523414637491 

 sufficient gap quot  2.04965140635595 
  10         20         30         40         50         60         70         80         90         100       
Errors in recognizing the intrinsic clustering 
kmeans++ 0 kmeans 49 

Decisions if the data set is clusterable or not - kmeans++  
clu if correct  100  clu if not correct  0  not clu if   correct  0  not clu if not correct  0 

 }%\Bem{ 
 \hline   DNase & 2g & 3 & 0.1 &  49  &  0 & 100 & 0 & 0 & 0 \\ 
\Bem{ 

------------------------------------------
The set to analyse DNase
 The numerical columns [1] "conc"    "density"

 The number of rows   176
 The sizes of ground truth clusters  [1] 22 66 22 22 44
\n }%\Bem{
\Bem{

 phase  orig  pfrak 0 ---- sufficient gap quot  0.00380107934075923 

 sufficient gap quot  0.00380107934075923 
  10         20         30         40         50         60         70         80         90         100       
Errors in recognizing the intrinsic clustering 
kmeans++ 35 kmeans 75 

Decisions if the data set is clusterable or not - kmeans++  
clu if correct  0  clu if not correct  0  not clu if   correct  65  not clu if not correct  35 

 }%\Bem{ 
 \hline   DNase & orig & 5 & 0 &  75  &  35 & 0 & 0 & 65 & 35 \\ 
\Bem{

 phase  orig  pfrak 0.1 ---- sufficient gap quot  0.00400912645257774 

 sufficient gap quot  0.00400912645257774 
  10         20         30         40         50         60         70         80         90         100       
Errors in recognizing the intrinsic clustering 
kmeans++ 30 kmeans 79 

Decisions if the data set is clusterable or not - kmeans++  
clu if correct  0  clu if not correct  0  not clu if   correct  70  not clu if not correct  30 

 }%\Bem{ 
 \hline   DNase & orig & 5 & 0.1 &  79  &  30 & 0 & 0 & 70 & 30 \\ 
\Bem{

 phase  g/2  pfrak 0 ---- sufficient gap quot  0.00380107934075923 

 sufficient gap quot  0.493036744642702 
  10         20         30         40         50         60         70         80         90         100       
Errors in recognizing the intrinsic clustering 
kmeans++ 0 kmeans 84 

Decisions if the data set is clusterable or not - kmeans++  
clu if correct  0  clu if not correct  0  not clu if   correct  100  not clu if not correct  0 

 }%\Bem{ 
 \hline   DNase & g/2 & 5 & 0 &  84  &  0 & 0 & 0 & 100 & 0 \\ 
\Bem{

 phase  g/2  pfrak 0.1 ---- sufficient gap quot  0.00400912645257774 

 sufficient gap quot  0.4938195159537 
  10         20         30         40         50         60         70         80         90         100       
Errors in recognizing the intrinsic clustering 
kmeans++ 0 kmeans 86 

Decisions if the data set is clusterable or not - kmeans++  
clu if correct  0  clu if not correct  0  not clu if   correct  100  not clu if not correct  0 

 }%\Bem{ 
 \hline   DNase & g/2 & 5 & 0.1 &  86  &  0 & 0 & 0 & 100 & 0 \\ 
\Bem{

 phase  g  pfrak 0 ---- sufficient gap quot  0.00380107934075923 

 sufficient gap quot  0.999999999999996 
  10         20         30         40         50         60         70         80         90         100       
Errors in recognizing the intrinsic clustering 
kmeans++ 0 kmeans 80 

Decisions if the data set is clusterable or not - kmeans++  
clu if correct  100  clu if not correct  0  not clu if   correct  0  not clu if not correct  0 

 }%\Bem{ 
 \hline   DNase & g & 5 & 0 &  80  &  0 & 100 & 0 & 0 & 0 \\ 
\Bem{

 phase  g  pfrak 0.1 ---- sufficient gap quot  0.00400912645257774 

 sufficient gap quot  0.999999999999923 
  10         20         30         40         50         60         70         80         90         100       
Errors in recognizing the intrinsic clustering 
kmeans++ 0 kmeans 79 

Decisions if the data set is clusterable or not - kmeans++  
clu if correct  100  clu if not correct  0  not clu if   correct  0  not clu if not correct  0 

 }%\Bem{ 
 \hline   DNase & g & 5 & 0.1 &  79  &  0 & 100 & 0 & 0 & 0 \\ 
\Bem{

 phase  2g  pfrak 0 ---- sufficient gap quot  0.00380107934075923 

 sufficient gap quot  2.01392651071466 
  10         20         30         40         50         60         70         80         90         100       
Errors in recognizing the intrinsic clustering 
kmeans++ 0 kmeans 83 

Decisions if the data set is clusterable or not - kmeans++  
clu if correct  100  clu if not correct  0  not clu if   correct  0  not clu if not correct  0 

 }%\Bem{ 
 \hline   DNase & 2g & 5 & 0 &  83  &  0 & 100 & 0 & 0 & 0 \\ 
\Bem{

 phase  2g  pfrak 0.1 ---- sufficient gap quot  0.00400912645257774 

 sufficient gap quot  2.01236096809262 
  10         20         30         40         50         60         70         80         90         100       
Errors in recognizing the intrinsic clustering 
kmeans++ 0 kmeans 81 

Decisions if the data set is clusterable or not - kmeans++  
clu if correct  100  clu if not correct  0  not clu if   correct  0  not clu if not correct  0 

 }%\Bem{ 
 \hline   DNase & 2g & 5 & 0.1 &  81  &  0 & 100 & 0 & 0 & 0 \\ 
\Bem{ 

------------------------------------------
The set to analyse iris
 The numerical columns [1] "Sepal.Length" "Sepal.Width"  "Petal.Length" "Petal.Width" 

 The number of rows   150
 The sizes of ground truth clusters  [1] 100  50
\n }%\Bem{
\Bem{

 phase  orig  pfrak 0 ---- sufficient gap quot  -0.0911315934039788 

 sufficient gap quot  -0.0911315934039788 
  10         20         30         40         50         60         70         80         90         100       
Errors in recognizing the intrinsic clustering 
kmeans++ 0 kmeans 0 

Decisions if the data set is clusterable or not - kmeans++  
clu if correct  0  clu if not correct  0  not clu if   correct  100  not clu if not correct  0 

 }%\Bem{ 
 \hline   iris & orig & 2 & 0 &  0  &  0 & 0 & 0 & 100 & 0 \\ 
\Bem{

 phase  orig  pfrak 0.1 ---- sufficient gap quot  -0.0821858847599162 

 sufficient gap quot  -0.0821858847599162 
  10         20         30         40         50         60         70         80         90         100       
Errors in recognizing the intrinsic clustering 
kmeans++ 0 kmeans 0 

Decisions if the data set is clusterable or not - kmeans++  
clu if correct  0  clu if not correct  0  not clu if   correct  100  not clu if not correct  0 

 }%\Bem{ 
 \hline   iris & orig & 2 & 0.1 &  0  &  0 & 0 & 0 & 100 & 0 \\ 
\Bem{

 phase  g/2  pfrak 0 ---- sufficient gap quot  -0.0911315934039788 

 sufficient gap quot  0.389763141982924 
  10         20         30         40         50         60         70         80         90         100       
Errors in recognizing the intrinsic clustering 
kmeans++ 0 kmeans 0 

Decisions if the data set is clusterable or not - kmeans++  
clu if correct  0  clu if not correct  0  not clu if   correct  100  not clu if not correct  0 

 }%\Bem{ 
 \hline   iris & g/2 & 2 & 0 &  0  &  0 & 0 & 0 & 100 & 0 \\ 
\Bem{

 phase  g/2  pfrak 0.1 ---- sufficient gap quot  -0.0821858847599162 

 sufficient gap quot  0.399385450847417 
  10         20         30         40         50         60         70         80         90         100       
Errors in recognizing the intrinsic clustering 
kmeans++ 0 kmeans 0 

Decisions if the data set is clusterable or not - kmeans++  
clu if correct  0  clu if not correct  0  not clu if   correct  100  not clu if not correct  0 

 }%\Bem{ 
 \hline   iris & g/2 & 2 & 0.1 &  0  &  0 & 0 & 0 & 100 & 0 \\ 
\Bem{

 phase  g  pfrak 0 ---- sufficient gap quot  -0.0911315934039788 

 sufficient gap quot  0.999999999999998 
  10         20         30         40         50         60         70         80         90         100       
Errors in recognizing the intrinsic clustering 
kmeans++ 0 kmeans 0 

Decisions if the data set is clusterable or not - kmeans++  
clu if correct  100  clu if not correct  0  not clu if   correct  0  not clu if not correct  0 

 }%\Bem{ 
 \hline   iris & g & 2 & 0 &  0  &  0 & 100 & 0 & 0 & 0 \\ 
\Bem{

 phase  g  pfrak 0.1 ---- sufficient gap quot  -0.0821858847599162 

 sufficient gap quot  0.999999999999999 
  10         20         30         40         50         60         70         80         90         100       
Errors in recognizing the intrinsic clustering 
kmeans++ 0 kmeans 0 

Decisions if the data set is clusterable or not - kmeans++  
clu if correct  100  clu if not correct  0  not clu if   correct  0  not clu if not correct  0 

 }%\Bem{ 
 \hline   iris & g & 2 & 0.1 &  0  &  0 & 100 & 0 & 0 & 0 \\ 
\Bem{

 phase  2g  pfrak 0 ---- sufficient gap quot  -0.0911315934039788 

 sufficient gap quot  2.22047371603415 
  10         20         30         40         50         60         70         80         90         100       
Errors in recognizing the intrinsic clustering 
kmeans++ 0 kmeans 0 

Decisions if the data set is clusterable or not - kmeans++  
clu if correct  100  clu if not correct  0  not clu if   correct  0  not clu if not correct  0 

 }%\Bem{ 
 \hline   iris & 2g & 2 & 0 &  0  &  0 & 100 & 0 & 0 & 0 \\ 
\Bem{

 phase  2g  pfrak 0.1 ---- sufficient gap quot  -0.0821858847599162 

 sufficient gap quot  2.20122909830517 
  10         20         30         40         50         60         70         80         90         100       
Errors in recognizing the intrinsic clustering 
kmeans++ 0 kmeans 0 

Decisions if the data set is clusterable or not - kmeans++  
clu if correct  100  clu if not correct  0  not clu if   correct  0  not clu if not correct  0 

 }%\Bem{ 
 \hline   iris & 2g & 2 & 0.1 &  0  &  0 & 100 & 0 & 0 & 0 \\ 
\Bem{ 

------------------------------------------
The set to analyse iris
 The numerical columns [1] "Sepal.Length" "Sepal.Width"  "Petal.Length" "Petal.Width" 

 The number of rows   150
 The sizes of ground truth clusters  [1] 53 50 47
\n }%\Bem{
\Bem{

 phase  orig  pfrak 0 ---- sufficient gap quot  -0.0421982929531296 

 sufficient gap quot  -0.0421982929531296 
  10         20         30         40         50         60         70         80         90         100       
Errors in recognizing the intrinsic clustering 
kmeans++ 17 kmeans 19 

Decisions if the data set is clusterable or not - kmeans++  
clu if correct  0  clu if not correct  0  not clu if   correct  83  not clu if not correct  17 

 }%\Bem{ 
 \hline   iris & orig & 3 & 0 &  19  &  17 & 0 & 0 & 83 & 17 \\ 
\Bem{

 phase  orig  pfrak 0.1 ---- sufficient gap quot  -0.0298356210274633 

 sufficient gap quot  -0.0232119752572255 
  10         20         30         40         50         60         70         80         90         100       
Errors in recognizing the intrinsic clustering 
kmeans++ 15 kmeans 23 

Decisions if the data set is clusterable or not - kmeans++  
clu if correct  0  clu if not correct  0  not clu if   correct  85  not clu if not correct  15 

 }%\Bem{ 
 \hline   iris & orig & 3 & 0.1 &  23  &  15 & 0 & 0 & 85 & 15 \\ 
\Bem{

 phase  g/2  pfrak 0 ---- sufficient gap quot  -0.0421982929531296 

 sufficient gap quot  0.462382190867003 
  10         20         30         40         50         60         70         80         90         100       
Errors in recognizing the intrinsic clustering 
kmeans++ 0 kmeans 35 

Decisions if the data set is clusterable or not - kmeans++  
clu if correct  0  clu if not correct  0  not clu if   correct  100  not clu if not correct  0 

 }%\Bem{ 
 \hline   iris & g/2 & 3 & 0 &  35  &  0 & 0 & 0 & 100 & 0 \\
\hline 
\Bem{

 phase  g/2  pfrak 0.1 ---- sufficient gap quot  -0.0298356210274633 

 sufficient gap quot  0.47010955859384 
 10         20         30         40         50         60         70       
 }% Bem
%DNase & original &2& 0 & 73 & 32  & 0  & 0  &  68 &  32  \\
 \hline  
\end{tabular}  
\end{center}  
\end{table} 
}%Bem 

%\IpiArchiv{
\begin{longtable}{|l|r|r|r|r|r|r|r|r|r|} 
\caption{Reconstruction of the clusters in real data from $R$ library $datasets$. Notation in text.   }
\label{tab:real}
\\
\hline
Dataset & gap type& clusters &$\mathfrak{p}$ &
\multicolumn{2}{|c|}{Errors}& 
\multicolumn{2}{|c|}{WC disc.}& 
\multicolumn{2}{|c|}{WC not disc. }
\\
\cline{5-10}
&&&& $k$-means   & $k$-means++ 
& cc 
& wc
& cc  
& wc 
\\
 \hline  
%\Bem{%super
 %\hline 
  iris & g/2 & 3 & 0.1 &  40  &  0 & 0 & 0 & 100 & 0 \\ 
\Bem{

 phase  g  pfrak 0 ---- sufficient gap quot  -0.0421982929531296 

 sufficient gap quot  0.999999999999997 
  10         20         30         40         50         60         70         80         90         100       
Errors in recognizing the intrinsic clustering 
kmeans++ 0 kmeans 40 

Decisions if the data set is clusterable or not - kmeans++  
clu if correct  100  clu if not correct  0  not clu if   correct  0  not clu if not correct  0 

 }%\Bem{ 
 \hline   iris & g & 3 & 0 &  40  &  0 & 100 & 0 & 0 & 0 \\ 
\Bem{

 phase  g  pfrak 0.1 ---- sufficient gap quot  -0.0298356210274633 

 sufficient gap quot  0.999999999999997 
  10         20         30         40         50         60         70         80         90         100       
Errors in recognizing the intrinsic clustering 
kmeans++ 0 kmeans 24 

Decisions if the data set is clusterable or not - kmeans++  
clu if correct  100  clu if not correct  0  not clu if   correct  0  not clu if not correct  0 

 }%\Bem{ 
 \hline   iris & g & 3 & 0.1 &  24  &  0 & 100 & 0 & 0 & 0 \\ 
\Bem{

 phase  2g  pfrak 0 ---- sufficient gap quot  -0.0421982929531296 

 sufficient gap quot  2.07523561826599 
  10         20         30         40         50         60         70         80         90         100       
Errors in recognizing the intrinsic clustering 
kmeans++ 0 kmeans 36 

Decisions if the data set is clusterable or not - kmeans++  
clu if correct  100  clu if not correct  0  not clu if   correct  0  not clu if not correct  0 

 }%\Bem{ 
 \hline   iris & 2g & 3 & 0 &  36  &  0 & 100 & 0 & 0 & 0 \\ 
\Bem{

 phase  2g  pfrak 0.1 ---- sufficient gap quot  -0.0298356210274633 

 sufficient gap quot  2.05978088281231 
  10         20         30         40         50         60         70         80         90         100       
Errors in recognizing the intrinsic clustering 
kmeans++ 0 kmeans 31 

Decisions if the data set is clusterable or not - kmeans++  
clu if correct  100  clu if not correct  0  not clu if   correct  0  not clu if not correct  0 

 }%\Bem{ 
 \hline   iris & 2g & 3 & 0.1 &  31  &  0 & 100 & 0 & 0 & 0 \\ 
\Bem{ 

------------------------------------------
The set to analyse iris
 The numerical columns [1] "Sepal.Length" "Sepal.Width"  "Petal.Length" "Petal.Width" 

 The number of rows   150
 The sizes of ground truth clusters  [1] 48 25 29 23 25
\n }%\Bem{
\Bem{

 phase  orig  pfrak 0 ---- sufficient gap quot  -0.0140192981240695 

 sufficient gap quot  -0.00646079055308779 
  10         20         30         40         50         60         70         80         90         100       
Errors in recognizing the intrinsic clustering 
kmeans++ 97 kmeans 94 

Decisions if the data set is clusterable or not - kmeans++  
clu if correct  0  clu if not correct  0  not clu if   correct  3  not clu if not correct  97 

 }%\Bem{ 
 \hline   iris & orig & 5 & 0 &  94  &  97 & 0 & 0 & 3 & 97 \\ 
\Bem{

 phase  orig  pfrak 0.1 ---- sufficient gap quot  -0.0127556666397006 

 sufficient gap quot  -0.00689588136388553 
  10         20         30         40         50         60         70         80         90         100       
Errors in recognizing the intrinsic clustering 
kmeans++ 95 kmeans 96 

Decisions if the data set is clusterable or not - kmeans++  
clu if correct  0  clu if not correct  0  not clu if   correct  5  not clu if not correct  95 

 }%\Bem{ 
 \hline   iris & orig & 5 & 0.1 &  96  &  95 & 0 & 0 & 5 & 95 \\ 
\Bem{

 phase  g/2  pfrak 0 ---- sufficient gap quot  -0.0140192981240695 

 sufficient gap quot  0.486489880171282 
  10         20         30         40         50         60         70         80         90         100       
Errors in recognizing the intrinsic clustering 
kmeans++ 0 kmeans 69 

Decisions if the data set is clusterable or not - kmeans++  
clu if correct  0  clu if not correct  0  not clu if   correct  100  not clu if not correct  0 

 }%\Bem{ 
 \hline   iris & g/2 & 5 & 0 &  69  &  0 & 0 & 0 & 100 & 0 \\ 
\Bem{

 phase  g/2  pfrak 0.1 ---- sufficient gap quot  -0.0127556666397006 

 sufficient gap quot  0.487215155323312 
  10         20         30         40         50         60         70         80         90         100       
Errors in recognizing the intrinsic clustering 
kmeans++ 0 kmeans 72 

Decisions if the data set is clusterable or not - kmeans++  
clu if correct  0  clu if not correct  0  not clu if   correct  100  not clu if not correct  0 

 }%\Bem{ 
 \hline   iris & g/2 & 5 & 0.1 &  72  &  0 & 0 & 0 & 100 & 0 \\ 
\Bem{

 phase  g  pfrak 0 ---- sufficient gap quot  -0.0140192981240695 

 sufficient gap quot  0.98936213934499 
  10         20         30         40         50         60         70         80         90         100       
Errors in recognizing the intrinsic clustering 
kmeans++ 0 kmeans 71 

Decisions if the data set is clusterable or not - kmeans++  
clu if correct  0  clu if not correct  0  not clu if   correct  100  not clu if not correct  0 

 }%\Bem{ 
 \hline   iris & g & 5 & 0 &  71  &  0 & 0 & 0 & 100 & 0 \\ 
\Bem{

 phase  g  pfrak 0.1 ---- sufficient gap quot  -0.0127556666397006 

 sufficient gap quot  0.989065529352742 
  10         20         30         40         50         60         70         80         90         100       
Errors in recognizing the intrinsic clustering 
kmeans++ 0 kmeans 68 

Decisions if the data set is clusterable or not - kmeans++  
clu if correct  0  clu if not correct  0  not clu if   correct  100  not clu if not correct  0 

 }%\Bem{ 
 \hline   iris & g & 5 & 0.1 &  68  &  0 & 0 & 0 & 100 & 0 \\ 
\Bem{

 phase  2g  pfrak 0 ---- sufficient gap quot  -0.0140192981240695 

 sufficient gap quot  1.99510665769241 
  10         20         30         40         50         60         70         80         90         100       
Errors in recognizing the intrinsic clustering 
kmeans++ 0 kmeans 72 

Decisions if the data set is clusterable or not - kmeans++  
clu if correct  100  clu if not correct  0  not clu if   correct  0  not clu if not correct  0 

 }%\Bem{ 
 \hline   iris & 2g & 5 & 0 &  72  &  0 & 100 & 0 & 0 & 0 \\ 
\Bem{

 phase  2g  pfrak 0.1 ---- sufficient gap quot  -0.0127556666397006 

 sufficient gap quot  1.99276627741158 
  10         20         30         40         50         60         70         80         90         100       
Errors in recognizing the intrinsic clustering 
kmeans++ 0 kmeans 71 

Decisions if the data set is clusterable or not - kmeans++  
clu if correct  100  clu if not correct  0  not clu if   correct  0  not clu if not correct  0 

 }%\Bem{ 
 \hline   iris & 2g & 5 & 0.1 &  71  &  0 & 100 & 0 & 0 & 0 \\ 
\Bem{ 

------------------------------------------
The set to analyse Theoph
 The numerical columns [1] "Wt"   "Dose" "Time" "conc"

 The number of rows   132
 The sizes of ground truth clusters  [1] 55 77
\n }%\Bem{
\Bem{

 phase  orig  pfrak 0 ---- sufficient gap quot  -0.150094436206911 

 sufficient gap quot  -0.150094436206911 
  10         20         30         40         50         60         70         80         90         100       
Errors in recognizing the intrinsic clustering 
kmeans++ 29 kmeans 8 

Decisions if the data set is clusterable or not - kmeans++  
clu if correct  0  clu if not correct  0  not clu if   correct  71  not clu if not correct  29 

 }%\Bem{ 
 \hline   Theoph & orig & 2 & 0 &  8  &  29 & 0 & 0 & 71 & 29 \\ 
\Bem{

 phase  orig  pfrak 0.1 ---- sufficient gap quot  -0.137191381605891 

 sufficient gap quot  -0.137191381605891 
  10         20         30         40         50         60         70         80         90         100       
Errors in recognizing the intrinsic clustering 
kmeans++ 17 kmeans 20 

Decisions if the data set is clusterable or not - kmeans++  
clu if correct  0  clu if not correct  0  not clu if   correct  83  not clu if not correct  17 

 }%\Bem{ 
 \hline   Theoph & orig & 2 & 0.1 &  20  &  17 & 0 & 0 & 83 & 17 \\ 
\Bem{

 phase  g/2  pfrak 0 ---- sufficient gap quot  -0.150094436206911 

 sufficient gap quot  0.383647293275971 
  10         20         30         40         50         60         70         80         90         100       
Errors in recognizing the intrinsic clustering 
kmeans++ 0 kmeans 0 

Decisions if the data set is clusterable or not - kmeans++  
clu if correct  0  clu if not correct  0  not clu if   correct  100  not clu if not correct  0 

 }%\Bem{ 
 \hline   Theoph & g/2 & 2 & 0 &  0  &  0 & 0 & 0 & 100 & 0 \\ 
\Bem{

 phase  g/2  pfrak 0.1 ---- sufficient gap quot  -0.137191381605891 

 sufficient gap quot  0.389275375187053 
  10         20         30         40         50         60         70         80         90         100       
Errors in recognizing the intrinsic clustering 
kmeans++ 0 kmeans 0 

Decisions if the data set is clusterable or not - kmeans++  
clu if correct  0  clu if not correct  0  not clu if   correct  100  not clu if not correct  0 

 }%\Bem{ 
 \hline   Theoph & g/2 & 2 & 0.1 &  0  &  0 & 0 & 0 & 100 & 0 \\ 
\Bem{

 phase  g  pfrak 0 ---- sufficient gap quot  -0.150094436206911 

 sufficient gap quot  1 
  10         20         30         40         50         60         70         80         90         100       
Errors in recognizing the intrinsic clustering 
kmeans++ 0 kmeans 0 

Decisions if the data set is clusterable or not - kmeans++  
clu if correct  100  clu if not correct  0  not clu if   correct  0  not clu if not correct  0 

 }%\Bem{ 
 \hline   Theoph & g & 2 & 0 &  0  &  0 & 100 & 0 & 0 & 0 \\ 
\Bem{

 phase  g  pfrak 0.1 ---- sufficient gap quot  -0.137191381605891 

 sufficient gap quot  0.999999999999999 
  10         20         30         40         50         60         70         80         90         100       
Errors in recognizing the intrinsic clustering 
kmeans++ 0 kmeans 0 

Decisions if the data set is clusterable or not - kmeans++  
clu if correct  100  clu if not correct  0  not clu if   correct  0  not clu if not correct  0 

 }%\Bem{ 
 \hline   Theoph & g & 2 & 0.1 &  0  &  0 & 100 & 0 & 0 & 0 \\ 
\Bem{

 phase  2g  pfrak 0 ---- sufficient gap quot  -0.150094436206911 

 sufficient gap quot  2.23270541344805 
  10         20         30         40         50         60         70         80         90         100       
Errors in recognizing the intrinsic clustering 
kmeans++ 0 kmeans 0 

Decisions if the data set is clusterable or not - kmeans++  
clu if correct  100  clu if not correct  0  not clu if   correct  0  not clu if not correct  0 

 }%\Bem{ 
 \hline   Theoph & 2g & 2 & 0 &  0  &  0 & 100 & 0 & 0 & 0 \\ 
\Bem{

 phase  2g  pfrak 0.1 ---- sufficient gap quot  -0.137191381605891 

 sufficient gap quot  2.22144924962589 
  10         20         30         40         50         60         70         80         90         100       
Errors in recognizing the intrinsic clustering 
kmeans++ 0 kmeans 0 

Decisions if the data set is clusterable or not - kmeans++  
clu if correct  100  clu if not correct  0  not clu if   correct  0  not clu if not correct  0 

 }%\Bem{ 
 \hline   Theoph & 2g & 2 & 0.1 &  0  &  0 & 100 & 0 & 0 & 0 \\ 
\Bem{ 

------------------------------------------
The set to analyse Theoph
 The numerical columns [1] "Wt"   "Dose" "Time" "conc"

 The number of rows   132
 The sizes of ground truth clusters  [1] 50 12 70
\n }%\Bem{
\Bem{

 phase  orig  pfrak 0 ---- sufficient gap quot  -0.0277200155474507 

 sufficient gap quot  -0.029994027589529 
  10         20         30         40         50         60         70         80         90         100       
Errors in recognizing the intrinsic clustering 
kmeans++ 81 kmeans 83 

Decisions if the data set is clusterable or not - kmeans++  
clu if correct  0  clu if not correct  0  not clu if   correct  19  not clu if not correct  81 

 }%\Bem{ 
 \hline   Theoph & orig & 3 & 0 &  83  &  81 & 0 & 0 & 19 & 81 \\ 
\Bem{

 phase  orig  pfrak 0.1 ---- sufficient gap quot  -0.0205868709282834 

 sufficient gap quot  -0.0396036504954812 
  10         20         30         40         50         60         70         80         90         100       
Errors in recognizing the intrinsic clustering 
kmeans++ 78 kmeans 83 

Decisions if the data set is clusterable or not - kmeans++  
clu if correct  0  clu if not correct  0  not clu if   correct  22  not clu if not correct  78 

 }%\Bem{ 
 \hline   Theoph & orig & 3 & 0.1 &  83  &  78 & 0 & 0 & 22 & 78 \\ 
\Bem{

 phase  g/2  pfrak 0 ---- sufficient gap quot  -0.0277200155474507 

 sufficient gap quot  0.473861023500863 
  10         20         30         40         50         60         70         80         90         100       
Errors in recognizing the intrinsic clustering 
kmeans++ 0 kmeans 21 

Decisions if the data set is clusterable or not - kmeans++  
clu if correct  0  clu if not correct  0  not clu if   correct  100  not clu if not correct  0 

 }%\Bem{ 
 \hline   Theoph & g/2 & 3 & 0 &  21  &  0 & 0 & 0 & 100 & 0 \\ 
\Bem{

 phase  g/2  pfrak 0.1 ---- sufficient gap quot  -0.0205868709282834 

 sufficient gap quot  0.478467866098741 
  10         20         30         40         50         60         70         80         90         100       
Errors in recognizing the intrinsic clustering 
kmeans++ 0 kmeans 32 

Decisions if the data set is clusterable or not - kmeans++  
clu if correct  0  clu if not correct  0  not clu if   correct  100  not clu if not correct  0 

 }%\Bem{ 
 \hline   Theoph & g/2 & 3 & 0.1 &  32  &  0 & 0 & 0 & 100 & 0 \\ 
\Bem{

 phase  g  pfrak 0 ---- sufficient gap quot  -0.0277200155474507 

 sufficient gap quot  1 
  10         20         30         40         50         60         70         80         90         100       
Errors in recognizing the intrinsic clustering 
kmeans++ 0 kmeans 32 

Decisions if the data set is clusterable or not - kmeans++  
clu if correct  100  clu if not correct  0  not clu if   correct  0  not clu if not correct  0 

 }%\Bem{ 
 \hline   Theoph & g & 3 & 0 &  32  &  0 & 100 & 0 & 0 & 0 \\ 
\Bem{

 phase  g  pfrak 0.1 ---- sufficient gap quot  -0.0205868709282834 

 sufficient gap quot  0.999999999999999 
  10         20         30         40         50         60         70         80         90         100       
Errors in recognizing the intrinsic clustering 
kmeans++ 0 kmeans 35 

Decisions if the data set is clusterable or not - kmeans++  
clu if correct  100  clu if not correct  0  not clu if   correct  0  not clu if not correct  0 

 }%\Bem{ 
 \hline   Theoph & g & 3 & 0.1 &  35  &  0 & 100 & 0 & 0 & 0 \\ 
\Bem{

 phase  2g  pfrak 0 ---- sufficient gap quot  -0.0277200155474507 

 sufficient gap quot  2.05227795299827 
  10         20         30         40         50         60         70         80         90         100       
Errors in recognizing the intrinsic clustering 
kmeans++ 1 kmeans 34 

Decisions if the data set is clusterable or not - kmeans++  
clu if correct  99  clu if not correct  0  not clu if   correct  0  not clu if not correct  1 

 }%\Bem{ 
 \hline   Theoph & 2g & 3 & 0 &  34  &  1 & 99 & 0 & 0 & 1 \\ 
\Bem{

 phase  2g  pfrak 0.1 ---- sufficient gap quot  -0.0205868709282834 

 sufficient gap quot  2.04306426780252 
  10         20         30         40         50         60         70         80         90         100       
Errors in recognizing the intrinsic clustering 
kmeans++ 0 kmeans 28 

Decisions if the data set is clusterable or not - kmeans++  
clu if correct  100  clu if not correct  0  not clu if   correct  0  not clu if not correct  0 

 }%\Bem{ 
 \hline   Theoph & 2g & 3 & 0.1 &  28  &  0 & 100 & 0 & 0 & 0 \\ 
\Bem{ 

------------------------------------------
The set to analyse Theoph
 The numerical columns [1] "Wt"   "Dose" "Time" "conc"

 The number of rows   132
 The sizes of ground truth clusters  [1] 30 23 12 19 48
\n }%\Bem{
\Bem{

 phase  orig  pfrak 0 ---- sufficient gap quot  -0.006221185009608 

 sufficient gap quot  -0.00612991987175502 
  10         20         30         40         50         60         70         80         90         100       
Errors in recognizing the intrinsic clustering 
kmeans++ 84 kmeans 84 

Decisions if the data set is clusterable or not - kmeans++  
clu if correct  0  clu if not correct  0  not clu if   correct  16  not clu if not correct  84 

 }%\Bem{ 
 \hline   Theoph & orig & 5 & 0 &  84  &  84 & 0 & 0 & 16 & 84 \\ 
\Bem{

 phase  orig  pfrak 0.1 ---- sufficient gap quot  -0.00474014737190679 

 sufficient gap quot  -0.0055588756766712 
  10         20         30         40         50         60         70         80         90         100       
Errors in recognizing the intrinsic clustering 
kmeans++ 80 kmeans 84 

Decisions if the data set is clusterable or not - kmeans++  
clu if correct  0  clu if not correct  0  not clu if   correct  20  not clu if not correct  80 

 }%\Bem{ 
 \hline   Theoph & orig & 5 & 0.1 &  84  &  80 & 0 & 0 & 20 & 80 \\ 
\Bem{

 phase  g/2  pfrak 0 ---- sufficient gap quot  -0.006221185009608 

 sufficient gap quot  0.490725188289394 
  10         20         30         40         50         60         70         80         90         100       
Errors in recognizing the intrinsic clustering 
kmeans++ 0 kmeans 68 

Decisions if the data set is clusterable or not - kmeans++  
clu if correct  0  clu if not correct  0  not clu if   correct  100  not clu if not correct  0 

 }%\Bem{ 
 \hline   Theoph & g/2 & 5 & 0 &  68  &  0 & 0 & 0 & 100 & 0 \\ 
\Bem{

 phase  g/2  pfrak 0.1 ---- sufficient gap quot  -0.00474014737190679 

 sufficient gap quot  0.491096106253125 
  10         20         30         40         50         60         70         80         90         100       
Errors in recognizing the intrinsic clustering 
kmeans++ 0 kmeans 73 

Decisions if the data set is clusterable or not - kmeans++  
clu if correct  0  clu if not correct  0  not clu if   correct  100  not clu if not correct  0 

 }%\Bem{ 
 \hline   Theoph & g/2 & 5 & 0.1 &  73  &  0 & 0 & 0 & 100 & 0 \\ 
\Bem{

 phase  g  pfrak 0 ---- sufficient gap quot  -0.006221185009608 

 sufficient gap quot  0.992609289893949 
  10         20         30         40         50         60         70         80         90         100       
Errors in recognizing the intrinsic clustering 
kmeans++ 0 kmeans 72 

Decisions if the data set is clusterable or not - kmeans++  
clu if correct  100  clu if not correct  0  not clu if   correct  0  not clu if not correct  0 

 }%\Bem{ 
 \hline   Theoph & g & 5 & 0 &  72  &  0 & 100 & 0 & 0 & 0 \\ 
\Bem{

 phase  g  pfrak 0.1 ---- sufficient gap quot  -0.00474014737190679 

 sufficient gap quot  0.991990000596154 
  10         20         30         40         50         60         70         80         90         100       
Errors in recognizing the intrinsic clustering 
kmeans++ 0 kmeans 72 

Decisions if the data set is clusterable or not - kmeans++  
clu if correct  100  clu if not correct  0  not clu if   correct  0  not clu if not correct  0 

 }%\Bem{ 
 \hline   Theoph & g & 5 & 0.1 &  72  &  0 & 100 & 0 & 0 & 0 \\ 
\Bem{

 phase  2g  pfrak 0 ---- sufficient gap quot  -0.006221185009608 

 sufficient gap quot  1.99637749310306 
  10         20         30         40         50         60         70         80         90         100       
Errors in recognizing the intrinsic clustering 
kmeans++ 0 kmeans 72 

Decisions if the data set is clusterable or not - kmeans++  
clu if correct  100  clu if not correct  0  not clu if   correct  0  not clu if not correct  0 

 }%\Bem{ 
 \hline   Theoph & 2g & 5 & 0 &  72  &  0 & 100 & 0 & 0 & 0 \\ 
\Bem{

 phase  2g  pfrak 0.1 ---- sufficient gap quot  -0.00474014737190679 

 sufficient gap quot  1.99377778928221 
  10         20         30         40         50         60         70         80         90         100       
Errors in recognizing the intrinsic clustering 
kmeans++ 0 kmeans 74 

Decisions if the data set is clusterable or not - kmeans++  
clu if correct  100  clu if not correct  0  not clu if   correct  0  not clu if not correct  0 

 }%\Bem{ 
 \hline   Theoph & 2g & 5 & 0.1 &  74  &  0 & 100 & 0 & 0 & 0 \\ 
\Bem{ 

------------------------------------------
The set to analyse attenu
 The numerical columns [1] "event" "mag"   "dist"  "accel"

 The number of rows   182
 The sizes of ground truth clusters  [1]  55 127
\n }%\Bem{
\Bem{

 phase  orig  pfrak 0 ---- sufficient gap quot  -0.155839906518548 

 sufficient gap quot  -0.155839906518548 
  10         20         30         40         50         60         70         80         90         100       
Errors in recognizing the intrinsic clustering 
kmeans++ 42 kmeans 44 

Decisions if the data set is clusterable or not - kmeans++  
clu if correct  0  clu if not correct  0  not clu if   correct  58  not clu if not correct  42 

 }%\Bem{ 
 \hline   attenu & orig & 2 & 0 &  44  &  42 & 0 & 0 & 58 & 42 \\ 
\Bem{

 phase  orig  pfrak 0.1 ---- sufficient gap quot  -0.127521900432507 

 sufficient gap quot  -0.127521900432507 
  10         20         30         40         50         60         70         80         90         100       
Errors in recognizing the intrinsic clustering 
kmeans++ 49 kmeans 41 

Decisions if the data set is clusterable or not - kmeans++  
clu if correct  0  clu if not correct  0  not clu if   correct  51  not clu if not correct  49 

 }%\Bem{ 
 \hline   attenu & orig & 2 & 0.1 &  41  &  49 & 0 & 0 & 51 & 49 \\ 
\Bem{

 phase  g/2  pfrak 0 ---- sufficient gap quot  -0.155839906518548 

 sufficient gap quot  0.387087740476297 
  10         20         30         40         50         60         70         80         90         100       
Errors in recognizing the intrinsic clustering 
kmeans++ 0 kmeans 0 

Decisions if the data set is clusterable or not - kmeans++  
clu if correct  0  clu if not correct  0  not clu if   correct  100  not clu if not correct  0 

 }%\Bem{ 
 \hline   attenu & g/2 & 2 & 0 &  0  &  0 & 0 & 0 & 100 & 0 \\ 
\Bem{

 phase  g/2  pfrak 0.1 ---- sufficient gap quot  -0.127521900432507 

 sufficient gap quot  0.40453799562228 
  10         20         30         40         50         60         70         80         90         100       
Errors in recognizing the intrinsic clustering 
kmeans++ 0 kmeans 0 

Decisions if the data set is clusterable or not - kmeans++  
clu if correct  0  clu if not correct  0  not clu if   correct  100  not clu if not correct  0 

 }%\Bem{ 
 \hline   attenu & g/2 & 2 & 0.1 &  0  &  0 & 0 & 0 & 100 & 0 \\ 
\Bem{

 phase  g  pfrak 0 ---- sufficient gap quot  -0.155839906518548 

 sufficient gap quot  1 
  10         20         30         40         50         60         70         80         90         100       
Errors in recognizing the intrinsic clustering 
kmeans++ 0 kmeans 0 

Decisions if the data set is clusterable or not - kmeans++  
clu if correct  100  clu if not correct  0  not clu if   correct  0  not clu if not correct  0 

 }%\Bem{ 
 \hline   attenu & g & 2 & 0 &  0  &  0 & 100 & 0 & 0 & 0 \\ 
\Bem{

 phase  g  pfrak 0.1 ---- sufficient gap quot  -0.127521900432507 

 sufficient gap quot  1 
  10         20         30         40         50         60         70         80         90         100       
Errors in recognizing the intrinsic clustering 
kmeans++ 0 kmeans 0 

Decisions if the data set is clusterable or not - kmeans++  
clu if correct  100  clu if not correct  0  not clu if   correct  0  not clu if not correct  0 

 }%\Bem{ 
 \hline   attenu & g & 2 & 0.1 &  0  &  0 & 100 & 0 & 0 & 0 \\ 
\Bem{

 phase  2g  pfrak 0 ---- sufficient gap quot  -0.155839906518548 

 sufficient gap quot  2.22582451904739 
  10         20         30         40         50         60         70         80         90         100       
Errors in recognizing the intrinsic clustering 
kmeans++ 0 kmeans 0 

Decisions if the data set is clusterable or not - kmeans++  
clu if correct  100  clu if not correct  0  not clu if   correct  0  not clu if not correct  0 

 }%\Bem{ 
 \hline   attenu & 2g & 2 & 0 &  0  &  0 & 100 & 0 & 0 & 0 \\ 
\Bem{

 phase  2g  pfrak 0.1 ---- sufficient gap quot  -0.127521900432507 

 sufficient gap quot  2.19092400875544 
  10         20         30         40         50         60         70         80         90         100       
Errors in recognizing the intrinsic clustering 
kmeans++ 0 kmeans 0 

Decisions if the data set is clusterable or not - kmeans++  
clu if correct  100  clu if not correct  0  not clu if   correct  0  not clu if not correct  0 

 }%\Bem{ 
 \hline   attenu & 2g & 2 & 0.1 &  0  &  0 & 100 & 0 & 0 & 0 \\ 
\Bem{ 

------------------------------------------
The set to analyse attenu
 The numerical columns [1] "event" "mag"   "dist"  "accel"

 The number of rows   182
 The sizes of ground truth clusters  [1] 81 58 43
\n }%\Bem{
\Bem{

 phase  orig  pfrak 0 ---- sufficient gap quot  -0.0487431812663528 

 sufficient gap quot  -0.0487431812663528 
  10         20         30         40         50         60         70         80         90         100       
Errors in recognizing the intrinsic clustering 
kmeans++ 61 kmeans 50 

Decisions if the data set is clusterable or not - kmeans++  
clu if correct  0  clu if not correct  0  not clu if   correct  39  not clu if not correct  61 

 }%\Bem{ 
 \hline   attenu & orig & 3 & 0 &  50  &  61 & 0 & 0 & 39 & 61 \\ 
\Bem{

 phase  orig  pfrak 0.1 ---- sufficient gap quot  -0.0442898136189204 

 sufficient gap quot  -0.0276166904046889 
  10         20         30         40         50         60         70         80         90         100       
Errors in recognizing the intrinsic clustering 
kmeans++ 58 kmeans 50 

Decisions if the data set is clusterable or not - kmeans++  
clu if correct  0  clu if not correct  0  not clu if   correct  42  not clu if not correct  58 

 }%\Bem{ 
 \hline   attenu & orig & 3 & 0.1 &  50  &  58 & 0 & 0 & 42 & 58 \\ 
\Bem{

 phase  g/2  pfrak 0 ---- sufficient gap quot  -0.0487431812663528 

 sufficient gap quot  0.308398667613849 
  10         20         30         40         50         60         70         80         90         100       
Errors in recognizing the intrinsic clustering 
kmeans++ 0 kmeans 33 

Decisions if the data set is clusterable or not - kmeans++  
clu if correct  0  clu if not correct  0  not clu if   correct  100  not clu if not correct  0 

 }%\Bem{ 
 \hline   attenu & g/2 & 3 & 0 &  33  &  0 & 0 & 0 & 100 & 0 \\ 
\Bem{

 phase  g/2  pfrak 0.1 ---- sufficient gap quot  -0.0442898136189204 

 sufficient gap quot  0.314913549291985 
  10         20         30         40         50         60         70         80         90         100       
Errors in recognizing the intrinsic clustering 
kmeans++ 2 kmeans 38 

Decisions if the data set is clusterable or not - kmeans++  
clu if correct  0  clu if not correct  0  not clu if   correct  98  not clu if not correct  2 

 }%\Bem{ 
 \hline   attenu & g/2 & 3 & 0.1 &  38  &  2 & 0 & 0 & 98 & 2 \\ 
\Bem{

 phase  g  pfrak 0 ---- sufficient gap quot  -0.0487431812663528 

 sufficient gap quot  0.677814323846153 
  10         20         30         40         50         60         70         80         90         100       
Errors in recognizing the intrinsic clustering 
kmeans++ 0 kmeans 43 

Decisions if the data set is clusterable or not - kmeans++  
clu if correct  0  clu if not correct  0  not clu if   correct  100  not clu if not correct  0 

 }%\Bem{ 
 \hline   attenu & g & 3 & 0 &  43  &  0 & 0 & 0 & 100 & 0 \\ 
\Bem{

 phase  g  pfrak 0.1 ---- sufficient gap quot  -0.0442898136189204 

 sufficient gap quot  0.681860361601568 
  10         20         30         40         50         60         70         80         90         100       
Errors in recognizing the intrinsic clustering 
kmeans++ 0 kmeans 39 

Decisions if the data set is clusterable or not - kmeans++  
clu if correct  0  clu if not correct  0  not clu if   correct  100  not clu if not correct  0 

 }%\Bem{ 
 \hline   attenu & g & 3 & 0.1 &  39  &  0 & 0 & 0 & 100 & 0 \\ 
\Bem{

 phase  2g  pfrak 0 ---- sufficient gap quot  -0.0487431812663528 

 sufficient gap quot  1.41664563631076 
  10         20         30         40         50         60         70         80         90         100       
Errors in recognizing the intrinsic clustering 
kmeans++ 0 kmeans 41 

Decisions if the data set is clusterable or not - kmeans++  
clu if correct  100  clu if not correct  0  not clu if   correct  0  not clu if not correct  0 

 }%\Bem{ 
 \hline   attenu & 2g & 3 & 0 &  41  &  0 & 100 & 0 & 0 & 0 \\ 
\Bem{

 phase  2g  pfrak 0.1 ---- sufficient gap quot  -0.0442898136189204 

 sufficient gap quot  1.41575398622075 
  10         20         30         40         50         60         70         80         90         100       
Errors in recognizing the intrinsic clustering 
kmeans++ 0 kmeans 36 

Decisions if the data set is clusterable or not - kmeans++  
clu if correct  100  clu if not correct  0  not clu if   correct  0  not clu if not correct  0 

 }%\Bem{ 
 \hline   attenu & 2g & 3 & 0.1 &  36  &  0 & 100 & 0 & 0 & 0 \\ 
\Bem{ 

------------------------------------------
The set to analyse attenu
 The numerical columns [1] "event" "mag"   "dist"  "accel"

 The number of rows   182
 The sizes of ground truth clusters  [1] 14 30 19 40 79
\n }%\Bem{
\Bem{

 phase  orig  pfrak 0 ---- sufficient gap quot  -0.00646178579854429 

 sufficient gap quot  -0.00740857761892024 
  10         20         30         40         50         60         70         80         90         100       
Errors in recognizing the intrinsic clustering 
kmeans++ 76 kmeans 80 

Decisions if the data set is clusterable or not - kmeans++  
clu if correct  0  clu if not correct  0  not clu if   correct  24  not clu if not correct  76 

 }%\Bem{ 
 \hline   attenu & orig & 5 & 0 &  80  &  76 & 0 & 0 & 24 & 76 \\ 
\Bem{

 phase  orig  pfrak 0.1 ---- sufficient gap quot  -0.0054452171787261 

 sufficient gap quot  -0.00352252536498841 
  10         20         30         40         50         60         70         80         90         100       
Errors in recognizing the intrinsic clustering 
kmeans++ 70 kmeans 79 

Decisions if the data set is clusterable or not - kmeans++  
clu if correct  0  clu if not correct  0  not clu if   correct  30  not clu if not correct  70 

 }%\Bem{ 
 \hline   attenu & orig & 5 & 0.1 &  79  &  70 & 0 & 0 & 30 & 70 \\ 
\Bem{

 phase  g/2  pfrak 0 ---- sufficient gap quot  -0.00646178579854429 

 sufficient gap quot  0.432569783366679 
  10         20         30         40         50         60         70         80         90         100       
Errors in recognizing the intrinsic clustering 
kmeans++ 0 kmeans 84 

Decisions if the data set is clusterable or not - kmeans++  
clu if correct  0  clu if not correct  0  not clu if   correct  100  not clu if not correct  0 

 }%\Bem{ 
 \hline   attenu & g/2 & 5 & 0 &  84  &  0 & 0 & 0 & 100 & 0 \\ 
\Bem{

 phase  g/2  pfrak 0.1 ---- sufficient gap quot  -0.0054452171787261 

 sufficient gap quot  0.432964620229812 
  10         20         30         40         50         60         70         80         90         100       
Errors in recognizing the intrinsic clustering 
kmeans++ 0 kmeans 89 

Decisions if the data set is clusterable or not - kmeans++  
clu if correct  0  clu if not correct  0  not clu if   correct  100  not clu if not correct  0 

 }%\Bem{ 
 \hline   attenu & g/2 & 5 & 0.1 &  89  &  0 & 0 & 0 & 100 & 0 \\ 
\Bem{

 phase  g  pfrak 0 ---- sufficient gap quot  -0.00646178579854429 

 sufficient gap quot  0.876139841714711 
  10         20         30         40         50         60         70         80         90         100       
Errors in recognizing the intrinsic clustering 
kmeans++ 0 kmeans 75 

Decisions if the data set is clusterable or not - kmeans++  
clu if correct  0  clu if not correct  0  not clu if   correct  100  not clu if not correct  0 

 }%\Bem{ 
 \hline   attenu & g & 5 & 0 &  75  &  0 & 0 & 0 & 100 & 0 \\ 
\Bem{

 phase  g  pfrak 0.1 ---- sufficient gap quot  -0.0054452171787261 

 sufficient gap quot  0.875873879565771 
  10         20         30         40         50         60         70         80         90         100       
Errors in recognizing the intrinsic clustering 
kmeans++ 0 kmeans 85 

Decisions if the data set is clusterable or not - kmeans++  
clu if correct  0  clu if not correct  0  not clu if   correct  100  not clu if not correct  0 

 }%\Bem{ 
 \hline   attenu & g & 5 & 0.1 &  85  &  0 & 0 & 0 & 100 & 0 \\ 
\Bem{

 phase  2g  pfrak 0 ---- sufficient gap quot  -0.00646178579854429 

 sufficient gap quot  1.76327995841078 
  10         20         30         40         50         60         70         80         90         100       
Errors in recognizing the intrinsic clustering 
kmeans++ 0 kmeans 82 

Decisions if the data set is clusterable or not - kmeans++  
clu if correct  100  clu if not correct  0  not clu if   correct  0  not clu if not correct  0 

 }%\Bem{ 
 \hline   attenu & 2g & 5 & 0 &  82  &  0 & 100 & 0 & 0 & 0 \\ 
\Bem{

 phase  2g  pfrak 0.1 ---- sufficient gap quot  -0.0054452171787261 

 sufficient gap quot  1.76169239823773 
  10         20         30         40         50         60         70         80         90         100       
Errors in recognizing the intrinsic clustering 
kmeans++ 0 kmeans 78 

Decisions if the data set is clusterable or not - kmeans++  
clu if correct  100  clu if not correct  0  not clu if   correct  0  not clu if not correct  0 

 }%\Bem{ 
 \hline   attenu & 2g & 5 & 0.1 &  78  &  0 & 100 & 0 & 0 & 0 \\ 
 \hline 
\Bem{ 

------------------------------------------
The set to analyse faithful
 The numerical columns [1] "eruptions" "waiting"  

 The number of rows   272
 The sizes of ground truth clusters  [1]  98 174
\n }%\Bem{
\Bem{

 phase  orig  pfrak 0 ---- sufficient gap quot  -0.00270817801835952 

 sufficient gap quot  -0.00270817801835952 
  10         20         30         40         50         60         70         80         90         100       
Errors in recognizing the intrinsic clustering 
kmeans++ 0 kmeans 0 

Decisions if the data set is clusterable or not - kmeans++  
clu if correct  0  clu if not correct  0  not clu if   correct  100  not clu if not correct  0 

 }%\Bem{ 

%\Bem{%super %
\Bem{ 

------------------------------------------
The set to analyse faithful
 The numerical columns [1] "eruptions" "waiting"  

 The number of rows   272
 The sizes of ground truth clusters  [1] 174  98
\n }%\Bem{
\Bem{

 phase  orig  pfrak 0 ---- sufficient gap quot  -0.00270817801835952 

 sufficient gap quot  -0.00270817801835952 
  10         20         30         40         50         60         70         80         90         100       
Errors in recognizing the intrinsic clustering 
kmeans++ 0 kmeans 0 

Decisions if the data set is clusterable or not - kmeans++  
clu if correct  0  clu if not correct  0  not clu if   correct  100  not clu if not correct  0 

 \hline 
 }%\Bem{ 
  faithful & orig & 2 & 0 &  0  &  0 & 0 & 0 & 100 & 0 \\ 
\Bem{

 phase  orig  pfrak 0.1 ---- sufficient gap quot  0.0195519965242614 

 sufficient gap quot  0.0195519965242614 
  10         20         30         40         50         60         70         80         90         100       
Errors in recognizing the intrinsic clustering 
kmeans++ 0 kmeans 0 

Decisions if the data set is clusterable or not - kmeans++  
clu if correct  0  clu if not correct  0  not clu if   correct  100  not clu if not correct  0 

 }%\Bem{ 
\hline 
   faithful & orig & 2 & 0.1 &  0  &  0 & 0 & 0 & 100 & 0 \\ 
\Bem{

 phase  g/2  pfrak 0 ---- sufficient gap quot  -0.00270817801835952 

 sufficient gap quot  0.380638061643645 
  10         20         30         40         50         60         70         80         90         100       
Errors in recognizing the intrinsic clustering 
kmeans++ 0 kmeans 0 

Decisions if the data set is clusterable or not - kmeans++  
clu if correct  0  clu if not correct  0  not clu if   correct  100  not clu if not correct  0 

 }%\Bem{ 
 \hline   faithful & g/2 & 2 & 0 &  0  &  0 & 0 & 0 & 100 & 0 \\ 
\Bem{

 phase  g/2  pfrak 0.1 ---- sufficient gap quot  0.0195519965242614 

 sufficient gap quot  0.39173318180355 
  10         20         30         40         50         60         70         80         90         100       
Errors in recognizing the intrinsic clustering 
kmeans++ 0 kmeans 0 

Decisions if the data set is clusterable or not - kmeans++  
clu if correct  0  clu if not correct  0  not clu if   correct  100  not clu if not correct  0 

 }%\Bem{ 
 \hline   faithful & g/2 & 2 & 0.1 &  0  &  0 & 0 & 0 & 100 & 0 \\ 
\Bem{

 phase  g  pfrak 0 ---- sufficient gap quot  -0.00270817801835952 

 sufficient gap quot  0.999999999999997 
  10         20         30         40         50         60         70         80         90         100       
Errors in recognizing the intrinsic clustering 
kmeans++ 0 kmeans 0 

Decisions if the data set is clusterable or not - kmeans++  
clu if correct  100  clu if not correct  0  not clu if   correct  0  not clu if not correct  0 

 }%\Bem{ 
 \hline   faithful & g & 2 & 0 &  0  &  0 & 100 & 0 & 0 & 0 \\ 
\Bem{

 phase  g  pfrak 0.1 ---- sufficient gap quot  0.0195519965242614 

 sufficient gap quot  1 
  10         20         30         40         50         60         70         80         90         100       
Errors in recognizing the intrinsic clustering 
kmeans++ 0 kmeans 0 

Decisions if the data set is clusterable or not - kmeans++  
clu if correct  100  clu if not correct  0  not clu if   correct  0  not clu if not correct  0 

 }%\Bem{ 
 \hline   faithful & g & 2 & 0.1 &  0  &  0 & 100 & 0 & 0 & 0 \\ 
\Bem{

 phase  2g  pfrak 0 ---- sufficient gap quot  -0.00270817801835952 

 sufficient gap quot  2.23872387671271 
  10         20         30         40         50         60         70         80         90         100       
Errors in recognizing the intrinsic clustering 
kmeans++ 0 kmeans 0 

Decisions if the data set is clusterable or not - kmeans++  
clu if correct  100  clu if not correct  0  not clu if   correct  0  not clu if not correct  0 

 }%\Bem{ 
 \hline   faithful & 2g & 2 & 0 &  0  &  0 & 100 & 0 & 0 & 0 \\ 
\Bem{

 phase  2g  pfrak 0.1 ---- sufficient gap quot  0.0195519965242614 

 sufficient gap quot  2.2165336363929 
  10         20         30         40         50         60         70         80         90         100       
Errors in recognizing the intrinsic clustering 
kmeans++ 0 kmeans 0 

Decisions if the data set is clusterable or not - kmeans++  
clu if correct  100  clu if not correct  0  not clu if   correct  0  not clu if not correct  0 

 }%\Bem{ 
 \hline   faithful & 2g & 2 & 0.1 &  0  &  0 & 100 & 0 & 0 & 0 \\ 
\Bem{ 

------------------------------------------
The set to analyse faithful
 The numerical columns [1] "eruptions" "waiting"  

 The number of rows   272
 The sizes of ground truth clusters  [1] 96 97 79
\n }%\Bem{
\Bem{

 phase  orig  pfrak 0 ---- sufficient gap quot  -0.0504779671381085 

 sufficient gap quot  -0.0504779671381085 
  10         20         30         40         50         60         70         80         90         100       
Errors in recognizing the intrinsic clustering 
kmeans++ 60 kmeans 59 

Decisions if the data set is clusterable or not - kmeans++  
clu if correct  0  clu if not correct  0  not clu if   correct  40  not clu if not correct  60 

 }%\Bem{ 
 \hline   faithful & orig & 3 & 0 &  59  &  60 & 0 & 0 & 40 & 60 \\ 
\Bem{

 phase  orig  pfrak 0.1 ---- sufficient gap quot  -0.0507772900415468 

 sufficient gap quot  -0.0507772900415468 
  10         20         30         40         50         60         70         80         90         100       
Errors in recognizing the intrinsic clustering 
kmeans++ 62 kmeans 59 

Decisions if the data set is clusterable or not - kmeans++  
clu if correct  0  clu if not correct  0  not clu if   correct  38  not clu if not correct  62 

 }%\Bem{ 
 \hline   faithful & orig & 3 & 0.1 &  59  &  62 & 0 & 0 & 38 & 62 \\ 
\Bem{

 phase  g/2  pfrak 0 ---- sufficient gap quot  -0.0504779671381085 

 sufficient gap quot  0.460697332393296 
  10         20         30         40         50         60         70         80         90         100       
Errors in recognizing the intrinsic clustering 
kmeans++ 0 kmeans 25 

Decisions if the data set is clusterable or not - kmeans++  
clu if correct  0  clu if not correct  0  not clu if   correct  100  not clu if not correct  0 

 }%\Bem{ 
 \hline   faithful & g/2 & 3 & 0 &  25  &  0 & 0 & 0 & 100 & 0 \\ 
\Bem{

 phase  g/2  pfrak 0.1 ---- sufficient gap quot  -0.0507772900415468 

 sufficient gap quot  0.458388330313244 
  10         20         30         40         50         60         70         80         90         100       
Errors in recognizing the intrinsic clustering 
kmeans++ 0 kmeans 29 

Decisions if the data set is clusterable or not - kmeans++  
clu if correct  0  clu if not correct  0  not clu if   correct  100  not clu if not correct  0 

 }%\Bem{ 
 \hline   faithful & g/2 & 3 & 0.1 &  29  &  0 & 0 & 0 & 100 & 0 \\ 
\Bem{

 phase  g  pfrak 0 ---- sufficient gap quot  -0.0504779671381085 

 sufficient gap quot  1.00000000000001 
  10         20         30         40         50         60         70         80         90         100       
Errors in recognizing the intrinsic clustering 
kmeans++ 0 kmeans 27 

Decisions if the data set is clusterable or not - kmeans++  
clu if correct  100  clu if not correct  0  not clu if   correct  0  not clu if not correct  0 

 }%\Bem{ 
 \hline   faithful & g & 3 & 0 &  27  &  0 & 100 & 0 & 0 & 0 \\ 
\Bem{

 phase  g  pfrak 0.1 ---- sufficient gap quot  -0.0507772900415468 

 sufficient gap quot  1 
  10         20         30         40         50         60         70         80         90         100       
Errors in recognizing the intrinsic clustering 
kmeans++ 0 kmeans 28 

Decisions if the data set is clusterable or not - kmeans++  
clu if correct  100  clu if not correct  0  not clu if   correct  0  not clu if not correct  0 

 }%\Bem{ 
 \hline   faithful & g & 3 & 0.1 &  28  &  0 & 100 & 0 & 0 & 0 \\ 
\Bem{

 phase  2g  pfrak 0 ---- sufficient gap quot  -0.0504779671381085 

 sufficient gap quot  2.0786053352134 
  10         20         30         40         50         60         70         80         90         100       
Errors in recognizing the intrinsic clustering 
kmeans++ 0 kmeans 29 

Decisions if the data set is clusterable or not - kmeans++  
clu if correct  100  clu if not correct  0  not clu if   correct  0  not clu if not correct  0 

 }%\Bem{ 
 \hline   faithful & 2g & 3 & 0 &  29  &  0 & 100 & 0 & 0 & 0 \\ 
\Bem{

 phase  2g  pfrak 0.1 ---- sufficient gap quot  -0.0507772900415468 

 sufficient gap quot  2.08322333937339 
  10         20         30         40         50         60         70         80         90         100       
Errors in recognizing the intrinsic clustering 
kmeans++ 0 kmeans 25 

Decisions if the data set is clusterable or not - kmeans++  
clu if correct  100  clu if not correct  0  not clu if   correct  0  not clu if not correct  0 

 }%\Bem{ 
 \hline   faithful & 2g & 3 & 0.1 &  25  &  0 & 100 & 0 & 0 & 0 \\ 
\Bem{ 

------------------------------------------
The set to analyse faithful
 The numerical columns [1] "eruptions" "waiting"  

 The number of rows   272
 The sizes of ground truth clusters  [1] 47 56 55 41 73
\n }%\Bem{
\Bem{

 phase  orig  pfrak 0 ---- sufficient gap quot  -0.0115724634675192 

 sufficient gap quot  -0.0115724634675192 
  10         20         30         40         50         60         70         80         90         100       
Errors in recognizing the intrinsic clustering 
kmeans++ 47 kmeans 72 

Decisions if the data set is clusterable or not - kmeans++  
clu if correct  0  clu if not correct  0  not clu if   correct  53  not clu if not correct  47 

 }%\Bem{ 
 \hline   faithful & orig & 5 & 0 &  72  &  47 & 0 & 0 & 53 & 47 \\ 
\Bem{

 phase  orig  pfrak 0.1 ---- sufficient gap quot  -0.00958636791762576 

 sufficient gap quot  -0.0100733745068692 
  10         20         30         40         50         60         70         80         90         100       
Errors in recognizing the intrinsic clustering 
kmeans++ 57 kmeans 65 

Decisions if the data set is clusterable or not - kmeans++  
clu if correct  0  clu if not correct  0  not clu if   correct  43  not clu if not correct  57 

 }%\Bem{ 
 \hline   faithful & orig & 5 & 0.1 &  65  &  57 & 0 & 0 & 43 & 57 \\ 
\Bem{

 phase  g/2  pfrak 0 ---- sufficient gap quot  -0.0115724634675192 

 sufficient gap quot  0.48998354816436 
  10         20         30         40         50         60         70         80         90         100       
Errors in recognizing the intrinsic clustering 
kmeans++ 0 kmeans 72 

Decisions if the data set is clusterable or not - kmeans++  
clu if correct  0  clu if not correct  0  not clu if   correct  100  not clu if not correct  0 

 }%\Bem{ 
 \hline   faithful & g/2 & 5 & 0 &  72  &  0 & 0 & 0 & 100 & 0 \\ 
\Bem{

 phase  g/2  pfrak 0.1 ---- sufficient gap quot  -0.00958636791762576 

 sufficient gap quot  0.491055454789815 
  10         20         30         40         50         60         70         80         90         100       
Errors in recognizing the intrinsic clustering 
kmeans++ 0 kmeans 77 

Decisions if the data set is clusterable or not - kmeans++  
clu if correct  0  clu if not correct  0  not clu if   correct  100  not clu if not correct  0 

 }%\Bem{ 
 \hline   faithful & g/2 & 5 & 0.1 &  77  &  0 & 0 & 0 & 100 & 0 \\ 
\Bem{

 phase  g  pfrak 0 ---- sufficient gap quot  -0.0115724634675192 

 sufficient gap quot  0.99999999999997 
  10         20         30         40         50         60         70         80         90         100       
Errors in recognizing the intrinsic clustering 
kmeans++ 0 kmeans 81 

Decisions if the data set is clusterable or not - kmeans++  
clu if correct  100  clu if not correct  0  not clu if   correct  0  not clu if not correct  0 

 }%\Bem{ 
 \hline   faithful & g & 5 & 0 &  81  &  0 & 100 & 0 & 0 & 0 \\ 
\Bem{

 phase  g  pfrak 0.1 ---- sufficient gap quot  -0.00958636791762576 

 sufficient gap quot  0.999999999999989 
  10         20         30         40         50         60         70         80         90         100       
Errors in recognizing the intrinsic clustering 
kmeans++ 0 kmeans 76 

Decisions if the data set is clusterable or not - kmeans++  
clu if correct  100  clu if not correct  0  not clu if   correct  0  not clu if not correct  0 

 }%\Bem{ 
 \hline   faithful & g & 5 & 0.1 &  76  &  0 & 100 & 0 & 0 & 0 \\ 
\Bem{

 phase  2g  pfrak 0 ---- sufficient gap quot  -0.0115724634675192 

 sufficient gap quot  2.02003290367128 
  10         20         30         40         50         60         70         80         90         100       
Errors in recognizing the intrinsic clustering 
kmeans++ 0 kmeans 77 

Decisions if the data set is clusterable or not - kmeans++  
clu if correct  100  clu if not correct  0  not clu if   correct  0  not clu if not correct  0 

 }%\Bem{ 
 \hline   faithful & 2g & 5 & 0 &  77  &  0 & 100 & 0 & 0 & 0 \\ 
\Bem{

 phase  2g  pfrak 0.1 ---- sufficient gap quot  -0.00958636791762576 

 sufficient gap quot  2.01788909042029 
  10         20         30         40         50         60         70         80         90         100       
Errors in recognizing the intrinsic clustering 
kmeans++ 0 kmeans 74 

Decisions if the data set is clusterable or not - kmeans++  
clu if correct  100  clu if not correct  0  not clu if   correct  0  not clu if not correct  0 

 }%\Bem{ 
 \hline   faithful & 2g & 5 & 0.1 &  74  &  0 & 100 & 0 & 0 & 0 \\ 
\Bem{ 

------------------------------------------
The set to analyse infert
 The numerical columns [1] "age"            "parity"         "induced"        "case"          
[5] "spontaneous"    "pooled.stratum"

 The number of rows   248
 The sizes of ground truth clusters  [1]  76 172
\n }%\Bem{
\Bem{

 phase  orig  pfrak 0 ---- sufficient gap quot  -0.144500839607118 

 sufficient gap quot  -0.138059286802988 
  10         20         30         40         50         60         70         80         90         100       
Errors in recognizing the intrinsic clustering 
kmeans++ 94 kmeans 95 

Decisions if the data set is clusterable or not - kmeans++  
clu if correct  0  clu if not correct  0  not clu if   correct  6  not clu if not correct  94 

 }%\Bem{ 
 \hline   infert & orig & 2 & 0 &  95  &  94 & 0 & 0 & 6 & 94 \\ 
\Bem{

 phase  orig  pfrak 0.1 ---- sufficient gap quot  -0.121679272201649 

 sufficient gap quot  -0.110579879401231 
  10         20         30         40         50         60         70         80         90         100       
Errors in recognizing the intrinsic clustering 
kmeans++ 94 kmeans 96 

Decisions if the data set is clusterable or not - kmeans++  
clu if correct  0  clu if not correct  0  not clu if   correct  6  not clu if not correct  94 

 }%\Bem{ 
 \hline   infert & orig & 2 & 0.1 &  96  &  94 & 0 & 0 & 6 & 94 \\ 
\Bem{

 phase  g/2  pfrak 0 ---- sufficient gap quot  -0.144500839607118 

 sufficient gap quot  0.396336495448809 
  10         20         30         40         50         60         70         80         90         100       
Errors in recognizing the intrinsic clustering 
kmeans++ 0 kmeans 0 

Decisions if the data set is clusterable or not - kmeans++  
clu if correct  0  clu if not correct  0  not clu if   correct  100  not clu if not correct  0 

 }%\Bem{ 
 \hline   infert & g/2 & 2 & 0 &  0  &  0 & 0 & 0 & 100 & 0 \\ 
\Bem{

 phase  g/2  pfrak 0.1 ---- sufficient gap quot  -0.121679272201649 

 sufficient gap quot  0.409258771512458 
  10         20         30         40         50         60         70         80         90         100       
Errors in recognizing the intrinsic clustering 
kmeans++ 0 kmeans 0 

Decisions if the data set is clusterable or not - kmeans++  
clu if correct  0  clu if not correct  0  not clu if   correct  100  not clu if not correct  0 

 }%\Bem{ 
 \hline   infert & g/2 & 2 & 0.1 &  0  &  0 & 0 & 0 & 100 & 0 \\ 
\Bem{

 phase  g  pfrak 0 ---- sufficient gap quot  -0.144500839607118 

 sufficient gap quot  1 
  10         20         30         40         50         60         70         80         90         100       
Errors in recognizing the intrinsic clustering 
kmeans++ 0 kmeans 0 

Decisions if the data set is clusterable or not - kmeans++  
clu if correct  100  clu if not correct  0  not clu if   correct  0  not clu if not correct  0 

 }%\Bem{ 
 \hline   infert & g & 2 & 0 &  0  &  0 & 100 & 0 & 0 & 0 \\ 
\Bem{

 phase  g  pfrak 0.1 ---- sufficient gap quot  -0.121679272201649 

 sufficient gap quot  1 
  10         20         30         40         50         60         70         80         90         100       
Errors in recognizing the intrinsic clustering 
kmeans++ 0 kmeans 0 

Decisions if the data set is clusterable or not - kmeans++  
clu if correct  100  clu if not correct  0  not clu if   correct  0  not clu if not correct  0 

 }%\Bem{ 
 \hline   infert & g & 2 & 0.1 &  0  &  0 & 100 & 0 & 0 & 0 \\ 
\Bem{

 phase  2g  pfrak 0 ---- sufficient gap quot  -0.144500839607118 

 sufficient gap quot  2.20732700910238 
  10         20         30         40         50         60         70         80         90         100       
Errors in recognizing the intrinsic clustering 
kmeans++ 0 kmeans 0 

Decisions if the data set is clusterable or not - kmeans++  
clu if correct  100  clu if not correct  0  not clu if   correct  0  not clu if not correct  0 

 }%\Bem{ 
 \hline   infert & 2g & 2 & 0 &  0  &  0 & 100 & 0 & 0 & 0 \\ 
\Bem{

 phase  2g  pfrak 0.1 ---- sufficient gap quot  -0.121679272201649 

 sufficient gap quot  2.18148245697507 
  10         20         30         40         50         60         70         80         90         100       
Errors in recognizing the intrinsic clustering 
kmeans++ 0 kmeans 0 

Decisions if the data set is clusterable or not - kmeans++  
clu if correct  100  clu if not correct  0  not clu if   correct  0  not clu if not correct  0 

 }%\Bem{ 
 \hline   infert & 2g & 2 & 0.1 &  0  &  0 & 100 & 0 & 0 & 0 \\ 
\Bem{ 

------------------------------------------
The set to analyse infert
 The numerical columns [1] "age"            "parity"         "induced"        "case"          
[5] "spontaneous"    "pooled.stratum"

 The number of rows   248
 The sizes of ground truth clusters  [1] 113  71  64
\n }%\Bem{
\Bem{

 phase  orig  pfrak 0 ---- sufficient gap quot  -0.0546985251160574 

 sufficient gap quot  -0.0621641571430663 
  10         20         30         40         50         60         70         80         90         100       
Errors in recognizing the intrinsic clustering 
kmeans++ 92 kmeans 90 

Decisions if the data set is clusterable or not - kmeans++  
clu if correct  0  clu if not correct  0  not clu if   correct  8  not clu if not correct  92 

 }%\Bem{ 
 \hline   infert & orig & 3 & 0 &  90  &  92 & 0 & 0 & 8 & 92 \\ 
\Bem{

 phase  orig  pfrak 0.1 ---- sufficient gap quot  -0.0477926151723594 

 sufficient gap quot  -0.039027873114514 
  10         20         30         40         50         60         70         80         90         100       
Errors in recognizing the intrinsic clustering 
kmeans++ 88 kmeans 95 

Decisions if the data set is clusterable or not - kmeans++  
clu if correct  0  clu if not correct  0  not clu if   correct  12  not clu if not correct  88 

 }%\Bem{ 
 \hline   infert & orig & 3 & 0.1 &  95  &  88 & 0 & 0 & 12 & 88 \\ 
\Bem{

 phase  g/2  pfrak 0 ---- sufficient gap quot  -0.0546985251160574 

 sufficient gap quot  0.386202915373546 
  10         20         30         40         50         60         70         80         90         100       
Errors in recognizing the intrinsic clustering 
kmeans++ 1 kmeans 34 

Decisions if the data set is clusterable or not - kmeans++  
clu if correct  0  clu if not correct  0  not clu if   correct  99  not clu if not correct  1 

 }%\Bem{ 
 \hline   infert & g/2 & 3 & 0 &  34  &  1 & 0 & 0 & 99 & 1 \\ 
\Bem{

 phase  g/2  pfrak 0.1 ---- sufficient gap quot  -0.0477926151723594 

 sufficient gap quot  0.395574924487622 
  10         20         30         40         50         60         70         80         90         100       
Errors in recognizing the intrinsic clustering 
kmeans++ 0 kmeans 31 

Decisions if the data set is clusterable or not - kmeans++  
clu if correct  0  clu if not correct  0  not clu if   correct  100  not clu if not correct  0 

 }%\Bem{ 
 \hline   infert & g/2 & 3 & 0.1 &  31  &  0 & 0 & 0 & 100 & 0 \\ 
\Bem{

 phase  g  pfrak 0 ---- sufficient gap quot  -0.0546985251160574 

 sufficient gap quot  0.849001474672417 
  10         20         30         40         50         60         70         80         90         100       
Errors in recognizing the intrinsic clustering 
kmeans++ 0 kmeans 27 

Decisions if the data set is clusterable or not - kmeans++  
clu if correct  0  clu if not correct  0  not clu if   correct  100  not clu if not correct  0 

 }%\Bem{ 
 \hline   infert & g & 3 & 0 &  27  &  0 & 0 & 0 & 100 & 0 \\ 
\Bem{

 phase  g  pfrak 0.1 ---- sufficient gap quot  -0.0477926151723594 

 sufficient gap quot  0.854564584702751 
  10         20         30         40         50         60         70         80         90         100       
Errors in recognizing the intrinsic clustering 
kmeans++ 0 kmeans 35 

Decisions if the data set is clusterable or not - kmeans++  
clu if correct  0  clu if not correct  0  not clu if   correct  100  not clu if not correct  0 

 }%\Bem{ 
 \hline   infert & g & 3 & 0.1 &  35  &  0 & 0 & 0 & 100 & 0 \\ 
\Bem{

 phase  2g  pfrak 0 ---- sufficient gap quot  -0.0546985251160574 

 sufficient gap quot  1.77459859327017 
  10         20         30         40         50         60         70         80         90         100       
Errors in recognizing the intrinsic clustering 
kmeans++ 1 kmeans 38 

Decisions if the data set is clusterable or not - kmeans++  
clu if correct  99  clu if not correct  0  not clu if   correct  0  not clu if not correct  1 

 }%\Bem{ 
 \hline   infert & 2g & 3 & 0 &  38  &  1 & 99 & 0 & 0 & 1 \\ 
\Bem{

 phase  2g  pfrak 0.1 ---- sufficient gap quot  -0.0477926151723594 

 sufficient gap quot  1.772543905133 
  10         20         30         40         50         60         70         80         90         100       
Errors in recognizing the intrinsic clustering 
kmeans++ 0 kmeans 26 

Decisions if the data set is clusterable or not - kmeans++  
clu if correct  100  clu if not correct  0  not clu if   correct  0  not clu if not correct  0 

 }%\Bem{ 
 \hline   infert & 2g & 3 & 0.1 &  26  &  0 & 100 & 0 & 0 & 0 \\ 
\Bem{ 

------------------------------------------
The set to analyse infert
 The numerical columns [1] "age"            "parity"         "induced"        "case"          
[5] "spontaneous"    "pooled.stratum"

 The number of rows   248
 The sizes of ground truth clusters  [1] 38 45 78 47 40
\n }%\Bem{
\Bem{

 phase  orig  pfrak 0 ---- sufficient gap quot  -0.0133175235070715 

 sufficient gap quot  -0.00830044248452065 
  10         20         30         40         50         60         70         80         90         100       
Errors in recognizing the intrinsic clustering 
kmeans++ 83 kmeans 82 

Decisions if the data set is clusterable or not - kmeans++  
clu if correct  0  clu if not correct  0  not clu if   correct  17  not clu if not correct  83 

 }%\Bem{ 
 \hline   infert & orig & 5 & 0 &  82  &  83 & 0 & 0 & 17 & 83 \\ 
\Bem{

 phase  orig  pfrak 0.1 ---- sufficient gap quot  -0.0110186494969437 

 sufficient gap quot  -0.0115833513318759 
  10         20         30         40         50         60         70         80         90         100       
Errors in recognizing the intrinsic clustering 
kmeans++ 92 kmeans 81 

Decisions if the data set is clusterable or not - kmeans++  
clu if correct  0  clu if not correct  0  not clu if   correct  8  not clu if not correct  92 

 }%\Bem{ 
 \hline   infert & orig & 5 & 0.1 &  81  &  92 & 0 & 0 & 8 & 92 \\ 
\Bem{

 phase  g/2  pfrak 0 ---- sufficient gap quot  -0.0133175235070715 

 sufficient gap quot  0.323250664503798 
  10         20         30         40         50         60         70         80         90         100       
Errors in recognizing the intrinsic clustering 
kmeans++ 1 kmeans 75 

Decisions if the data set is clusterable or not - kmeans++  
clu if correct  0  clu if not correct  0  not clu if   correct  99  not clu if not correct  1 

 }%\Bem{ 
 \hline   infert & g/2 & 5 & 0 &  75  &  1 & 0 & 0 & 99 & 1 \\ 
\Bem{

 phase  g/2  pfrak 0.1 ---- sufficient gap quot  -0.0110186494969437 

 sufficient gap quot  0.323704802061944 
  10         20         30         40         50         60         70         80         90         100       
Errors in recognizing the intrinsic clustering 
kmeans++ 0 kmeans 76 

Decisions if the data set is clusterable or not - kmeans++  
clu if correct  0  clu if not correct  0  not clu if   correct  100  not clu if not correct  0 

 }%\Bem{ 
 \hline   infert & g/2 & 5 & 0.1 &  76  &  0 & 0 & 0 & 100 & 0 \\ 
\Bem{

 phase  g  pfrak 0 ---- sufficient gap quot  -0.0133175235070715 

 sufficient gap quot  0.664278599256132 
  10         20         30         40         50         60         70         80         90         100       
Errors in recognizing the intrinsic clustering 
kmeans++ 0 kmeans 86 

Decisions if the data set is clusterable or not - kmeans++  
clu if correct  0  clu if not correct  0  not clu if   correct  100  not clu if not correct  0 

 }%\Bem{ 
 \hline   infert & g & 5 & 0 &  86  &  0 & 0 & 0 & 100 & 0 \\ 
\Bem{

 phase  g  pfrak 0.1 ---- sufficient gap quot  -0.0110186494969437 

 sufficient gap quot  0.663970869215573 
  10         20         30         40         50         60         70         80         90         100       
Errors in recognizing the intrinsic clustering 
kmeans++ 0 kmeans 79 

Decisions if the data set is clusterable or not - kmeans++  
clu if correct  0  clu if not correct  0  not clu if   correct  100  not clu if not correct  0 

 }%\Bem{ 
 \hline   infert & g & 5 & 0.1 &  79  &  0 & 0 & 0 & 100 & 0 \\ 
\Bem{

 phase  2g  pfrak 0 ---- sufficient gap quot  -0.0133175235070715 

 sufficient gap quot  1.34633446876081 
  10         20         30         40         50         60         70         80         90         100       
Errors in recognizing the intrinsic clustering 
kmeans++ 0 kmeans 79 

Decisions if the data set is clusterable or not - kmeans++  
clu if correct  100  clu if not correct  0  not clu if   correct  0  not clu if not correct  0 

 }%\Bem{ 
 \hline   infert & 2g & 5 & 0 &  79  &  0 & 100 & 0 & 0 & 0 \\ 
\Bem{

 phase  2g  pfrak 0.1 ---- sufficient gap quot  -0.0110186494969437 

 sufficient gap quot  1.34450300352286 
  10         20         30         40         50         60         70         80         90         100       
Errors in recognizing the intrinsic clustering 
kmeans++ 0 kmeans 71 

Decisions if the data set is clusterable or not - kmeans++  
clu if correct  100  clu if not correct  0  not clu if   correct  0  not clu if not correct  0 

 }%\Bem{ 
 \hline   infert & 2g & 5 & 0.1 &  71  &  0 & 100 & 0 & 0 & 0 \\ 
\Bem{ 

------------------------------------------
The set to analyse quakes
 The numerical columns [1] "lat"  "long" "mag" 

 The number of rows   1000
 The sizes of ground truth clusters  [1] 795 205
\n }%\Bem{
\Bem{

 phase  orig  pfrak 0 ---- sufficient gap quot  -0.13504546318426 

 sufficient gap quot  -0.13504546318426 
  10         20         30         40         50         60         70         80         90         100       
Errors in recognizing the intrinsic clustering 
kmeans++ 34 kmeans 31 

Decisions if the data set is clusterable or not - kmeans++  
clu if correct  0  clu if not correct  0  not clu if   correct  66  not clu if not correct  34 

 }%\Bem{ 
 \hline   quakes & orig & 2 & 0 &  31  &  34 & 0 & 0 & 66 & 34 \\ 
\Bem{

 phase  orig  pfrak 0.1 ---- sufficient gap quot  -0.107445469848881 

 sufficient gap quot  -0.107445469848881 
  10         20         30         40         50         60         70         80         90         100       
Errors in recognizing the intrinsic clustering 
kmeans++ 23 kmeans 29 

Decisions if the data set is clusterable or not - kmeans++  
clu if correct  0  clu if not correct  0  not clu if   correct  77  not clu if not correct  23 

 }%\Bem{ 
 \hline   quakes & orig & 2 & 0.1 &  29  &  23 & 0 & 0 & 77 & 23 \\ 
\Bem{

 phase  g/2  pfrak 0 ---- sufficient gap quot  -0.13504546318426 

 sufficient gap quot  0.399166609166306 
  10         20         30         40         50         60         70         80         90         100       
Errors in recognizing the intrinsic clustering 
kmeans++ 0 kmeans 0 

Decisions if the data set is clusterable or not - kmeans++  
clu if correct  0  clu if not correct  0  not clu if   correct  100  not clu if not correct  0 

 }%\Bem{ 
 \hline   quakes & g/2 & 2 & 0 &  0  &  0 & 0 & 0 & 100 & 0 \\ 
\Bem{

 phase  g/2  pfrak 0.1 ---- sufficient gap quot  -0.107445469848881 

 sufficient gap quot  0.409022691684821 
  10         20         30         40         50         60         70         80         90         100       
Errors in recognizing the intrinsic clustering 
kmeans++ 0 kmeans 0 

Decisions if the data set is clusterable or not - kmeans++  
clu if correct  0  clu if not correct  0  not clu if   correct  100  not clu if not correct  0 

 }%\Bem{ 
 \hline   quakes & g/2 & 2 & 0.1 &  0  &  0 & 0 & 0 & 100 & 0 \\ 
\Bem{

 phase  g  pfrak 0 ---- sufficient gap quot  -0.13504546318426 

 sufficient gap quot  1 
  10         20         30         40         50         60         70         80         90         100       
Errors in recognizing the intrinsic clustering 
kmeans++ 0 kmeans 0 

Decisions if the data set is clusterable or not - kmeans++  
clu if correct  100  clu if not correct  0  not clu if   correct  0  not clu if not correct  0 

 }%\Bem{ 
 \hline   quakes & g & 2 & 0 &  0  &  0 & 100 & 0 & 0 & 0 \\ 
\Bem{

 phase  g  pfrak 0.1 ---- sufficient gap quot  -0.107445469848881 

 sufficient gap quot  1 
  10         20         30         40         50         60         70         80         90         100       
Errors in recognizing the intrinsic clustering 
kmeans++ 0 kmeans 0 

Decisions if the data set is clusterable or not - kmeans++  
clu if correct  100  clu if not correct  0  not clu if   correct  0  not clu if not correct  0 

 }%\Bem{ 
 \hline   quakes & g & 2 & 0.1 &  0  &  0 & 100 & 0 & 0 & 0 \\ 
\Bem{

 phase  2g  pfrak 0 ---- sufficient gap quot  -0.13504546318426 

 sufficient gap quot  2.20166678166739 
  10         20         30         40         50         60         70         80         90         100       
Errors in recognizing the intrinsic clustering 
kmeans++ 0 kmeans 0 

Decisions if the data set is clusterable or not - kmeans++  
clu if correct  100  clu if not correct  0  not clu if   correct  0  not clu if not correct  0 

 }%\Bem{ 
 \hline   quakes & 2g & 2 & 0 &  0  &  0 & 100 & 0 & 0 & 0 \\ 
\Bem{

 phase  2g  pfrak 0.1 ---- sufficient gap quot  -0.107445469848881 

 sufficient gap quot  2.18195461663037 
  10         20         30         40         50         60         70         80         90         100       
Errors in recognizing the intrinsic clustering 
kmeans++ 0 kmeans 0 

Decisions if the data set is clusterable or not - kmeans++  
clu if correct  100  clu if not correct  0  not clu if   correct  0  not clu if not correct  0 

 }%\Bem{ 
 \hline   quakes & 2g & 2 & 0.1 &  0  &  0 & 100 & 0 & 0 & 0 \\ 
\Bem{ 

------------------------------------------
The set to analyse quakes
 The numerical columns [1] "lat"  "long" "mag" 

 The number of rows   1000
 The sizes of ground truth clusters  [1] 479 200 321
\n }%\Bem{
\Bem{

 phase  orig  pfrak 0 ---- sufficient gap quot  -0.0431411524201149 

 sufficient gap quot  -0.0431411524201149 
  10         20         30         40         50         60         70         80         90         100       
Errors in recognizing the intrinsic clustering 
kmeans++ 0 kmeans 0 

Decisions if the data set is clusterable or not - kmeans++  
clu if correct  0  clu if not correct  0  not clu if   correct  100  not clu if not correct  0 

 }%\Bem{ 
 \hline   quakes & orig & 3 & 0 &  0  &  0 & 0 & 0 & 100 & 0 \\ 
\Bem{

 phase  orig  pfrak 0.1 ---- sufficient gap quot  -0.0343904657883239 

 sufficient gap quot  -0.0343904657883239 
  10         20         30         40         50         60         70         80         90         100       
Errors in recognizing the intrinsic clustering 
kmeans++ 0 kmeans 0 

Decisions if the data set is clusterable or not - kmeans++  
clu if correct  0  clu if not correct  0  not clu if   correct  100  not clu if not correct  0 

 }%\Bem{ 
 \hline   quakes & orig & 3 & 0.1 &  0  &  0 & 0 & 0 & 100 & 0 \\ 
\Bem{

 phase  g/2  pfrak 0 ---- sufficient gap quot  -0.0431411524201149 

 sufficient gap quot  0.267516043145007 
  10         20         30         40         50         60         70         80         90         100       
Errors in recognizing the intrinsic clustering 
kmeans++ 0 kmeans 31 

Decisions if the data set is clusterable or not - kmeans++  
clu if correct  0  clu if not correct  0  not clu if   correct  100  not clu if not correct  0 

 }%\Bem{ 
 \hline   quakes & g/2 & 3 & 0 &  31  &  0 & 0 & 0 & 100 & 0 \\ 
\Bem{

 phase  g/2  pfrak 0.1 ---- sufficient gap quot  -0.0343904657883239 

 sufficient gap quot  0.269802036449878 
  10         20         30         40         50         60         70         80         90         100       
Errors in recognizing the intrinsic clustering 
kmeans++ 0 kmeans 35 

Decisions if the data set is clusterable or not - kmeans++  
clu if correct  0  clu if not correct  0  not clu if   correct  100  not clu if not correct  0 

 }%\Bem{ 
 \hline   quakes & g/2 & 3 & 0.1 &  35  &  0 & 0 & 0 & 100 & 0 \\ 
\Bem{

 phase  g  pfrak 0 ---- sufficient gap quot  -0.0431411524201149 

 sufficient gap quot  0.589131688174818 
  10         20         30         40         50         60         70         80         90         100       
Errors in recognizing the intrinsic clustering 
kmeans++ 0 kmeans 32 

Decisions if the data set is clusterable or not - kmeans++  
clu if correct  0  clu if not correct  0  not clu if   correct  100  not clu if not correct  0 

 }%\Bem{ 
 \hline   quakes & g & 3 & 0 &  32  &  0 & 0 & 0 & 100 & 0 \\ 
\Bem{

 phase  g  pfrak 0.1 ---- sufficient gap quot  -0.0343904657883239 

 sufficient gap quot  0.590204776216722 
  10         20         30         40         50         60         70         80         90         100       
Errors in recognizing the intrinsic clustering 
kmeans++ 0 kmeans 25 

Decisions if the data set is clusterable or not - kmeans++  
clu if correct  0  clu if not correct  0  not clu if   correct  100  not clu if not correct  0 

 }%\Bem{ 
 \hline   quakes & g & 3 & 0.1 &  25  &  0 & 0 & 0 & 100 & 0 \\ 
\Bem{

 phase  2g  pfrak 0 ---- sufficient gap quot  -0.0431411524201149 

 sufficient gap quot  1.23236297823443 
  10         20         30         40         50         60         70         80         90         100       
Errors in recognizing the intrinsic clustering 
kmeans++ 0 kmeans 33 

Decisions if the data set is clusterable or not - kmeans++  
clu if correct  100  clu if not correct  0  not clu if   correct  0  not clu if not correct  0 

 }%\Bem{ 
 \hline   quakes & 2g & 3 & 0 &  33  &  0 & 100 & 0 & 0 & 0 \\ 
\Bem{

 phase  2g  pfrak 0.1 ---- sufficient gap quot  -0.0343904657883239 

 sufficient gap quot  1.23101025575041 
  10         20         30         40         50         60         70         80         90         100       
Errors in recognizing the intrinsic clustering 
kmeans++ 0 kmeans 34 

Decisions if the data set is clusterable or not - kmeans++  
clu if correct  100  clu if not correct  0  not clu if   correct  0  not clu if not correct  0 

 }%\Bem{ 
 \hline   quakes & 2g & 3 & 0.1 &  34  &  0 & 100 & 0 & 0 & 0 \\ 
\Bem{ 

------------------------------------------
The set to analyse quakes
 The numerical columns [1] "lat"  "long" "mag" 

 The number of rows   1000
 The sizes of ground truth clusters  [1] 239 196  81 171 313
\n }%\Bem{
\Bem{

 phase  orig  pfrak 0 ---- sufficient gap quot  -0.0104600574997993 

 sufficient gap quot  -0.0104600574997993 
  10         20         30         40         50         60         70         80         90         100       
Errors in recognizing the intrinsic clustering 
kmeans++ 85 kmeans 93 

Decisions if the data set is clusterable or not - kmeans++  
clu if correct  0  clu if not correct  0  not clu if   correct  15  not clu if not correct  85 

 }%\Bem{ 
 \hline   quakes & orig & 5 & 0 &  93  &  85 & 0 & 0 & 15 & 85 \\ 
\Bem{

 phase  orig  pfrak 0.1 ---- sufficient gap quot  -0.00807844603556033 

 sufficient gap quot  -0.0106317714113256 
  10         20         30         40         50         60         70         80         90         100       
Errors in recognizing the intrinsic clustering 
kmeans++ 88 kmeans 81 

Decisions if the data set is clusterable or not - kmeans++  
clu if correct  0  clu if not correct  0  not clu if   correct  12  not clu if not correct  88 

 }%\Bem{ 
 \hline   quakes & orig & 5 & 0.1 &  81  &  88 & 0 & 0 & 12 & 88 \\ 
\Bem{

 phase  g/2  pfrak 0 ---- sufficient gap quot  -0.0104600574997993 

 sufficient gap quot  0.218836097277098 
  10         20         30         40         50         60         70         80         90         100       
Errors in recognizing the intrinsic clustering 
kmeans++ 2 kmeans 78 

Decisions if the data set is clusterable or not - kmeans++  
clu if correct  0  clu if not correct  0  not clu if   correct  98  not clu if not correct  2 

 }%\Bem{ 
 \hline   quakes & g/2 & 5 & 0 &  78  &  2 & 0 & 0 & 98 & 2 \\ 
\Bem{

 phase  g/2  pfrak 0.1 ---- sufficient gap quot  -0.00807844603556033 

 sufficient gap quot  0.255561221497479 
  10         20         30         40         50         60         70         80         90         100       
Errors in recognizing the intrinsic clustering 
kmeans++ 0 kmeans 70 

Decisions if the data set is clusterable or not - kmeans++  
clu if correct  0  clu if not correct  0  not clu if   correct  100  not clu if not correct  0 

 }%\Bem{ 
 \hline   quakes & g/2 & 5 & 0.1 &  70  &  0 & 0 & 0 & 100 & 0 \\ 
\Bem{

 phase  g  pfrak 0 ---- sufficient gap quot  -0.0104600574997993 

 sufficient gap quot  0.444997029084706 
  10         20         30         40         50         60         70         80         90         100       
Errors in recognizing the intrinsic clustering 
kmeans++ 0 kmeans 78 

Decisions if the data set is clusterable or not - kmeans++  
clu if correct  0  clu if not correct  0  not clu if   correct  100  not clu if not correct  0 

 }%\Bem{ 
 \hline   quakes & g & 5 & 0 &  78  &  0 & 0 & 0 & 100 & 0 \\ 
\Bem{

 phase  g  pfrak 0.1 ---- sufficient gap quot  -0.00807844603556033 

 sufficient gap quot  0.5177367192501 
  10         20         30         40         50         60         70         80         90         100       
Errors in recognizing the intrinsic clustering 
kmeans++ 0 kmeans 80 

Decisions if the data set is clusterable or not - kmeans++  
clu if correct  0  clu if not correct  0  not clu if   correct  100  not clu if not correct  0 

 }%\Bem{ 
 \hline   quakes & g & 5 & 0.1 &  80  &  0 & 0 & 0 & 100 & 0 \\ 
\Bem{

 phase  2g  pfrak 0 ---- sufficient gap quot  -0.0104600574997993 

 sufficient gap quot  0.897318892699918 
  10         20         30         40         50         60         70         80         90         100       
Errors in recognizing the intrinsic clustering 
kmeans++ 0 kmeans 80 

Decisions if the data set is clusterable or not - kmeans++  
clu if correct  0  clu if not correct  0  not clu if   correct  100  not clu if not correct  0 

 }%\Bem{ 
 \hline   quakes & 2g & 5 & 0 &  80  &  0 & 0 & 0 & 100 & 0 \\ 
\Bem{

 phase  2g  pfrak 0.1 ---- sufficient gap quot  -0.00807844603556033 

 sufficient gap quot  1.04208771475528 
  10         20         30         40         50         60         70         80         90         100       
Errors in recognizing the intrinsic clustering 
kmeans++ 0 kmeans 77 

Decisions if the data set is clusterable or not - kmeans++  
clu if correct  100  clu if not correct  0  not clu if   correct  0  not clu if not correct  0 

 }%\Bem{ 
 \hline   quakes & 2g & 5 & 0.1 &  77  &  0 & 100 & 0 & 0 & 0 \\  
\hline 
\Bem{ 

------------------------------------------
The set to analyse randu
 The numerical columns [1] "x" "y" "z"

 The number of rows   400
 The sizes of ground truth clusters  [1] 203 197
\n }%\Bem{
\Bem{

 phase  orig  pfrak 0 ---- sufficient gap quot  -0.164520903734925 

 sufficient gap quot  -0.165309960498347 
  10         20         30         40         50         60         70         80         90         100       
Errors in recognizing the intrinsic clustering 
kmeans++ 69 kmeans 62 

Decisions if the data set is clusterable or not - kmeans++  
clu if correct  0  clu if not correct  0  not clu if   correct  31  not clu if not correct  69 
\hline
 }%\Bem{ 
    randu & orig & 2 & 0 &  62  &  69 & 0 & 0 & 31 & 69 \\ 
\Bem{

 phase  orig  pfrak 0.1 ---- sufficient gap quot  -0.133632181962324 

 sufficient gap quot  -0.132550625077814 
  10         20         30         40         50         60         70         80         90         100       
Errors in recognizing the intrinsic clustering 
kmeans++ 64 kmeans 63 

Decisions if the data set is clusterable or not - kmeans++  
clu if correct  0  clu if not correct  0  not clu if   correct  36  not clu if not correct  64 

 }%\Bem{ 
 \hline   randu & orig & 2 & 0.1 &  63  &  64 & 0 & 0 & 36 & 64 \\ 
\Bem{

 phase  g/2  pfrak 0 ---- sufficient gap quot  -0.164520903734925 

 sufficient gap quot  0.378804582021212 
  10         20         30         40         50         60         70         80         90         100       
Errors in recognizing the intrinsic clustering 
kmeans++ 0 kmeans 0 

Decisions if the data set is clusterable or not - kmeans++  
clu if correct  0  clu if not correct  0  not clu if   correct  100  not clu if not correct  0 

 }%\Bem{ 
 \hline   randu & g/2 & 2 & 0 &  0  &  0 & 0 & 0 & 100 & 0 \\ 
\Bem{

 phase  g/2  pfrak 0.1 ---- sufficient gap quot  -0.133632181962324 

 sufficient gap quot  0.390136749301832 
  10         20         30         40         50         60         70         80         90         100       
Errors in recognizing the intrinsic clustering 
kmeans++ 0 kmeans 0 

Decisions if the data set is clusterable or not - kmeans++  
clu if correct  0  clu if not correct  0  not clu if   correct  100  not clu if not correct  0 

 }%\Bem{ 
 \hline   randu & g/2 & 2 & 0.1 &  0  &  0 & 0 & 0 & 100 & 0 \\ 
\Bem{

 phase  g  pfrak 0 ---- sufficient gap quot  -0.164520903734925 

 sufficient gap quot  0.999999999999999 
  10         20         30         40         50         60         70         80         90         100       
Errors in recognizing the intrinsic clustering 
kmeans++ 0 kmeans 0 

Decisions if the data set is clusterable or not - kmeans++  
clu if correct  100  clu if not correct  0  not clu if   correct  0  not clu if not correct  0 

 }%\Bem{ 
 \hline   randu & g & 2 & 0 &  0  &  0 & 100 & 0 & 0 & 0 \\ 
\Bem{

 phase  g  pfrak 0.1 ---- sufficient gap quot  -0.133632181962324 

 sufficient gap quot  1 
  10         20         30         40         50         60         70         80         90         100       
Errors in recognizing the intrinsic clustering 
kmeans++ 0 kmeans 0 

Decisions if the data set is clusterable or not - kmeans++  
clu if correct  100  clu if not correct  0  not clu if   correct  0  not clu if not correct  0 

 }%\Bem{ 
 \hline   randu & g & 2 & 0.1 &  0  &  0 & 100 & 0 & 0 & 0 \\ 
\Bem{

 phase  2g  pfrak 0 ---- sufficient gap quot  -0.164520903734925 

 sufficient gap quot  2.24239083595758 
  10         20         30         40         50         60         70         80         90         100       
Errors in recognizing the intrinsic clustering 
kmeans++ 0 kmeans 0 

Decisions if the data set is clusterable or not - kmeans++  
clu if correct  100  clu if not correct  0  not clu if   correct  0  not clu if not correct  0 

 }%\Bem{ 
 \hline   randu & 2g & 2 & 0 &  0  &  0 & 100 & 0 & 0 & 0 \\ 
\Bem{

 phase  2g  pfrak 0.1 ---- sufficient gap quot  -0.133632181962324 

 sufficient gap quot  2.21972650139634 
  10         20         30         40         50         60         70         80         90         100       
Errors in recognizing the intrinsic clustering 
kmeans++ 0 kmeans 0 

Decisions if the data set is clusterable or not - kmeans++  
clu if correct  100  clu if not correct  0  not clu if   correct  0  not clu if not correct  0 

 }%\Bem{ 
 \hline   randu & 2g & 2 & 0.1 &  0  &  0 & 100 & 0 & 0 & 0 \\ 
\Bem{ 

------------------------------------------
The set to analyse randu
 The numerical columns [1] "x" "y" "z"

 The number of rows   400
 The sizes of ground truth clusters  [1] 116 149 135
\n }%\Bem{
\Bem{

 phase  orig  pfrak 0 ---- sufficient gap quot  -0.0559116707326986 

 sufficient gap quot  -0.0569620250719292 
  10         20         30         40         50         60         70         80         90         100       
Errors in recognizing the intrinsic clustering 
kmeans++ 95 kmeans 93 

Decisions if the data set is clusterable or not - kmeans++  
clu if correct  0  clu if not correct  0  not clu if   correct  5  not clu if not correct  95 

 }%\Bem{ 
 \hline   randu & orig & 3 & 0 &  93  &  95 & 0 & 0 & 5 & 95 \\ 
\Bem{

 phase  orig  pfrak 0.1 ---- sufficient gap quot  -0.0441962692615901 

 sufficient gap quot  -0.0459485421185808 
  10         20         30         40         50         60         70         80         90         100       
Errors in recognizing the intrinsic clustering 
kmeans++ 93 kmeans 97 

Decisions if the data set is clusterable or not - kmeans++  
clu if correct  0  clu if not correct  0  not clu if   correct  7  not clu if not correct  93 

 }%\Bem{ 
 \hline   randu & orig & 3 & 0.1 &  97  &  93 & 0 & 0 & 7 & 93 \\ 
\Bem{

 phase  g/2  pfrak 0 ---- sufficient gap quot  -0.0559116707326986 

 sufficient gap quot  0.420751689665258 
  10         20         30         40         50         60         70         80         90         100       
Errors in recognizing the intrinsic clustering 
kmeans++ 0 kmeans 16 

Decisions if the data set is clusterable or not - kmeans++  
clu if correct  0  clu if not correct  0  not clu if   correct  100  not clu if not correct  0 

 }%\Bem{ 
 \hline   randu & g/2 & 3 & 0 &  16  &  0 & 0 & 0 & 100 & 0 \\ 
\Bem{

 phase  g/2  pfrak 0.1 ---- sufficient gap quot  -0.0441962692615901 

 sufficient gap quot  0.447024038614953 
  10         20         30         40         50         60         70         80         90         100       
Errors in recognizing the intrinsic clustering 
kmeans++ 0 kmeans 21 

Decisions if the data set is clusterable or not - kmeans++  
clu if correct  0  clu if not correct  0  not clu if   correct  100  not clu if not correct  0 

 }%\Bem{ 
 \hline   randu & g/2 & 3 & 0.1 &  21  &  0 & 0 & 0 & 100 & 0 \\ 
\Bem{

 phase  g  pfrak 0 ---- sufficient gap quot  -0.0559116707326986 

 sufficient gap quot  0.928742272362285 
  10         20         30         40         50         60         70         80         90         100       
Errors in recognizing the intrinsic clustering 
kmeans++ 0 kmeans 34 

Decisions if the data set is clusterable or not - kmeans++  
clu if correct  0  clu if not correct  0  not clu if   correct  100  not clu if not correct  0 

 }%\Bem{ 
 \hline   randu & g & 3 & 0 &  34  &  0 & 0 & 0 & 100 & 0 \\ 
\Bem{

 phase  g  pfrak 0.1 ---- sufficient gap quot  -0.0441962692615901 

 sufficient gap quot  0.977689761621097 
  10         20         30         40         50         60         70         80         90         100       
Errors in recognizing the intrinsic clustering 
kmeans++ 0 kmeans 29 

Decisions if the data set is clusterable or not - kmeans++  
clu if correct  0  clu if not correct  0  not clu if   correct  100  not clu if not correct  0 

 }%\Bem{ 
 \hline   randu & g & 3 & 0.1 &  29  &  0 & 0 & 0 & 100 & 0 \\ 
\Bem{

 phase  2g  pfrak 0 ---- sufficient gap quot  -0.0559116707326986 

 sufficient gap quot  1.94472343775633 
  10         20         30         40         50         60         70         80         90         100       
Errors in recognizing the intrinsic clustering 
kmeans++ 0 kmeans 30 

Decisions if the data set is clusterable or not - kmeans++  
clu if correct  100  clu if not correct  0  not clu if   correct  0  not clu if not correct  0 

 }%\Bem{ 
 \hline   randu & 2g & 3 & 0 &  30  &  0 & 100 & 0 & 0 & 0 \\ 
\Bem{

 phase  2g  pfrak 0.1 ---- sufficient gap quot  -0.0441962692615901 

 sufficient gap quot  2.03902120763338 
  10         20         30         40         50         60         70         80         90         100       
Errors in recognizing the intrinsic clustering 
kmeans++ 1 kmeans 41 

Decisions if the data set is clusterable or not - kmeans++  
clu if correct  99  clu if not correct  0  not clu if   correct  0  not clu if not correct  1 

 }%\Bem{ 
 \hline   randu & 2g & 3 & 0.1 &  41  &  1 & 99 & 0 & 0 & 1 \\ 
\Bem{ 

------------------------------------------
The set to analyse randu
 The numerical columns [1] "x" "y" "z"

 The number of rows   400
 The sizes of ground truth clusters  [1] 83 80 89 74 74
\n }%\Bem{
\Bem{

 phase  orig  pfrak 0 ---- sufficient gap quot  -0.0129921053705772 

 sufficient gap quot  -0.0136834659513274 
  10         20         30         40         50         60         70         80         90         100       
Errors in recognizing the intrinsic clustering 
kmeans++ 100 kmeans 99 

Decisions if the data set is clusterable or not - kmeans++  
clu if correct  0  clu if not correct  0  not clu if   correct  0  not clu if not correct  100 

 }%\Bem{ 
 \hline   randu & orig & 5 & 0 &  99  &  100 & 0 & 0 & 0 & 100 \\ 
\Bem{

 phase  orig  pfrak 0.1 ---- sufficient gap quot  -0.00987440659117326 

 sufficient gap quot  -0.00888528200551756 
  10         20         30         40         50         60         70         80         90         100       
Errors in recognizing the intrinsic clustering 
kmeans++ 98 kmeans 99 

Decisions if the data set is clusterable or not - kmeans++  
clu if correct  0  clu if not correct  0  not clu if   correct  2  not clu if not correct  98 

 }%\Bem{ 
 \hline   randu & orig & 5 & 0.1 &  99  &  98 & 0 & 0 & 2 & 98 \\ 
\Bem{

 phase  g/2  pfrak 0 ---- sufficient gap quot  -0.0129921053705772 

 sufficient gap quot  0.379854703001759 
  10         20         30         40         50         60         70         80         90         100       
Errors in recognizing the intrinsic clustering 
kmeans++ 0 kmeans 76 

Decisions if the data set is clusterable or not - kmeans++  
clu if correct  0  clu if not correct  0  not clu if   correct  100  not clu if not correct  0 

 }%\Bem{ 
 \hline   randu & g/2 & 5 & 0 &  76  &  0 & 0 & 0 & 100 & 0 \\ 
\Bem{

 phase  g/2  pfrak 0.1 ---- sufficient gap quot  -0.00987440659117326 

 sufficient gap quot  0.488441970440319 
  10         20         30         40         50         60         70         80         90         100       
Errors in recognizing the intrinsic clustering 
kmeans++ 1 kmeans 77 

Decisions if the data set is clusterable or not - kmeans++  
clu if correct  0  clu if not correct  0  not clu if   correct  99  not clu if not correct  1 

 }%\Bem{ 
 \hline   randu & g/2 & 5 & 0.1 &  77  &  1 & 0 & 0 & 99 & 1 \\ 
\Bem{

 phase  g  pfrak 0 ---- sufficient gap quot  -0.0129921053705772 

 sufficient gap quot  0.781586615853916 
  10         20         30         40         50         60         70         80         90         100       
Errors in recognizing the intrinsic clustering 
kmeans++ 0 kmeans 80 

Decisions if the data set is clusterable or not - kmeans++  
clu if correct  0  clu if not correct  0  not clu if   correct  100  not clu if not correct  0 

 }%\Bem{ 
 \hline   randu & g & 5 & 0 &  80  &  0 & 0 & 0 & 100 & 0 \\ 
\Bem{

 phase  g  pfrak 0.1 ---- sufficient gap quot  -0.00987440659117326 

 sufficient gap quot  1 
  10         20         30         40         50         60         70         80         90         100       
Errors in recognizing the intrinsic clustering 
kmeans++ 0 kmeans 71 

Decisions if the data set is clusterable or not - kmeans++  
clu if correct  100  clu if not correct  0  not clu if   correct  0  not clu if not correct  0 

 }%\Bem{ 
 \hline   randu & g & 5 & 0.1 &  71  &  0 & 100 & 0 & 0 & 0 \\ 
\Bem{

 phase  2g  pfrak 0 ---- sufficient gap quot  -0.0129921053705772 

 sufficient gap quot  1.58505044155823 
  10         20         30         40         50         60         70         80         90         100       
Errors in recognizing the intrinsic clustering 
kmeans++ 0 kmeans 81 

Decisions if the data set is clusterable or not - kmeans++  
clu if correct  100  clu if not correct  0  not clu if   correct  0  not clu if not correct  0 

 }%\Bem{ 
 \hline   randu & 2g & 5 & 0 &  81  &  0 & 100 & 0 & 0 & 0 \\ 
\Bem{

 phase  2g  pfrak 0.1 ---- sufficient gap quot  -0.00987440659117326 

 sufficient gap quot  2.02311605911936 
  10         20         30         40         50         60         70         80         90         100       
Errors in recognizing the intrinsic clustering 
kmeans++ 0 kmeans 72 

Decisions if the data set is clusterable or not - kmeans++  
clu if correct  100  clu if not correct  0  not clu if   correct  0  not clu if not correct  0 

 }%\Bem{ 
 \hline   randu & 2g & 5 & 0.1 &  72  &  0 & 100 & 0 & 0 & 0 \\ 
%
 \hline  
\end{longtable}

%} % \IpiArchiv

In Section \ref{sec:gapinsufficiency} we drew attention to the fact that fixing gap is an insufficient criterion to define well-clusterability for $k$-means like  (or centre-based) algorithms.
The unbalanced sample can lead to change in clustering optimum position. 
To confirm it, we performed an experiment reported in Table 
\ref{tab:fixdistM} on synthetic data for a number of algorithms from $k$-means family and some other. 
 We took the $k$-means implementation $kmeans$ from the  $R$ package $stats$ in versions "Hartigan-Wong", "Lloyd", "Forgy",  "MacQueen", which were run with $nstart$ parameters equal 1, 10 and 20. Additionally, we experimented with the function $cmeans$ (implementing Fuzzy $c$-means algorithm) from the package $e1071$ in versions “cmeans" and "ufcl”. For comparison, we ran the single link algorithm as implemented in the $hclust$ function of the package $stats$. We implemented our own version of initialization of the $k$-means++ algorithm and used it in combination with the standard $kmeans$ of $R$. Two variants of $k$-means++ were used: one with a single start and one with two starts.  

%Experimental conditions are the same as previously except that 
We fix the gap parameter to be equal $g/r_{max}=2$, as e.g. proposed in so-called perfect clusterability criterion. 

With increase of the spread between cluster sizes 
we observe clearly that all but single link algorithm have increasing difficulty in detecting the "perfect" clusters. 
The additionally displayed indicator $relQ$ (relative quality, the quotient of the achieved $k$-means cost function to the smallest one) explains the reason: $k$-means optimizes $k$-means clustering function and single link does not. 
 
In  Table \ref{tab:synth} and Table \ref{tab:real}, illustrating the second and third experiment sets, 
we use the following abbreviations in column titles:
Errors $k$-means means the number of errors (not recognizing the correct clustering) 
per 100 runs by the $k$-means algorithm with random initialisation with 1 start. 
Errors $k$-means++ means the same respectively.   
WC disc. means the number of times per 100 runs that the $k$-means++ discovered the well-clusterable structure in the data, whereas WC not disc.  means the number of times $k$-means++ did not discover such a structure. 
Both are split into two categories: 
cc - correct clustering in the sense of the assumed cluster structure,
wc - wrong clustering in the same sense. 

The setup of the second experiment is as follows: 
a generator with parameters $k$ (number of clusters), $M$, $m$ (minimum and maximum cluster size), $d$ (dimensionality), $\mathfrak{p}$ (share of variance outside of core) and $gp$ (the minimum gap size compared to the one required to testify that well-clusterable structure was found). 
Some parameters are fixed for this experiment: 	$M= 75, 
	m= 45, 
	d= 10$, and only $k, \mathfrak{p}$ and $gp$ are varied. 
In each run, a new sample is generated. 

You see from Table \ref{tab:synth} that $k$-means++ almost always discovers the well-clusterable structure in the data when it exists (it failed one time in 1800 experiments) and when it failed, the discovered structure was non well-clusterable one. In 1800 experiments, when the well-clusterable structure did not exist, it did not discover a well-clusterable structure. 
It is worth noting that given the gap between the clusters is half as big as the well-clusterability gap, $k$-means++ discovered nearly always (except for 4 cases) the intrinsic structure in the data. 
Interestingly, $k$-means discovers the intrinsic structure in the data when $k=2$. 
But it fails when the number of clusters is bigger and the increase of the gaps between clusters does not improve its performance.

The setup for the third experiment is as follows. 
Datasets from the R library with at least 100 records were selected
("DNase", 
"iris", "Theoph", 
     "attenu" ,
"faithful" , "infert" , 
"quakes"  ,
   "randu" 
). 
Non-numeric columns were removed, the numerical columns were normalized (mean=0, stddev=1) and a clustering via $k$-means++ into $k$ (=2,3,5) clusters with 200 starts was performed and this was considered as the "ground truth" clustering. 
Then 100 times clustering experiments were performed computing the statistics as   in the second experiment  for the original data set and for data set modified so that the gaps between clusters were equal $g/2,g,2g$ where $g$ is the gap required by the well-clusterability condition. 
Table \ref{tab:real} presents the results for the first set and partially the second one only as the other are boringly\footnote{More experimental results are made available at 
\url{www.ipipan.waw.pl/~klopotek/ipi_archiv/kMeansEasyClustering.pdf}
} similar. 

One sees that both $k$-means++ and $k$-means do not perform well for the original data set. 
But if the gaps between clusters amount to at least half the well-clusterability gap, $k$-means++ discovers the intrinsic clustering. $k$-means does so only when $k=2$. 
$k$-means++ discovers always the well clusterable structure when it exists and does not if does not exist. 
If $k>2$, $k$-means is not helped by increasing the gap between clusters. 

Summarizing,
we have demonstrated empirically that 
(1) The gap size alone does not guarantee the discovery of clusters seen by human inspection for the class of $k$-means-like algorithms. 
(2)    If a dataset is well-clusterable as defined 
in Theorem \ref{th:clusterabilitydecidable} 
or Theorem \ref{th:coreclusterabilitydecidable}
(based on Definition \ref{def:WellClus})
then $k$-means++ is able to identify the best clustering (both for real world datasets and synthetic ones).
  (3) If $k$-means++ cannot find a clustering satisfying well-clusterability, there is no good clustering structure, fitting those definitions, hidden in data (with high probability) for all $k$-means style algorithms.

\IpiArchiv{

%-------------------TAB------------------- 
\begin{table} 
\caption{Dependence of the number of errors on   $k$. Other parameters fixed at  $d$=2 $M$=15, $m$=9 $breakgap$=1.
 .}
\label{tab:k}
\begin{center}  
\begin{tabular}{|l|r|r|r|r|} 
\hline  
$k=$& 2& 4& 8& 16    \\ 
\hline  
Hartigan-Wong 1x& 0 $\pm$ 0& 583 $\pm$ 158.5& 938.3 $\pm$ 18.7& 999.4 $\pm$ 0.8 \\
1000 execs in s / relQ & 0.24 / 1& 0.3 / 701.98& 0.44 / 54638.17& 0.68 / 3276669.6 \\
\hline  
Lloyd 1x& 0 $\pm$ 0& 655.8 $\pm$ 78.2& 952.7 $\pm$ 9.4& 999.7 $\pm$ 0.5 \\
1000 execs in s / relQ & 0.23 / 1& 0.29 / 1056.44& 0.43 / 71083.66& 0.66 / 4201145 \\
\hline  
Forgy 1x& 0 $\pm$ 0& 645 $\pm$ 75.3& 950.1 $\pm$ 8.1& 999.9 $\pm$ 0.3 \\
1000 execs in s / relQ & 0.23 / 1& 0.28 / 1112.96& 0.43 / 71431.91& 0.66 / 4175279.99 \\
\hline  
MacQueen 1x& 0 $\pm$ 0& 642.5 $\pm$ 64.2& 949.9 $\pm$ 14.7& 999.6 $\pm$ 0.7 \\
\hline  
\end{tabular}  
\end{center}  
\end{table}

%-------------------TAB------------------- 
\begin{table} 
\caption{Dependence of the number of errors on   $d$. Other parameters fixed at  $k$=5 $M$=15, $m$=9 $breakgap$=1.
 .}
\label{tab:d}
\begin{center}  
\begin{tabular}{|l|r|r|r|r|} 
\hline  
$d=$& 2& 4& 8& 16    \\ 
\hline  
Hartigan-Wong 1x& 718.9 $\pm$ 65.5& 695.8 $\pm$ 46& 748.9 $\pm$ 43.2& 750.8 $\pm$ 28.7 \\
1000 execs in s / relQ & 0.34 / 1762.09& 0.42 / 749.98& 0.59 / 474.19& 0.92 / 447.61 \\
\hline  
Lloyd 1x& 741.8 $\pm$ 59.9& 721.2 $\pm$ 31.9& 779.8 $\pm$ 35.2& 764.7 $\pm$ 14.2 \\
1000 execs in s / relQ & 0.34 / 2108.83& 0.41 / 860.33& 0.57 / 520.88& 0.9 / 470.34 \\
\hline  
Forgy 1x& 742.5 $\pm$ 52.2& 718.5 $\pm$ 28.7& 773.6 $\pm$ 41.9& 756.3 $\pm$ 16.7 \\
1000 execs in s / relQ & 0.33 / 2116.47& 0.41 / 844.73& 0.57 / 517.54& 0.89 / 467.36 \\
\hline  
MacQueen 1x& 747 $\pm$ 47.3& 716.2 $\pm$ 38.5& 765.4 $\pm$ 46.7& 751.2 $\pm$ 22.3 \\
\hline  
\end{tabular}  
\end{center}  
\end{table}

%-------------------TAB------------------- 
\begin{table} 
\caption{Dependence of the number of errors on   $M$ to $m$ proportion. Other parameters fixed at  $k$=5 $d$=2 $breakgap$=1.
 .}
\label{tab:M}
\begin{center}  
\begin{tabular}{|l|r|r|r|r|} 
\hline  
$M,m=$& 15 , 9& 30 , 9& 60 , 9& 120 , 9    \\ 
\hline  
Hartigan-Wong 1x& 723.7 $\pm$ 43.5& 801 $\pm$ 58.1& 762.9 $\pm$ 49.4& 736.1 $\pm$ 76.4 \\
1000 execs in s / relQ & 0.36 / 6216.55& 0.37 / 10441.43& 0.41 / 2427.55& 0.45 / 5018.41 \\
\hline  
Lloyd 1x& 730.4 $\pm$ 50.7& 804.9 $\pm$ 67.4& 777.3 $\pm$ 54.6& 733.8 $\pm$ 88.3 \\
1000 execs in s / relQ & 0.35 / 7085.08& 0.36 / 11054.14& 0.4 / 2456.29& 0.45 / 5112.28 \\
\hline  
Forgy 1x& 731.3 $\pm$ 50.2& 809.8 $\pm$ 67& 775.2 $\pm$ 57.9& 748.9 $\pm$ 81 \\
1000 execs in s / relQ & 0.34 / 7404.61& 0.36 / 10798.16& 0.41 / 2508.64& 0.45 / 5033.54 \\
\hline  
MacQueen 1x& 726.8 $\pm$ 44.9& 809 $\pm$ 70.7& 777.7 $\pm$ 53& 741.3 $\pm$ 88.3 \\
\hline  
\end{tabular}  
\end{center}  
\end{table}

%-------------------TAB------------------- 
\begin{table} 
\caption{Dependence of the number of errors on   cluster size, $M$ and $m$ kept proportional. Other parameters fixed at  $k$=5 $d$=2 $breakgap$=1.
 .}
\label{tab:Mm}
\begin{center}  
\begin{tabular}{|l|r|r|r|r|} 
\hline  
$M,m=$& 15 , 9& 30 , 18& 60 , 36& 120 , 72    \\ 
\hline  
Hartigan-Wong 1x& 714.5 $\pm$ 55.3& 726.7 $\pm$ 41& 751.6 $\pm$ 35.1& 739.4 $\pm$ 48.6 \\
1000 execs in s / relQ & 0.36 / 2222.82& 0.38 / 9506.9& 0.4 / 2452.25& 0.48 / 1017.77 \\
\hline  
Lloyd 1x& 731.6 $\pm$ 39.9& 737.9 $\pm$ 34.2& 743.6 $\pm$ 42.8& 733.5 $\pm$ 34.2 \\
1000 execs in s / relQ & 0.35 / 2542.88& 0.37 / 9537.46& 0.4 / 2376.59& 0.48 / 1011.3 \\
\hline  
Forgy 1x& 725.9 $\pm$ 46.8& 734.9 $\pm$ 49& 756.1 $\pm$ 39& 739.2 $\pm$ 36.8 \\
1000 execs in s / relQ & 0.35 / 2599.24& 0.37 / 9968.81& 0.39 / 2606.82& 0.48 / 1007.41 \\
\hline  
MacQueen 1x& 736.8 $\pm$ 41.3& 735.2 $\pm$ 39.8& 753 $\pm$ 42.5& 734.5 $\pm$ 35.1 \\
\hline  
\end{tabular}  
\end{center}  
\end{table}

%-------------------TAB------------------- 
\begin{table} 
\caption{Dependence of the number of errors on   $breakgap$. Other parameters fixed at  $k$=5 $d$=2 $M$=15, $m$=9.
 .}
\label{tab:breakgap}
\begin{center}  
\begin{tabular}{|l|r|r|r|r|} 
\hline  
$breakgap=$& 1& 2& 4& 8    \\ 
\hline  
Hartigan-Wong 1x& 727.8 $\pm$ 47.5& 713.8 $\pm$ 35.1& 697.3 $\pm$ 69.2& 663.3 $\pm$ 42.7 \\
1000 execs in s / relQ & 0.36 / 1312.52& 0.36 / 710.88& 0.37 / 133.68& 0.36 / 55.72 \\
\hline  
Lloyd 1x& 735.7 $\pm$ 42.4& 728 $\pm$ 35.7& 717.7 $\pm$ 29.6& 690.8 $\pm$ 32.5 \\
1000 execs in s / relQ & 0.35 / 1469.84& 0.35 / 720.23& 0.37 / 143.39& 0.36 / 61.49 \\
\hline  
Forgy 1x& 733.1 $\pm$ 39.7& 729.8 $\pm$ 27.5& 721.2 $\pm$ 39.9& 689.2 $\pm$ 22.3 \\
1000 execs in s / relQ & 0.35 / 1421.55& 0.34 / 715.94& 0.37 / 145.41& 0.35 / 63.16 \\
\hline  
MacQueen 1x& 736.1 $\pm$ 43.4& 717.4 $\pm$ 35.9& 711.9 $\pm$ 51.5& 684.9 $\pm$ 29.1 \\
\hline  
\end{tabular}  
\end{center}  
\end{table}

%-------------------TAB------------------- 
\begin{table} 
\caption{Dependence of the number of errors on   $k$. Other parameters fixed at  $d$=2 $M$=15, $m$=9 $breakgap$=1 $\mathfrak{p}$=0.1.
 .}
\label{tab:k_core}
\begin{center}  
\begin{tabular}{|l|r|r|r|r|} 
\hline  
$k=$& 2& 4& 8& 16    \\ 
\hline  
Hartigan-Wong 1x& 0 $\pm$ 0& 531 $\pm$ 93.8& 941.3 $\pm$ 11.5& 999.3 $\pm$ 0.8 \\
1000 execs in s / relQ & 0.25 / 1& 0.33 / 1484.76& 0.49 / 494624.56& 0.81 / 52872221.3 \\
\hline  
Lloyd 1x& 0 $\pm$ 0& 520.8 $\pm$ 97.7& 942.3 $\pm$ 11.7& 999.4 $\pm$ 0.7 \\
1000 execs in s / relQ & 0.24 / 1& 0.32 / 1492.16& 0.48 / 502271.64& 0.8 / 58530442.16 \\
\hline  
Forgy 1x& 0 $\pm$ 0& 524.9 $\pm$ 81.9& 940 $\pm$ 12.2& 998.8 $\pm$ 1.1 \\
1000 execs in s / relQ & 0.24 / 1& 0.32 / 1548& 0.48 / 520427.63& 0.8 / 54534626.71 \\
\hline  
MacQueen 1x& 0 $\pm$ 0& 520.5 $\pm$ 82.6& 945 $\pm$ 13.2& 999.4 $\pm$ 0.5 \\
\hline  
\end{tabular}  
\end{center}  
\end{table}

%-------------------TAB------------------- 
\begin{table} 
\caption{Dependence of the number of errors on   $d$. Other parameters fixed at  $k$=5 $M$=15, $m$=9 $breakgap$=1 $\mathfrak{p}$=0.1.
 .}
\label{tab:d_core}
\begin{center}  
\begin{tabular}{|l|r|r|r|r|} 
\hline  
$d=$& 2& 4& 8& 16    \\ 
\hline  
Hartigan-Wong 1x& 707.1 $\pm$ 50.5& 744.6 $\pm$ 38.5& 753.3 $\pm$ 21.4& 764.7 $\pm$ 19.4 \\
1000 execs in s / relQ & 0.36 / 8483.66& 0.45 / 3683.63& 0.62 / 1761.1& 0.97 / 2022.89 \\
\hline  
Lloyd 1x& 717 $\pm$ 52.5& 760.8 $\pm$ 39.9& 753.4 $\pm$ 18.8& 770.1 $\pm$ 16.8 \\
1000 execs in s / relQ & 0.35 / 8896.63& 0.43 / 3873.14& 0.61 / 1802.6& 0.95 / 2088.42 \\
\hline  
Forgy 1x& 711.6 $\pm$ 53.1& 755.1 $\pm$ 36.5& 756 $\pm$ 16.9& 765.2 $\pm$ 25.3 \\
1000 execs in s / relQ & 0.35 / 8982.63& 0.43 / 3741.16& 0.61 / 1795.73& 0.95 / 2105.67 \\
\hline  
MacQueen 1x& 720.5 $\pm$ 44.3& 753 $\pm$ 35.8& 748.9 $\pm$ 20.4& 769.5 $\pm$ 25 \\
\hline  
\end{tabular}  
\end{center}  
\end{table}

%-------------------TAB------------------- 
\begin{table} 
\caption{Dependence of the number of errors on   $M$ to $m$ proportion. Other parameters fixed at  $k$=5 $d$=2 $breakgap$=1 $\mathfrak{p}$=0.1.
 .}
\label{tab:M_core}
\begin{center}  
\begin{tabular}{|l|r|r|r|r|} 
\hline  
$M,m=$& 15 , 9& 30 , 9& 60 , 9& 120 , 9    \\ 
\hline  
Hartigan-Wong 1x& 709.6 $\pm$ 40.7& 778.2 $\pm$ 20.9& 789 $\pm$ 36.6& 765.3 $\pm$ 42.5 \\
1000 execs in s / relQ & 0.52 / 9331.26& 0.51 / 9154.2& 0.56 / 11397.3& 0.68 / 58999.38 \\
\hline  
Lloyd 1x& 730.6 $\pm$ 38.3& 781.8 $\pm$ 19.6& 786.1 $\pm$ 35.1& 780.7 $\pm$ 49.8 \\
1000 execs in s / relQ & 0.5 / 10181.63& 0.49 / 9648.81& 0.54 / 11275.24& 0.67 / 58343.78 \\
\hline  
Forgy 1x& 723.8 $\pm$ 38.2& 786.5 $\pm$ 28& 792.6 $\pm$ 37& 778 $\pm$ 42.6 \\
1000 execs in s / relQ & 0.49 / 10529.17& 0.49 / 9641& 0.55 / 11645.35& 0.68 / 58111.99 \\
\hline  
MacQueen 1x& 733.6 $\pm$ 32.4& 780.2 $\pm$ 24.7& 784.3 $\pm$ 38.6& 767.7 $\pm$ 49.8 \\
\hline  
\end{tabular}  
\end{center}  
\end{table}

%-------------------TAB------------------- 
\begin{table} 
\caption{Dependence of the number of errors on   cluster size, $M$ and $m$ kept proportional. Other parameters fixed at  $k$=5 $d$=2 $breakgap$=1 $\mathfrak{p}$=0.1.
 .}
\label{tab:Mm_core}
\begin{center}  
\begin{tabular}{|l|r|r|r|r|} 
\hline  
$M,m=$& 15 , 9& 30 , 18& 60 , 36& 120 , 72    \\ 
\hline  
Hartigan-Wong 1x& 739.4 $\pm$ 28.1& 741 $\pm$ 44& 745.7 $\pm$ 70& 713 $\pm$ 64.6 \\
1000 execs in s / relQ & 0.37 / 42678.96& 0.38 / 20676.98& 0.42 / 5466.12& 0.49 / 138752.72 \\
\hline  
Lloyd 1x& 734.4 $\pm$ 28.5& 741.4 $\pm$ 48.9& 743.5 $\pm$ 69.7& 710.4 $\pm$ 70.6 \\
1000 execs in s / relQ & 0.35 / 46792.33& 0.37 / 20423.87& 0.41 / 5537.44& 0.49 / 151687.77 \\
\hline  
Forgy 1x& 734.1 $\pm$ 34.3& 745 $\pm$ 41.4& 751.3 $\pm$ 74.9& 713.5 $\pm$ 69.7 \\
1000 execs in s / relQ & 0.35 / 46112.12& 0.36 / 20150.75& 0.41 / 5548.93& 0.49 / 144663.49 \\
\hline  
MacQueen 1x& 738.8 $\pm$ 33.8& 739.2 $\pm$ 49.6& 743.6 $\pm$ 77.5& 710.9 $\pm$ 66.3 \\
\hline  
\end{tabular}  
\end{center}  
\end{table}

%-------------------TAB------------------- 
\begin{table} 
\caption{Dependence of the number of errors on   $breakgap$. Other parameters fixed at  $k$=5 $d$=2 $M$=15, $m$=9 $\mathfrak{p}$=0.1.
 .}
\label{tab:breakgap_core}
\begin{center}  
\begin{tabular}{|l|r|r|r|r|} 
\hline  
$breakgap=$& 1& 2& 4& 8    \\ 
\hline  
Hartigan-Wong 1x& 737.8 $\pm$ 36.7& 704.8 $\pm$ 62.6& 712.5 $\pm$ 62.9& 711.6 $\pm$ 41.7 \\
1000 execs in s / relQ & 0.49 / 14832.58& 0.49 / 1527.1& 0.47 / 874.07& 0.47 / 221.09 \\
\hline  
Lloyd 1x& 744.5 $\pm$ 29.1& 724.8 $\pm$ 52.9& 726.7 $\pm$ 53.6& 731.8 $\pm$ 28.8 \\
1000 execs in s / relQ & 0.48 / 15121.58& 0.47 / 1830.33& 0.46 / 923.38& 0.46 / 246.31 \\
\hline  
Forgy 1x& 743.1 $\pm$ 30& 717.1 $\pm$ 60.7& 714.1 $\pm$ 48.2& 729.5 $\pm$ 42 \\
1000 execs in s / relQ & 0.48 / 15136.66& 0.48 / 1764& 0.46 / 938.35& 0.46 / 239.11 \\
\hline  
MacQueen 1x& 736.7 $\pm$ 21.8& 721.8 $\pm$ 58.6& 721.6 $\pm$ 57.6& 727.1 $\pm$ 31.7 \\
\hline  
\end{tabular}  
\end{center}  
\end{table}

%-------------------TAB------------------- 
\begin{table} 
\caption{Dependence of the number of errors on   $\mathfrak{p}$. Other parameters fixed at  $k$=5 $d$=2 $M$=15, $m$=9 $breakgap$=1.
 .}
\label{tab:pfrak_core}
\begin{center}  
\begin{tabular}{|l|r|r|r|r|} 
\hline  
$\mathfrak{p}$=& 0.1& 0.2& 0.3& 0.4    \\ 
\hline  
Hartigan-Wong 1x& 709.6 $\pm$ 40.7& 739.7 $\pm$ 37.2& 727.8 $\pm$ 40& 727.3 $\pm$ 30.9 \\
1000 execs in s / relQ & 0.51 / 9331.26& 0.53 / 6214.92& 0.52 / 14411.23& 0.49 / 11570.51 \\
\hline  
Lloyd 1x& 730.6 $\pm$ 38.3& 754.2 $\pm$ 37.9& 725.1 $\pm$ 35.1& 745.7 $\pm$ 37.2 \\
1000 execs in s / relQ & 0.49 / 10181.63& 0.51 / 6489.79& 0.51 / 14022.92& 0.47 / 11998.16 \\
\hline  
Forgy 1x& 723.8 $\pm$ 38.2& 749.9 $\pm$ 46.6& 725.1 $\pm$ 37& 736 $\pm$ 30.3 \\
1000 execs in s / relQ & 0.49 / 10529.17& 0.51 / 6483.41& 0.51 / 15049.05& 0.47 / 12702.78 \\
\hline  
MacQueen 1x& 733.6 $\pm$ 32.4& 745.6 $\pm$ 40.3& 728 $\pm$ 37& 737.4 $\pm$ 31.9 \\
\hline  
\end{tabular}  
\end{center}  
\end{table}

A number of further experiments have been performed on synthetic data in order to show the behaviour of several algorithms from $k$-means family on data that were generated with $k$-means++ clusterability in mind.  

\Bem{We took the $k$-means implementation $kmeans$ from the  $R$ package $stats$ in versions "Hartigan-Wong", "Lloyd", "Forgy",  "MacQueen", which were run with $nstart$ parameters equal 1, 10 and 20. Additionally, we experimented with the function $cmeans$ (implementing Fuzzy $c$-means algorithm) from the package $e1071$ in versions “cmeans" and "ufcl”. For comparison, we ran the single link algorithm as implemented in the $hclust$ function of the package $stats$. We implemented our own version of initialization of the $k$-means++ algorithm and used it in combination with the standard $kmeans$ of $R$. Two variants of $k$-means++ were used: one with a single start and one with two starts.  
}

We tested the impact of the parameters $k$ (the number of clusters), $d$ (the number of dimensions),  the cluster size and the quotient of the maximum to the minimum cluster size. Additionally, we checked to what extent the reduction of the gap between clusters impacts the performance of the clustering ($breakgap$ parameter indicates by what number the calculated clusterability gap was divided).  

The impact was measured as follows: 
10 different cluster sets were generated randomly for each value of the parameter (e.g. for $k=$2, 4, 8, and 16 clusters), separated by appropriate gap. Then for 1000 times each of the algorithms was applied. Each outcome was classified as either correct or incorrect. The outcome was correct if exactly the predefined set of clusters was discovered by the run of the algorithm. Otherwise an error was reported.  
Each cell in the tables contains four numbers: 
the average number of errors over the 10 cluster sets (number of errors out of 1000 runs), plus the standard deviation,   the time in seconds taken by 1000 runs of the algorithm and the average obtained relative quality. The relative quality was measured as a quotient of the $Q$ cost function value divided by the minimal one obtained by any algorithm for a given cluster set.

We investigated both the case of gap separation of clusters
(see tables \ref{tab:k},
\ref{tab:d},
\ref{tab:M},
\ref{tab:Mm},
\ref{tab:breakgap})
 and the gap separation of cores. 
(see tables \ref{tab:k_core},
\ref{tab:d_core},
\ref{tab:M_core},
\ref{tab:Mm_core},
\ref{tab:breakgap_core},
\ref{tab:pfrak_core}).
In the case of cores, correctness of proper classification of core elements was considered only. 

Two basic insights can be gained from all these tables. 
The well-clusterability concept defined in this paper is well-suited for the single link algorithm. 
Single link is designed to identify data where there are large gaps between clusters. It is also significantly quicker than $k$-means++.
$k$-means++ is the second best performing algorithm, but at the same time the slowest one. The poor comparison of speed results from the fact that only small data sets were used. Single link speed grows however quadratically in the size of the data set, while $k$-means speed depends quadratically on $k$ and linearly on the on the size of the data set. Besides, the memory consumption of single link is quadratic in the size of the data, while that of $k$-means++ is linear. Hence single link is a competitor for toy examples only. 

Table  \ref{tab:k} demonstrates that two-cluster task is easy for all the algorithms except for $ufcl$ variant of Fuzzy-$c$-means. 
However, with an increase in the number of clusters, the capability of detecting the optimal clustering deteriorates strongly except for the two mentioned algorithms. The $cmeans$ variant of Fuzzy $c$-means is the least affected among algorithms 1:14.
Understandably, the variants of $k$-means with more restarts perform better than those with smaller number of restarts. 
 Same conclusions can be drawn from the core-based clusterability experiment summarized in Table \ref{tab:k_core}, but of course the off-core elements contribute to worsening of the performance.  

The impact of dimensionality is shown in Tables \ref{tab:d} and   \ref{tab:d_core}. 
The increase of dimensionality slightly deteriorates the performance (compared to the impact of the number of clusters), and in case of $cmeans$ it seems to improve it. 

Tables \ref{tab:Mm} and  \ref{tab:Mm_core} present the influence of the cluster sizes on the performance. The increase of cluster size negatively influences the performance except for $cmeans$ which seems to take advantage of the increased sample size. 

The spread of sample sizes, as illustrated by Tables \ref{tab:M} and  \ref{tab:M_core} apparently does not affect the performance (though again $cmeans$ takes advantage of it in the case represented by Table \ref{tab:M}). This would mean that the spread is quite well captured by our formula on $g$. 

The effect of decreasing the gap size 
below the value determined by our formulas, as shown in  Tables \ref{tab:breakgap} and  \ref{tab:breakgap_core}  surprisingly worsens the performance of $cmeans$ and $k$-means++ only. 
Traditional $k$-means algorithms with 20 restarts outperform $k$-means++ apparently. 
This would mean that with small distances $k$-means++ loses its advantage  of high probability of hitting distant points because these distant points may be points within the same cluster rather than points of distinct clusters. 

 As visible from Table \ref{tab:pfrak_core}, the share of off-core variance seems not to affect the performance of the algorithm, except for deteriorating $cmeans$ performance. 
}%Bem 

%Bem

\section{Discussion}\label{sec:discussion}
Ben-David \cite{Ben-David:2015} investigated several clusterability notions under the following criteria:
\begin{itemize}
\item practical relevance - most real world data sets should be clusterable;
\item existence of quick clustering algorithms 
\item existence of quick clusterability testing algorithms 
\item some popular algorithms should perform well on clusterable data
\end{itemize}

As the practical relevance is concerned, Ben-David found that the clusterability criteria he investigated are usually not met because of the impractically large gap between cluster centres is required. 
Our proposed clusterability criteria are as impractical as the ones he investigated - large gaps are required and their size grows with the spread of cluster sizes, the number of clusters, the cluster radius. 
This is due among others to the fact that $k$-means does not explore directly the   gaps between clusters. 
With this respect our proposal does not constitute a progress. 

As the algorithm complexity is concerned, $k$-means++ has no computational disadvantage over the algorithms Ben-David considered. The gain we have is that usually one iteration of $k$-means is needed to establish the final result (the whole complexity resides in the initialisation stage). 

With respect to efficient testability of the clusterability conditions Ben-David complains about the problem that checking clusterability condition requires knowledge of the optimal clustering that is NP-hard to find. 
With our proposal we  make here a major progress. 
Given an output clustering, if we check that the distances between clusters match the requirements imposed on the gap $g$, we are assured that we have found the optimal clustering. 
On the other hand, if the data is clusterable in our sense, then it is highly unlikely that the standard $k$-means++ algorithm does not find it. 

With respect to the last requirement, 
we can say that $k$-means++ is actually a popular algorithm and it behaves reasonably.

\section{Conclusions}\label{sec:conclusions}

We have defined the notion of a well-clusterable data set from the point of view of the objective of $k$-means clustering algorithm and common sense in two variants - without any data points  in the large gaps between clusters and with data points there.   
The novelty introduced here, compared to other work on this topic, is that 
one can a posteriori (after running $k$-means) check if the data set is well-clusterable or not.

Let make a comparison to the results of other investigators in the realm of well-clusterability, in particular presented in   
\cite{Ackerman:2009,Ostrovsky:2013, Balcan:2009, Awasthi:2012,%
Balcan:2008, Ackerman:2016, Epter:1999}.

If the data is well-clusterable according to criteria of Perturbation Robustness,  or $\epsilon$-Separatedness, or 
$(c, \epsilon)$-Approximation-
Stability or $\alpha$-Centre Stability or $(1+\alpha)$ Weak Deletion Stability or Perfect Separation, one can reconstruct the well-clustered structure using appropriate algorithm. But only in case of Perfect Separation or Nice Separation, you can decide that you have found the  structure, if you have found it. Note that you have no warranty that you will find Nice Separation if it is there. 
But for none of these ways of understanding well-clusterability we are able to decide (neither a priori nor a posteriori) that the data is not well-clusterable if the well-clustered structure was not found (unless by brute force). 

The only exception constitutes formally the method of Multi-modality Detection, which tells us a priori that the data is or is not well-clusterable. However, as we have demonstrated, data can be easily found that can foolish this method, so that it discovers well-clusterability in case when there is none. 

Under the definitions of well-clusterability presented in this paper, we get a completely new situation. It is guaranteed that if the well-clusterable structure is there, it will be detected with high probability.  One can check a posteriori that the structure found is the well-clusterable structure if it is so, with 100\% certainty.  Furthermore if the ($k$-means++) algorithm did not find a well-clusterable structure then with high probability it is not there in the data.

The paper contains a couple of other, minor contributions. The concept of cluster cores has been introduced such that if a seed of $k$-means once hits each core then there is guarantee that none of the cluster centres will ever leave the cluster. It has been shown that the number of reruns of $k$-means++ is small when a desired probability of success in finding the well-clusterability is being targeted. 
Numerical examples show that several orders of magnitude smaller gaps between clusters, compared to \cite{Ostrovsky:2013}, are required in order to state that the data is well clusterable, and still the probability of detecting the well clusterable structure is much higher (even close to one in a single $k$-means++ run). 

The procedure elaborated for constructing a well clusterable data set,  ensuring that the $k$-means cost function absolute minimum is reached for a predefined data partition may find applications in some testing procedures of clustering algorithms. 

Of course a number of research questions with respect to the topic of this paper remain open. 
First of all the issue of constructing tight (or at least tighter) bounds for estimation of required gaps between clusters. Second an investigation how the violations of these minimum values influence the capability of $k$-means algorithms to detect either the absolute minimum of their cost function or achieving a partition that is intuitively considered by humans as "good clustering".

\bibliographystyle{plain}
\bibliography{V3_EasyClustering_bib}
\end{document}